\documentclass[pdflatex,sn-mathphys-num]{sn-jnl}

\usepackage{graphicx}%
\usepackage{multirow}%
\usepackage{amsmath,amssymb,amsfonts}%
\usepackage{amsthm}%
\usepackage{mathrsfs}%
\usepackage[title]{appendix}%
\usepackage{xcolor}%
\usepackage{textcomp}%
\usepackage{manyfoot}%
\usepackage{booktabs}%
\usepackage{algorithm}%
\usepackage{algorithmicx}%
\usepackage{algpseudocode}%
\usepackage{listings}%
\usepackage{todonotes}%
\usepackage{colortbl}
\usepackage{tikz}
\usetikzlibrary{shapes,arrows,positioning,calc,3d}
\usepackage{acro}
\DeclareAcronym{ksd}{
short = KSD,
long = kernelized Stein discrepancy
}

\DeclareAcronym{rkhs}{
short = RKHS,
long = reproducing kernel Hilbert space
}

\DeclareAcronym{uda}{
short = UDA,
long = unsupervised domain adaptation
}

\DeclareAcronym{mmd}{
short = MMD,
long = maximum mean discrepancy
}

\DeclareAcronym{tll}{
short = TLL,
long = Transfer Learning Library
}

\usepackage{amsmath,amsfonts,bm}
\usepackage{xcolor}
\usepackage{fontawesome}

\def\eqref#1{(\ref{#1})}

\def\1{\bm{1}}

\def\eps{{\epsilon}}

\def\inv{{^{-1}}}
\newcommand{\q}{_{\mathcal{D}_T}}

\def\rd{{\textnormal{d}}}

\def\ermS{{\textnormal{S}}}

\DeclareMathAlphabet{\mathsfit}{\encodingdefault}{\sfdefault}{m}{sl}
\SetMathAlphabet{\mathsfit}{bold}{\encodingdefault}{\sfdefault}{bx}{n}

\def\gA{{\mathcal{A}}}

\def\gD{{\mathcal{D}}}

\def\gF{{\mathcal{F}}}

\def\gH{{\mathcal{H}}}

\def\gL{{\mathcal{L}}}

\def\gN{{\mathcal{N}}}

\def\sE{{\mathbb{E}}}

\newcommand{\E}{\mathbb{E}}

\newcommand{\T}{\top}

\usepackage{listings}
\usepackage{color}

\theoremstyle{thmstyleone}%
\newtheorem{theorem}{Theorem}
\theoremstyle{thmstyletwo}%
\newtheorem{assumption}{Assumption}

\theoremstyle{thmstylethree}%
\newtheorem{definition}{Definition}%

\title[Stein Discrepancy for UDA]{Stein Discrepancy for Unsupervised Domain Adaptation}

\author[1]{\fnm{Anneke} \sur{von Seeger}}\email{vonse006@umn.edu}

\author[2]{\fnm{Dongmian} \sur{Zou}}\email{dongmian.zou@duke.edu}

\author*[1]{\fnm{Gilad} \sur{Lerman}}\email{lerman@umn.edu}

\affil[1]{\orgdiv{School of Mathematics}, \orgname{University of Minnesota}, \orgaddress{\street{Church Street SE}, \city{Minneapolis}, \postcode{55455}, \state{Minnesota}, \country{USA}}}

\affil[2]{\orgdiv{Data Science Research Center}, \orgname{Duke Kunshan University}, \orgaddress{\street{Duke Avenue}, \city{Kunshan}, \postcode{215316}, \state{Jiangsu}, \country{China}}}

\begin{document}

\abstract{
\Ac{uda} aims to improve model performance on an unlabeled target domain using a related, labeled source domain.
A common approach aligns source and target feature distributions by minimizing a distance between them, often using symmetric measures such as \ac{mmd}.
However, these methods struggle when target data is scarce.
We propose a novel \ac{uda} framework that leverages Stein discrepancy, an asymmetric measure that depends on the target distribution only through its score function, making it particularly suitable for low-data target regimes.
Our proposed method has kernelized and adversarial forms and supports flexible modeling of the target distribution via Gaussian, GMM, or VAE models.
We derive a generalization bound on the target error and a convergence rate for the empirical Stein discrepancy in the two-sample setting.
Empirically, our method consistently outperforms prior \ac{uda} approaches under limited target data across multiple benchmarks.
}

\acresetall

\maketitle

\section{Introduction}\label{sec:introduction}
\acresetall

Deep learning methods have been shown to outperform humans on tasks like image classification~\citep{he2015delving}, but they typically require large amounts of labeled training data and assume that the training and test data are independent and identically distributed.
In practice, both of these requirements can be difficult to satisfy.
\Ac{uda} addresses both of these challenges: it leverages information from a labeled source dataset to improve accuracy on a related but unlabeled target dataset~\citep{ben-david_analysis_2007, ben-david_theory_2010}.
Since unlabeled data is often easier and cheaper to obtain than labeled data, and relaxing the assumption that training and test data are identically distributed broadens the range of applicable datasets, \Ac{uda} has become a crucial research area for solving real-world problems.

A common approach to \ac{uda} is feature alignment~\citep{ganin_unsupervised_2015, ganin_domain-adversarial_2016, long_learning_2015}, whose goal is to learn feature representations that are informative for downstream tasks but invariant across domains.
This can be accomplished by introducing a regularization term in the loss function that seeks to minimize the distance between the source and target feature distributions.
Previous methods have used a variety of distances, including Wasserstein distance~\citep{shen_wasserstein_2018} and \ac{mmd}~\citep{long_deep_2017}, to estimate the dissimilarity between distributions.

Existing \ac{uda} methods rely on a large, unlabeled target dataset to perform feature alignment.
However, in some scenarios of interest, only a small amount of target data is available; we refer to this as the scarce target setting.
For example, in online user training, a model might be trained on large amounts of data generated by many users and then fine-tuned to give personalized results using the much smaller amount of data associated with a single user.
A \ac{uda} method that requires balanced sample sizes or abundant target data would be challenging to apply to a new user, who might have only a small amount of data available.

In such regimes, traditional symmetric distances, such as \ac{mmd} or Wasserstein, become unreliable, as they depend equally on both domains.
Stein discrepancy~\citep{stein1972bound} is a statistical distance particularly suited to settings with limited target samples.
It compares distributions by applying a Stein operator $\gA_q$ to functions from a function class $\gF$ and maximizing the resulting quantity over $\gF$.
Computable Stein discrepancies, in which the maximization has a closed-form solution, are of particular interest; \ac{ksd} is a computable Stein discrepancy that arises when $\gF$ is the unit ball in a \ac{rkhs}. 
Unlike integral probability metrics such as Wasserstein distance or \ac{mmd}, which require taking expectations over both distributions, Stein discrepancy avoids integration over the target by relying on the score function.
This asymmetry makes it effective in the scarce target setting.

In this work, we propose a novel UDA method based on Stein discrepancy; the proposed method has adversarial and kernelized forms.
The dependence of Stein discrepancy on the score function requires careful modeling of the target distribution, which our proposed method supports through options for Gaussian, GMM, or VAE priors on the target distribution.
The proposed method integrates naturally into existing UDA pipelines such as FixMatch and SPA.
Theoretically, we prove a generalization bound on the target error and a convergence guarantee for the empirical Stein discrepancy estimate.
Numerical experiments show that our approach outperforms existing methods using other domain discrepancy measures on benchmark tasks in the scarce target regime.

\subsection{Related work}

We review related works in domain adaptation and Stein discrepancy which are relevant to machine learning. 
We refer the reader to \citet{liu2022deep} and \citet{anastasiou_steins_2023} for more comprehensive reviews of these topics.

\paragraph{Domain adaptation}

\citet{ben-david_analysis_2007, ben-david_theory_2010} lay a theoretical foundation for \ac{uda}, proving a generalization bound on the target error that depends on source error and the distance between source and target distributions.
This motivated the development of feature alignment methods, which aim to learn representations that are simultaneously discriminative between classes and domain-invariant.
While the original theory uses $\mathcal{H}$-divergence, which is difficult to calculate in practice, later methods explore alternatives including Wasserstein distance \citep{courty_joint_2017}, nuclear Wasserstein \citep{chen2022reusing}, Jensen-Shannon divergence \citep{shui_novel_2022}, $\alpha$-R\'enyi divergence \citep{mansour2009multiple}, and KL divergence \citep{nguyenkl}.
Several \ac{uda} methods use \ac{mmd} \citep{long_learning_2015, long_deep_2017, rozantsev2018beyond}, which are of particular interest because of theoretical connections between \ac{mmd} and \ac{ksd}.

A very popular strategy for feature alignment uses an adversarial objective \citep{ganin_unsupervised_2015,liu_coupled_2016, zhang_bridging_2019}.
These methods are formulated as a competition between a feature generator, which extracts domain-invariant features, and a domain discriminator, which attempts to classify a sample as coming from the source or target domain.
\citet{long_conditional_2018} extended this idea to a conditional adversarial method, inspired by conditional GANs, and more recent work uses f-divergences to measure the distance between domains as part of an adversarial approach \citep{acuna2021f}.

Other approaches to domain adaptation exist, many of which are complementary to feature alignment approaches.
Importance weighting, which emphasizes source samples that are similar to the target distribution \citep{pmlr-v28-gong13, Long_2014_CVPR}, is an early and important technique for \ac{uda}. 
Another effective class of techniques, known as self-training methods, involve pseudo-labeling target samples before training a classifier on the target domain \citep{Zou_2019_ICCV}; FixMatch is a widely used method from this class \citep{sohn2020fixmatch}.
Several recent methods focus on improving the optimization method, including SDAT, which smooths the optimization landscape \citep{rangwani2022closer}, and gradient harmonization, which seeks to resolve or reduce conflicts between the direction of gradients from minimizing the classification error and the distance between domains simultaneously \citep{huang_gradient_2024}.
Graph-based methods represent the source and target features as graphs and align the source and target domains by aligning characteristics of their graphs; SPA attempts to align the graph spectra \citep{xiao2024spa}.
These methods can often be combined with feature alignment methods for improved performance by including additional components in the loss function, such as a feature alignment loss component and a self-training or graph-based loss component.
We provide examples of this by integrating the proposed method with SPA and with FixMatch \citep{xiao2024spa, sohn2020fixmatch}.

While the methods above are popular, they face significant challenges.
Adversarial methods, for example, can suffer from unstable training.
More recent work by \citet{zhao2019learning} showed that feature alignment can fail as a strategy for domain adaptation when the label distributions of the source and target are not close.
Strategies such as Implicit Alignment can improve the performance of feature alignment methods in the presence of label shift \citep{jiang2020implicit}.
Finally, many of these methods experience a severe decline in accuracy as the amount of target data is reduced.
On the other hand, the proposed Stein discrepancy-based method for \ac{uda} demonstrates a much smaller decline in accuracy at low levels of target data.

Domain adaptation has been extended in several directions from \ac{uda}.
Semi-supervised domain adaptation has access to a small number of labels for the target distribution.
Multi-source domain adaptation attempts to leverage information from several source domains at once, while multi-target attempts to improve performance over several target domains \citep{zhao2020multisourcedomainadaptationdeep}.
Open set domain adaptation allows new classes in the target dataset that are not part of the source dataset \citep{panareda_busto_open_2017}.
Finally, domain generalization and few-shot learning are both similar to the scarce target setting for \ac{uda}.
Domain generalization extends multi-source \ac{uda} by assuming that there is no access to the target data set; the goal is to learn features that are invariant to unseen distributions \citep{wang_generalizing_2022}.
Domain generalization can be viewed as an extreme version of the scarce target setting, with no target data available, but does not leverage target information when it is available.
Few-shot learning can refer to a broad class of methods focused on learning from few data points; however, unlike the scarce target setting, the data in few-shot learning is usually labeled \citep{parnami2022learning}.

\paragraph{Stein discrepancy}

Stein's method was introduced to bound distances between probability distributions \citep{stein1972bound}.
\citet{gorham_measuring_2015} build on Stein's method by formalizing Stein discrepancy as a measure of distributional difference, particularly for assessing approximate MCMC samples.
Its key advantage is applicability to unnormalized distributions, making it valuable for Bayesian inference.
Since then, Stein discrepancies have gained popularity in machine learning and statistics. 
Originally, computing Stein discrepancies involved a maximization step, 
but \citet{liu_kernelized_2016}, \citet{chwialkowski_kernel_2016}, and \citet{gorham2017measuring} 
independently developed \ac{ksd}, which provides a closed-form solution using \ac{rkhs}.
\citet{gorham2017measuring} also establish theoretical conditions under which convergence in \ac{ksd} guarantees weak convergence between distributions.
They demonstrate that in dimensions $d \geq 3$, commonly used kernels such as Gaussian and Matérn fail to detect when a sample is not converging to the target, highlighting the importance of kernel choice in practical applications.

Stein discrepancy has powered advances in inference and sampling, including Stein variational gradient descent \citep{liu_stein_2016}, Stein points \citep{chen2018stein}, and Stein thinning \citep{riabiz2022optimal}, and has been extended and improved for different applications in several subsequent works.
One area of research focuses on improving the statistical power and optimality of the \ac{ksd} test itself.
For example, \citet{schrab2022ksd} propose KSDAgg, which aggregates tests across different kernels, avoiding the loss of statistical power that comes from splitting the data to select the most effective kernel. 
Similarly, \citet{hagrass2024minimax} propose a spectral regularization of \ac{ksd}, motivated by insights from operator theory, which achieves minimax optimal goodness-of-fit testing.
This is of particular interest in the scarce target setting, as a minimax optimal test statistic may make more efficient use of scarce data.
Finally, \citet{xu_kernelised_2022} extend \ac{ksd} to non-parametric settings, where the score function of an implicit model is estimated.
This extension enables the development of two-sample tests for implicit generative models, a crucial tool in machine learning.
These applications are particularly relevant for our work, since models can generate unlimited synthetic data while real datasets remain limited.
The resulting test must handle unbalanced sample sizes, a challenge that is directly relevant to the scarce target setting in \ac{uda}. 
However, to the best of our knowledge, Stein discrepancy has not previously been applied to \ac{uda}.

\subsection{Outline}

We first review kernelized Stein discrepancy in \S~\ref{subsec:preliminaries}. 
In \S~\ref{subsec:applicationToDA}, we introduce an algorithm applying Stein discrepancy to \ac{uda}, supported by theory in \S~\ref{subsec-boundsOnTarget} and \S~\ref{subsec:convergence-rate}.
\S~\ref{sec:experiments} contains empirical results, demonstrating that the proposed method performs well in the scarce target setting.
Finally, \S~\ref{sec:discussion} discusses the strengths and limitations of our approach and concludes.

\section{Method}\label{sec:method}

We begin with an overview of Stein discrepancies and \ac{ksd}, before describing how to apply it to domain adaptation and providing theoretical results related to generalization error and convergence rate.

\subsection{Preliminaries}\label{subsec:preliminaries}
Given a distribution $q$, the starting point for Stein discrepancy is a set of functions $\gF$ known as the Stein class: $\{ f \in \gF :  \lim_{ \| x \| \to \infty}  \langle f(x), q(x) \rangle = 0\}.$
For many distributions of interest, including discrete distributions, distributions of compact support, Gaussians, and combinations of Gaussians, 
this is a mild requirement and most functions of interest are contained in $\gF$.
A Stein operator $\gA_q$ acts on functions from $\gF$.
We will focus on the score-Stein operator: \[\gA_q f(x) = f(x)^\T \nabla_x \log q(x) + \nabla_x \cdot f(x),\] also called the Langevin Stein operator, which requires the additional assumption that $\sE_{x \sim q} \| \nabla \log q(x) \| \leq \infty$ \citep{anastasiou_steins_2023}.
Finally, Stein's identity, which holds for any $f \in \gF$, states $  \sE_{x \sim q} [ \gA_q f(x) ] = 0$

If the expectation over $q$ in Stein's identity is replaced by expectation over another smooth distribution $p$, then a simple calculation shows 
\begin{align*} \sE_{x \sim p}  [ \gA_q f] = & \sE_{x \sim p} \left [ \gA_q f - \gA_p f \right ] \\
= \sE_{x \sim p} & \left[ f(x)^T \left( \nabla_x \log q(x) - \nabla_x \log p(x) \right) \right].
\end{align*}
Finding the most discriminant $f \in \gF$ gives a measure of the distance between $p$ and $q$.
\begin{definition}[Stein discrepancy]
For smooth distributions $p$ and $q$, the Stein discrepancy is defined as
\begin{equation}\label{eq-steinDisc}
        \ermS (p,q) = \sup_{f \in \gF} \sE_{x \sim p} [ \gA_q f(x) ]. 
\end{equation}
\end{definition}

The choice of the function class $\gF$ is crucial: $\gF$ should be large enough to detect differences between any two distributions of interest, while being simple enough that identifying the most discriminant $f \in \gF$ is tractable.
When $\gF$ is the unit ball of an \ac{rkhs}, the optimization has a closed form solution \citep{chwialkowski_kernel_2016,  gorham2017measuring, liu_kernelized_2016}.
An \ac{rkhs} is a Hilbert space $\gH$ associated with a reproducing kernel, $k(\cdot,\cdot)$, which is positive definite and satisfy the reproducing property: $f(x) = \langle f(\cdot), k(x, \cdot) \rangle_\gH$, for any $f \in \gF$.
A common choice of kernel is the radial basis function (RBF) kernel: $k(x, x') = \exp(-\| x - x' \|^2 / (2 \sigma^2))$, where $\sigma$ is the bandwidth.
Due to the reproducing property of the kernel, $\sE_{x \sim p} [ \gA_q f(x)]$ can be rewritten as an inner product: \[\sE_{x \sim p} [ \gA_q f(x)] = \left\langle f(\cdot ) , \sE_{x \sim p} [\gA_q k(x, \cdot) ] \right\rangle_\gH. \]
Maximizing over an inner product is straightforward. The closed form solution is called the \acl{ksd} (KSD)\footnote{\citet{chwialkowski_kernel_2016} and \citet{anastasiou_steins_2023} define \ac{ksd} with a square root over the expectation. We follow the definition from \citet{liu_kernelized_2016}, which does not include the square root, and note that since \ac{ksd} is being applied to minimize the distance between two distributions, the presence or absence of a square root will not change the minimizer.}: 
\begin{equation}\label{eq-kernelSteinDisc}
    \ermS (p,q) = \sE_{x, x' \sim p} [ \gA_q \gA_q k(x, x')].
\end{equation}

Given an independent, identically distributed sample $\{ x_i \}_{i=1}^n$ and a score function for $q$, denoted $s_q(x)$, \ac{ksd} can be estimated by a U-statistic:
\begin{equation}
        \hat{\ermS}(p,q) = \frac{1}{n(n-1)} \sum_{1 \leq i \neq j \leq n} u_q(x_i, x_j),
\end{equation}
where
\begin{align}\label{def:ustat}
        u_q (x, x' ) &= s_q (x)^\T k(x, x') s_q(x') 
        + s_q(x)^\T \nabla_{x'} k(x, x') \nonumber \\
        &~+ \nabla_{x} k(x, x')^\T s_q(x')  
        + \text{trace} ( \nabla_{x, x'} k(x, x') ).
\end{align}
This U-statistic provides a minimum-variance, unbiased estimate of $\ermS (p,q)$.

Stein discrepancy is closely related to integral probability metrics (IPMs).
IPMs include many probability metrics of interest, several of which have been applied to previous \ac{uda} methods, and can be written as \[\rd_{\gF}(p,q) = \sup_{f \in \gF} \sE_p [ f(x) ] - \sE_q[f(x)] .\]
For instance, if $\gF$ is the set of 1-Lipschitz functions, $\rd_{\gF}(p,q)$ is the 1-Wasserstein distance between the distributions.
If $\gF$ is the unit ball of an \ac{rkhs}, then $\rd_{\gF}(p,q)$ is the \ac{mmd}.
An IPM can be rewritten as a Stein discrepancy for test functions $h$ that solve the Stein equation: $\gA_q f(x) = h(x) - \sE_q [ h(x) ]$, and a solution is guaranteed to exist for many settings of interest \citep{anastasiou_steins_2023}.
The advantage in rewriting an IPM as a Stein discrepancy is that instead of taking an expectation over both distributions, as required to calculate an IPM, calculating a Stein discrepancy only requires an expectation over one distribution; the second distribution influences the Stein discrepancy only through its score function.

\subsection{Methodology}\label{subsec:applicationToDA}

To apply Stein discrepancy to domain adaptation, let $x_S, x_T$ denote samples drawn from the source and target distributions $\gD_S, \gD_T$ respectively.
Since this method is focused on \ac{uda}, source labels $y_S$ are available but no target labels $y_T$ are available.

The proposed method is a feature alignment method; the goal is to learn features that are informative for classification but invariant between domains.
To accomplish this, features $z = g(x)$ are extracted by a function $g$, which is identical in both source and target domains.
Training seeks to simultaneously minimize the transfer loss $\gL_{\text{D}}$, which measures the distance between the source and target domains, and the classification loss on the source domain $\gL_{\text{C}}$. 
Any standard classification loss can be used, such as cross-entropy loss.
Our method uses Stein discrepancy as the transfer loss and we derive two forms: an adversarial form, based on \eqref{eq-steinDisc}:
\[\gL_{\text{D}}(\gD_S, \gD_T) = \sup_{f \in \gF} \E_{\gD_S} [ \gA_{\gD_T} f(x) ]\]
and a kernelized form, based on \eqref{eq-kernelSteinDisc}:
\[\gL_{\text{D}} (\gD_S, \gD_T) = \E_{x,x' \sim \gD_S} [\gA_{\gD_T} \gA_{\gD_T} k(x,x')].\]
Training the adversarial form requires a min-max optimization because estimating $\gL_{\text{D}}(\gD_S, \gD_T)$ requires maximizing to find the most discriminant $f$.
Then given an estimate of $\gL_{\text{D}} (\gD_S,\gD_T)$, $\gL_{\text{D}}(\gD_S,\gD_T)$ and $\gL_{\text{C}}$ are both minimized.
Training the kernelized form requires only minimization.
The architecture for both forms is shown in Figure \ref{fig:architecture-diagram}.

\begin{figure}[h!]
    \centering

\begin{minipage}{0.48\columnwidth}
    \centering
\resizebox{\linewidth}{!}{
    \begin{tikzpicture}[
       box3d/.style n args={3}{
            append after command={
                \pgfextra{
                    \draw[fill=gray!20] 
                        (\tikzlastnode.north east) -- ++(0.2,0.2) -- 
                        ++(0,-#2) -- ++(-0.2,-0.2) -- cycle;
                    \draw[fill=gray!5] 
                        (\tikzlastnode.north east) -- ++(-#1,0) -- 
                        ++(0.2,0.2) -- ++(#1,0) -- cycle;
                }
            },
            minimum width=#1,
            minimum height=#2,
            draw,
            fill=white
        },
        gbox/.style={box3d={0.8cm}{2.5cm}{gray}},
        cbox/.style={box3d={0.8cm}{1.5cm}{gray}},
        bigarrow/.style={->, thick},
    ]
    
    \node[font=\Large] (xs) at (0,1) {$x_S$};
    \node[font=\Large] (xt) at (0,-1) {$x_T$};
    
    \node[gbox] (g) at (2,0) {$g$};
    \node[cbox] (c) at (6,2) {$c$};
    
    \node[font=\Large] (zs) at (4,1) {$z_S$};
    \node[font=\Large] (zt) at (4,-1) {$z_T$};
    
        \node[draw, inner sep=0pt, minimum width=1.5cm, minimum height=1cm] (gaussian) at (6,-1) {
            \begin{tikzpicture}
            \draw[smooth, domain=-1:1, variable=\x, scale=0.5] 
                plot ({\x},{exp(-(\x)^2/0.2)});
            \end{tikzpicture}
        };
    
    \node[above, font=\large\bfseries, anchor=north, xshift=1.5cm, yshift=0.75cm] at (current bounding box.north) {Kernelized Architecture};
    
    \node[below] at (gaussian.south) {$\gD_T$};
    
    \node (lce) at (9,2) {$ \mathcal{L}_{C}(\hat{y}_s, y_s)$};
    
    \node[right] (lksd) at (8,0) {$\mathcal{L}_{D}(\gD_S, \gD_T) $};
    
    \draw[bigarrow] (xs) -- ($(g.north west)+(-0.1,-0.25)$);
    \draw[bigarrow] (xt) -- ($(g.south west)+(-0.1,0.25)$);
    \draw[bigarrow] ($(g.north east)+(0.3,-0.25)$)--  (zs);
    \draw[bigarrow] ($(g.south east)+(0.3,0.25)$) --  (zt);
    
    \draw[bigarrow] (zs) -- ($(c.west)+(-0.1, 0)$);
    \draw[bigarrow] ($(c.east)+(0.3, 0)$) -- (lce);
    \draw[bigarrow] (zs) -- (lksd);
    
    \draw[bigarrow] (zt) -- ($(gaussian.west)+(-0.1,0)$);
    
    \draw[bigarrow] ($(gaussian.east)+(0.1,0)$) -- node[midway, sloped, above, font=\tiny] {$\gA_{D_T}$} (lksd);
    
    \end{tikzpicture}
}
\end{minipage}
\hfill
\begin{minipage}{0.48\columnwidth}
\centering
\resizebox{\linewidth}{!}{
    \begin{tikzpicture}[
       box3d/.style n args={3}{
            append after command={
                \pgfextra{
                    \draw[fill=gray!20] 
                        (\tikzlastnode.north east) -- ++(0.2,0.2) -- 
                        ++(0,-#2) -- ++(-0.2,-0.2) -- cycle;
                    \draw[fill=gray!5] 
                        (\tikzlastnode.north east) -- ++(-#1,0) -- 
                        ++(0.2,0.2) -- ++(#1,0) -- cycle;
                }
            },
            minimum width=#1,
            minimum height=#2,
            draw,
            fill=white
        },
        gbox/.style={box3d={0.8cm}{2.5cm}{gray}},
        cbox/.style={box3d={0.8cm}{1.5cm}{gray}},
        bigarrow/.style={->, thick},
    ]
    
    \node[font=\Large] (xs) at (0,1) {$x_S$};
    \node[font=\Large] (xt) at (0,-1) {$x_T$};
    
    \node[gbox] (g) at (2,0) {$g$};
    \node[cbox] (c) at (6,3) {$c$};
    \node[cbox] (f) at (6,1) {$f$};
    
    \node[font=\Large] (zs) at (4,1) {$z_S$};
    \node[font=\Large] (zt) at (4,-1) {$z_T$};
    
        \node[draw, inner sep=0pt, minimum width=1.5cm, minimum height=1cm] (gaussian) at (6,-1) {
            \begin{tikzpicture}
            \draw[smooth, domain=-1:1, variable=\x, scale=0.5] 
                plot ({\x},{exp(-(\x)^2/0.2)});
            \end{tikzpicture}
        };
    
    \node[above, font=\large\bfseries, anchor=north, xshift=1.5cm, yshift=0.75cm] at (current bounding box.north) {Adversarial Architecture};

    \node[below] at (gaussian.south) {$\gD_T$};
    
    \node (lce) at (9,3) {$\mathcal{L}_{C}(\hat{y}_s, y_s)$};
    
    \node[right] (lsd) at (8,0) {$\mathcal{L}_{D}(\gD_S, \gD_T) $};
    
    \draw[bigarrow] (xs) -- ($(g.north west)+(-0.1,-0.25)$);
    \draw[bigarrow] (xt) -- ($(g.south west)+(-0.1,0.25)$);
    \draw[bigarrow] ($(g.north east)+(0.3,-0.25)$)--  (zs);
    \draw[bigarrow] ($(g.south east)+(0.3,0.25)$) --  (zt);
    
    \draw[bigarrow] (zs) -- ($(c.west)+(-0.1, 0)$);
    \draw[bigarrow] (zs) -- ($(f.west)+(-0.1, 0)$);
    \draw[bigarrow] ($(c.east)+(0.3, 0)$) -- (lce);
    \draw[bigarrow] ($(f.east)+(0.3, 0)$) -- (lsd);
    
    \draw[bigarrow] (zt) -- ($(gaussian.west)+(-0.1,0)$);
    
    \draw[bigarrow] ($(gaussian.east)+(0.1,0)$) -- node[midway, sloped, above, font=\tiny] {$\gA_{D_T}$} (lsd);
    
    \end{tikzpicture}
    
}
\end{minipage}


    \caption{Architecture for Stein discrepancy-based \ac{uda}. Source and target data, $x_S, x_T$ pass through a feature extractor $g$.
    Source features $z_S$ classified by $c$ and classification loss $\gL_{\text{C}}$ is calculated.
    Target features $z_T$ are used to estimate a target distribution; the score function is $\nabla \log \gD_T$ is used in the Stein operator $\gA_{\gD_T}$.
    Top (kernelized architecture): $\gL_{\text{D}}$ is defined according to Eq. \eqref{eq-kernelSteinDisc}: $\gL_{\text{D}}= \E_{z_S} [ \gA_{\gD_T} \gA_{\gD_T} k(z_S, z_S')]$.
    Training minimizes $\gL_{\text{C}} + \lambda \gL_{\text{D}}$ over $g, c$, where $\lambda$ is a trade-off parameter between the two losses.
    Bottom (adversarial architecture): $\gL_{\text{D}}$ is defined according to Eq. \eqref{eq-steinDisc}: $\gL_{\text{D}}= \max_{f \in \gF} \E_{z_S} [ \gA_{\gD_T} f(z_S)]$.
    Training maximizes over $f$ to estimate $\gL_{\text{D}}$ and minimizes $\gL_{\text{C}} + \lambda \gL_{\text{D}}$ over $g,c$.
    }    
    \label{fig:architecture-diagram}
\end{figure}

The asymmetry in Stein discrepancy gives an advantage in the scarce target setting to Stein discrepancy-based methods over \ac{uda} methods based on traditional IPMs, such as Wasserstein distance and \ac{mmd} \citep{courty_joint_2017, long_learning_2015, long_deep_2017}.
Since the expectation is taken only over $\gD_S$, and $\gD_T$ only appears via its score function, randomness enters $\gL_{\text{D}} (\gD_S, \gD_T)$ mainly through the samples $x,x' \sim \gD_S$, and $\gL_{\text{D}} (\gD_S, \gD_T)$ can be accurately estimated even when only a small amount of target data is available.

Estimating the Stein discrepancy requires a score function for $\gD_T$ to be expressed in a parametric form. This parametric form must be simple enough to admit an explicit and tractable computation of the score function, while being flexible enough to describe complex distributions from real data.
We propose three possible models for the target distribution: a Gaussian model, a Gaussian mixture model (GMM), and a variational autoencoder (VAE).

A Gaussian distribution $\gN(\mu,\Sigma)$ with mean $\mu$ and covariance $\Sigma$ has a simple score function:
\begin{equation}
   \nabla \log \gD_T(z) = - \Sigma\inv (z - \mu).
\end{equation}
In our method, we estimate the parameters using the sample mean and sample covariance from the data.

A GMM, a weighted sum of $k$ Gaussians with weights $\{ w_i\}_{i=1}^k$ where $\sum_{i=1}^k w_i =1$, has a score function that is closely related to the Gaussian score function:
\begin{align*}
   &  \nabla \log \gD_{T}(z)  = -\sum_{i=1}^k \gamma_i(z) \Sigma_i^{-1}(z - \mu_i), \\
    & \quad \gamma_i(z) = \frac{w_i \mathcal{N}(z|\mu_i, \Sigma_i)}{\sum_{j=1}^k w_j \mathcal{N}(z|\mu_j, \Sigma_j)}.
\end{align*}
The weights and parameters can be estimated using the EM algorithm or updated via gradient descent.

A VAE consists of an encoder $\mathbf{E}$ and decoder $\mathbf{D}$ that map between data and a lower-dimensional latent space with a simple prior, typically $\xi \sim \gN(\mu, \Sigma)$.
Sampling from this latent space and decoding yields samples from the target distribution \citep{luo_understanding_2022}. The score function for the target distribution $p(z)$ is given by:
\begin{equation}\label{def-VAEScoreFunction}
         \nabla_z \log p(z)  = \E_{q (\xi | z)} \left[ \nabla_z p (z|\xi) p (z | \xi)\right],
\end{equation}
where
\begin{equation*}
\begin{aligned}
    &\quad \nabla_z p (z | \xi) p (z | \xi) = \\
    & \left [ \frac{1}{(2 \pi)^{d/2} }  \exp  \left ( - \frac{\| z - \mathbf{D}(\xi) \|^2}{2} \right )\right  ]^2  \left ( \mathbf{D}(\xi) - z  \right). 
\end{aligned}
\end{equation*}
The parameters of the VAE are trained together with other parameters in our model.

We present the detailed derivation of the score function of a VAE.
Let $p(z)$ be the data distribution. 
Let $p(z | \xi) = \textbf{D} (\xi)$ represent the likelihood and $q(\xi | z) =\textbf{E}(z)$ represent the approximate posterior.
We estimate the score function $\nabla_z \log p(z)$ as follows:
\begin{align}
    \nabla_z \log p(z) & = \frac{1}{p(z)} \int \nabla_z p(z, \xi) d\xi \nonumber \\
    & = \frac{1}{p(z)} \int \frac{\nabla_z p(z, \xi) q( \xi | z)}{q(\xi | z)} d \xi \nonumber \\
    & = \E_{q ( \xi | z)} \left [ \frac{\nabla_z p(z | \xi) p (\xi)}{p(z) q (\xi | z) }\right] \nonumber \\
    & \approx \E_{q(\xi | z) } \left [ \frac{\nabla_z p(z | \xi) p(\xi) }{p(z) p (\xi | z) }\right ] \nonumber \\
    & = \E_{q(\xi | z)} \left [ \nabla_z p(z|\xi) p(z | \xi) \right ]. \label{eq:eqdpp}
\end{align}
Assuming $p(z | \xi)$ is a Gaussian distribution:
\begin{align}
    p(z| \xi) & = \frac{1}{(2 \pi)^{d/2}} \exp \left \{ - \frac{\| z - \mathbf{D}(\xi) \| ^2}{ 2} \right \}, \label{eq:pzxi}
\end{align}
where $d$ is the dimension of the feature space.
Combining \eqref{eq:eqdpp} and \eqref{eq:pzxi} yields
\begin{align*}
    &\nabla_z \log p(z) \\
    &= \left[ \frac{1}{(2 \pi)^{d/2}} \exp \left \{ -\frac{\| z - \mathbf{D}(\xi) \|^2  }{2} \right \} \right]^2 (\mathbf{D}(\xi) - z),
\end{align*}
where $\xi \sim \mathbf{E}(z)$.


Each model for the target distribution offers a tradeoff.
The Gaussian is simple and stable but lacks flexibility.
The VAE is the most expressive, but introduces additional training complexity and potential instability, since its parameters must be optimized while the feature distribution is also shifting.
A GMM can provide a good balance between flexibility and stable training.

While assuming a simple parametric form, such as a Gaussian or GMM, for high-dimensional input data would be restrictive, it is important to distinguish between the input space and the latent feature space induced by the network.
Deep feature extractors, including the ResNet architectures used in our numerical experiments, are designed to disentangle complex data manifolds into structured, low-dimensional representations.
Theoretical frameworks such as the manifold hypothesis \citep{bengio2013representation} and empirical observations of neural collapse \citep{papyan2020prevalence} suggest that well-trained networks impose a strong inductive bias, driving features towards low-rank, clustered configurations in the feature space.
Consequently, the feature space is amenable to parametric modeling.

In this context, the choice of a parametric model, whether Gaussian, GMM, or VAE, serves as a tractable proxy for estimating the domain discrepancy, not as a ground-truth generative description.
This proxy remains effective even under domain shift because the feature extractor is initialized with robust pre-trained weight and is simultaneously constrained by the source classification task.
As a result, the target features inherit a preliminary, albeit shifted, structure.
The parametric model captures this relatively simple initial geometry, and through minimization of Stein discrepancy, guides the target features into tighter alignment.
Our empirical results demonstrate that, effectively regularized by the feature extractor's inductive bias, even simple parametric assumptions give effective adaptation results.

\subsection{Bounds on target error}\label{subsec-boundsOnTarget}

We prove a generalization bound on the target error, making use of theoretical framework developed for domain adaptation \citep{ben-david_analysis_2007, ben-david_theory_2010, long_learning_2015} and Stein discrepancies \citep{anastasiou_steins_2023, liu_kernelized_2016}.

\begin{theorem}\label{thm:generalization}
    Let $\gD_S, \gD_T$ be probability distributions on the feature space $X$ and $\gF$ be the unit ball of an \ac{rkhs} with kernel $k(x,x')$, with $x, x' \in X$.
    Let $f^*_S$ and $f^*_T$ denote the true labeling functions associated with the source and target distributions, respectively. Let $\eps_T(f) = \E_{x \sim \gD_T} [  | f(x) - f^*_T(x) | ]$ be the error function in the target domain, and ${\eps}_S(f)$ defined similarly for the source domain.
    Then the following bound holds for any labeling function $f \in \gF$:
    \begin{equation}
        \eps_T(f) \leq \eps_S(f) + 2 \sqrt{\ermS(D_S, D_T)} + C,
    \end{equation}
    where $\ermS(\cdot,\cdot)$ is the Stein discrepancy, 
     and $C$ depends on $\gF$ and sample size.
\end{theorem}

Theorem~\ref{thm:generalization} suggests that minimizing \ac{ksd} and the source error together will minimize the target error. 
Similar generalization bounds have been shown for other discrepancies, in particular for \ac{mmd}, and the proof relies on a connection between \ac{ksd} and \ac{mmd} \citep{liu_kernelized_2016}.

\begin{proof}
We first show that \ac{ksd} can be viewed as a special case of squared \ac{mmd}. 
Since \ac{ksd} provides the exact Stein discrepancy when maximizing over a \ac{rkhs}, showing the relationship between \ac{ksd} and \ac{mmd} establishes the bound for the general Stein discrepancy.

Since $\gF$ is the unit ball of an RKHS, the MMD between $\gD_S$ and $\gD_T$ can be written as
\begin{equation*}
d_{\text{MMD}}(\gD_S, \gD_T) = \sup_{f \in \gF} \left |\E_{x \sim \gD_S} [f(x)]  - \E_{x \sim \gD_T} [f(x)] \right |.
\end{equation*}
Moreover, \citet{gretton_kernel_2012} showed that the squared MMD can be written in a kernelized form 
\begin{align*}
    d_{\text{MMD}}^2 & (\gD_s, \gD_T) = \E_{x, x' \sim \gD_S} [k(x, x')] - \\
    & 
    2 \E_{x \sim \gD_S, y \sim \gD_T} [k(x,y)] + \E_{y, y' \sim \gD_T} [ k(y,y')]. 
\end{align*}

\citet{liu_kernelized_2016} verified that $\gA_{\gD_T} \gA_{\gD_T} k(x,x')$ is a valid positive definite kernel and is contained in the Stein class of $\gD_S$, as long as the same is true for $k(x,x')$.
Using Stein's identity, which states that $\E_q [ \gA_q f(x) ] = 0$, it follows that the squared \ac{mmd} and Stein discrepancy are equivalent:
\begin{align*}
    d_{\text{MMD}}^2(\gD_s, \gD_T) &  = \E_{x, x' \sim \gD_S} [\gA_{\gD_T} \gA_{\gD_T} k(x,x')] \\
    & \quad - 2 \E_{x \sim \gD_S} \E_{y \sim \gD_T} [\gA_{\gD_T} \gA_{\gD_T} k(x,y)] \\
    & \quad + \E_{y \sim \gD_T} \E_{y' \sim \gD_T} [ \gA_{\gD_T} \gA_{\gD_T} k(y,y')] \\
    & = \E_{x, x' \sim \gD_S} [\gA_{\gD_T} \gA_{\gD_T} k(x,x')] \\
    & \quad - 2 \E_{x \sim \gD_S} [0] + \E_{y \sim \gD_T} [0] \\
    & = \E_{x, x' \sim \gD_S} [\gA_{\gD_T} \gA_{\gD_T} k(x,x')] \\
    & = \ermS(\gD_S, \gD_T).
\end{align*}

The proof follows from combining the above equation and the result of \citet{long_learning_2015}, which states that $$\eps_T(f) \leq \eps_S(f) + 2 d_{\text{MMD}}(\gD_S, \gD_T) + C,$$ where $C$ depends on the VC dimension of $\gF$, the sample size from each distribution, and the smallest possible test errors in both domains.
\end{proof}


\subsection{Convergence rate of empirical estimate}\label{subsec:convergence-rate}

\citet{liu_kernelized_2016} provide a convergence rate of $O(1/\sqrt{n})$ for an empirical estimate of the Stein discrepancy in goodness-of-fit testing, under the assumption that samples are available from one distribution, the exact score function of the other distribution is known, and the two distributions are not equal.
Estimating the score function from samples introduces an additional source of error, and the setting shifts from goodness-of-fit to two-sample testing.
Theorem \ref{thm:error-rate} establishes a convergence rate for \ac{ksd} in two-sample testing.
To the best of our knowledge, this is the first such  rate for \ac{ksd}, when the score function is estimated from samples rather than known exactly.

Let $s_q(x) = \nabla_x \log q(x)$ be the true score function of the target distribution $q = \gD_T$, which depends on parameters $\theta$.
Let $\hat{\theta}_m$ be an estimate of $q$'s parameters based on $m$ samples.
Define the plug-in score estimator as $\hat{s}_q(x) = s(x ; \hat{\theta}_m)$.
Replacing the true score $s_q$ with the estimated score $\hat{s}_q$ in the U-statistic from Eq. \eqref{def:ustat} gives the estimated U-statistic $\hat{u}(p,q)$.
The final statistic based on $n$ samples from $p = \gD_S$ calculated using this estimated $\hat{u}(p,q)$:
\begin{equation}
    \hat{S}_{\hat{u}}(p,q) = \frac{1}{n(n-1)}\sum_{1 \leq i \neq j \leq n} \hat{u}_q(x,x')
\end{equation}

\begin{assumption}\label{assump:m-estimator}
Given distribution $\mathcal{X}$, an M-estimator $\psi(\mathcal{X}, \theta)$ and population parameter $\theta^*$, assume that $\theta^*$ is the unique value for which $\E [ \psi(\mathcal{X}, \theta^*) ] = 0$, and that the estimator has finite variance, i.e. $\E [ \| \psi(\mathcal{X}, \theta^*) \|^2 ] < \infty$.
In addition, assume that $\psi$ satisfies the following regularity conditions:
$\psi(\mathcal{X}, \theta)$ is continuously differentiable in an open neighborhood of $\theta^*$,
$H_0 = \E \left[ \nabla_{\theta} \psi (\mathcal{X}, \theta^*) \right]$ is invertible,
and $\E \left[ \sup_{\theta \in N(\theta^*) } \| \nabla_{\theta} \psi (\mathcal{X}, \theta ) \| \right] < \infty$.
\end{assumption}

\begin{theorem}\label{thm:error-rate}
Assume that the source distribution $\mathcal{D}_S$ is independent of $\mathcal{D}_T$, $\mathcal{D}_S \neq \mathcal{D}_T$, and $\mathcal{D}_S$ has continuous density.
Assume that $\mathcal{D}_T$ has score function $s_{\mathcal{D}_T}$, which is differentiable with respect to parameters $\theta$.
Assume that $\hat{\theta}_m$ can be estimated by an M-estimator $\psi(\mathcal{D}_T, \theta)$ that satisfies Assumption \ref{assump:m-estimator}.
Assume the kernel $k(x,x')$ is integrally strictly positive definite,
$\mathbb{E}_{x, x' \sim \mathcal{D}_S}[ u\q (x,x')^2 ] :=\sigma_u^2 < \infty$, and $\mathbb{E}_{x,x' \sim \mathcal{D}_S}[\nabla u\q(x,x'; \theta^*)] := G $, where $G$ is finite.
Also, let $g_{p,q}(x) = p(x) (s_q(x)  - s_p(x))$ and assume $\| g_{\mathcal{D}_S, \mathcal{D}_T} \|^2_2 < \infty$, 

Suppose we have $m$ i.i.d. samples from the target distribution $\mathcal{D}_T$ and $n$ i.i.d. samples from the source distribution $\mathcal{D}_S$.
Then for sufficiently large sample sizes $m$ and $n$, $\hat{\ermS}_{\hat{u}}(\mathcal{D}_S, \mathcal{D}_T)$ converges to $\ermS(\mathcal{D}_S, \mathcal{D}_T)$ with rate $O(n^{-1/2} + m^{-1/2})$.
\end{theorem}


    The error in $\hat{S}_{\hat{u}}(\mathcal{D}_S, \mathcal{D}_T) - S(\mathcal{D}_S, \mathcal{D}_T)$ comes from two sources: approximating $S$ by $\hat{S}$ and approximating $\hat{S}$ by $\hat{S}_{\hat{u}}$.
The first approximation has variance $\sigma_u^2/n$  \citep{liu_kernelized_2016}.
In the second approximation, estimating the parameter $\hat{\theta_m}$ converges at rate $O(\sqrt{m})$ by standard results on M-estimators and estimating $\hat{S}_{\hat{u}}$ inherits that rate by the delta method.
Combining these two gives an overall rate of $O(n^{-1/2} + m^{-1/2})$.


\begin{proof}
    Let $\{x_i\}, \{ y_i\}$ denote samples from $\gD_S, \gD_T$, respectively.
    Split up $\hat{S}_{\hat{u}} - S$ into $E_1 = (\hat{S}_{\hat{u}} - \hat{S})$ and $E_2 = (\hat{S}- S)$, so $\hat{S}_{\hat{u}} - S = E_1 + E_2$.
    From \citet[Theorem 4.1]{liu_kernelized_2016}, $\sqrt{n} (S(\mathcal{D}_S, \mathcal{D}_T) - \hat{S}(\mathcal{D}_S, \mathcal{D}_T)) \overset{d}{\to} N(0, \sigma^2_u)$ for sufficiently large $n$.
    
    We would like to show that $\sqrt{m} E_1$ converges in distribution to a normal distribution with mean 0.
    We will apply the delta method and standard results of M-estimators to show convergence in distribution of $\hat{u}\q$ to $u\q$.
    The main idea is that given convergence in distribution of a parameter $\theta_m$ to a true value $\theta^*$ and a differentiable function $f$, $f(\theta_m)$ will converge at the same rate, rescaled by $\nabla_{\theta} f(\theta^*)$.
    See \citet[Chapters 3 and 5]{van2000asymptotic} for a rigorous explanation of the delta method and M-estimators.

    Fix a pair $x, x'$.
    $\sqrt{m} \left( \hat{\theta}_m - \theta^* \right) \overset{d}\to N(0, \Sigma_{\theta})$ for sufficiently large $m$ by standard results on M-estimators \citep{van2000asymptotic}.
    From the definition of $u\q(x,x')$, it is clear that as long as $s\q$ is differentiable with respect to $\theta$, which is true by assumption, $u\q$ is also differentiable.
    Then applying the delta method \citep{van2000asymptotic}, it follows that
    \begin{align*}
        \sqrt{m} \left(\hat{u}\q (x,x') - u\q(x,x') \right) \overset{d}{\to}  \\
        N \left(0, \nabla_{\theta}u\q(\theta^*) \Sigma_{\theta} \nabla_{\theta}u\q(\theta^*)^T \right).
    \end{align*} 
    In addition, the difference between $\sqrt{m}(\hat{u}\q (x,x') - u\q(x,x')) $ and $\sqrt{m} \nabla_{\theta} u\q (x,x') (\hat{\theta}_m - \theta^*)$ converges to 0 in probability.
    Therefore we can rewrite $E_1$ as \begin{align*}
       &  \frac{1}{n(n-1)} \sum_{1 \leq i \neq j \leq n} \hat{u}\q(x_i, x_j) - u\q(x_i, x_j) \\
       & =\frac{1}{n(n-1)} \sum_{1 \leq i \neq j \leq n} \nabla_{\theta} u\q(x_i, x_j)(\hat{\theta}_m - \theta^*) .
    \end{align*}
    The average of the gradient terms converges to a constant by the weak law of large numbers for U-statistics: \begin{align*}
    & \frac{1}{n(n-1)} \sum_{1 \leq i \neq j \leq n} \nabla_{\theta} u\q(x_i, x_j) \overset{p}{\to} \\
    & \mathbb{E}_{x,x' \sim \mathcal{D}_S} [ \nabla_{\theta} u(x,x')] = G.
    \end{align*}
    Therefore by Slutsky's theorem, \[\sqrt{m} E_1 \overset{d}{\to}  N \left(0, G \Sigma_{\theta} G^T \right). \]

    Having established $E_1 = O(m^{-1/2})$ and $E_2 = O(n^{-1/2})$, the total error is the sum $\hat{S}_{\hat{u}} - S = E_1 + E_2$.
    By standard properties of error rate analysis, the rate of the sum is  $O(\max\{ n^{-1/2}, m^{-1/2}\} ).$
    Since $ \max\{ n^{-1/2}, m^{-1/2}\} \leq n^{-1/2} + m^{-1/2}$, we conclude that  the overall error rate is $O(n^{-1/2} + m^{-1/2})$.
\end{proof}

Theorem~\ref{thm:error-rate} verifies that the empirical estimate of the Stein discrepancy using samples from both distributions, instead of samples from one distribution and the exact score function from the other, still converges to the true value.
It describes a broad class of distributions for which \ac{ksd} can be reliably estimated from samples from two distributions.
In particular, the three target distributions that we evaluate for our method, Gaussian, GMM, and VAE with normal latent distribution, are examples of the class of problems our theorem addresses. 
The Gaussian distribution is a textbook example: if the parameter vector is $\theta = (\mu, \Sigma)$, the M-estimator of its mean and covariance  is defined by the function $\psi(X, \theta) = \left[ (X-\mu), \text{vec}\left( X X^T - \Sigma \right) \right]^T$.
The finite variance requirement in Assumption~\ref{assump:m-estimator} then implies that the distribution must have finite fourth moments.
While a rigorous verification of the regularity conditions for GMMs and VAEs is complex and beyond the scope of this paper, our empirical results suggest the convergence holds.


The number of target samples $m$ also dictates the validity of our asymptotic analysis for the $E_1$ error term.
Our proof's convergence rate depends on the asymptotic normality of the M-estimator $\hat{\theta}_m$, which itself depends on $m$ being large enough for the Central Limit Theorem to apply.
While our final asymptotic rate does not specify a concrete number of samples required, this prerequisite for the M-estimator provides a practical lower bound.
To return to the example of a Gaussian target distribution, a heuristic threshold of $m \geq 30$ is often used in practice, to consider the CLT a reasonable approximation.
Our experiments, which use a minimum of 32 target samples in the scarce setting, are chosen to respect this practical consideration.

For context, this rate is the same as the rate of convergence for \ac{mmd} \citep{gretton_kernel_2012}.
While \ac{ksd} and \ac{mmd} have the same theoretical rate, the usefulness of a rate cannot be separated from the stability of the estimator and
\ac{ksd} has been observed to outperform \ac{mmd} under sample imbalance and small sample sizes.
\citet{xu_kernelised_2022} show that the type I error rate in two-sample testing for \ac{mmd} can be as high as 100\% with samples of sizes 50 and 1000 and significance level 0.05.
On the other hand, \ac{ksd} is not negatively impacted by unbalanced sample sizes to the same extent and maintains an error rate under 10\% in the same setting \citep{xu_kernelised_2022}.

This observed finite-sample advantage, despite an identical asymptotic rate, is not unique to our setting.
A strong precedent is found in the analogous problem of composite goodness-of-fit (CGOF) testing, which aims to estimate the parameters of a distribution under the assumption that it comes from a certain parametric family before computing \ac{ksd} or \ac{mmd} for goodness-of-fit testing.
CGOF is very similar to our setting, where the target distribution is assumed to come from a distribution family and the parameters are estimated before calculating \ac{ksd}.
In CGOF, \citet{key2025compositegoodnessoffittestskernels}
 also observes that \ac{ksd} has higher statistical power than \ac{mmd} across sample sizes ranging from 10 to 500, even as the power of both tests converges for larger samples.

These experimental results, combined with Theorem~\ref{thm:error-rate}, raise a key theoretical question: if \ac{ksd} and \ac{mmd} have the same asymptotic rate of convergence, why does \ac{ksd} consistently offer an advantage over \ac{mmd} in the scarce target regime?
\citet{key2025compositegoodnessoffittestskernels} strengthens the hypothesis that the asymptotic rate is not the sole determinant of practical performance.
\citet{key2025compositegoodnessoffittestskernels} hypothesize that \ac{ksd} shows higher power than \ac{mmd} for composite goodness-of-fit testing because \ac{ksd} uses the score function directly, while \ac{mmd} requires sampling from the target distribution.
Another possible explanation is that for small sample sizes, assuming a parametric distribution to estimate \ac{ksd} has a favorable bias-variance tradeoff.
\ac{mmd} is defined by the distance between mean embeddings, which require estimating expectations over the entire support of both distributions \citep{gretton_kernel_2012}.
This does not require any parametric assumptions, but the empirical distribution is a high-variance proxy for the true distribution when few samples are available.

In contrast, \ac{ksd} replaces the high-variance, non-parametric estimate of the target expectation with a lower variance, parametric estimate of the target score function, $\nabla_x \log q(x)$.
This idea is the cornerstone of score-based generative models, which have demonstrated that learning the score function is a powerful and sample-efficient proxy for learning the full data density, which is often intractable \citep{song2019generative, chen2022sampling}.
The success of these models is partly because the score function can often be accurately approximated with a parametric model, such as a neural network, even when the underlying density is complex.
By leveraging a parametric score, \ac{ksd} can capture meaningful distributional properties from far fewer samples than MMD.

Assuming a parametric distribution for \ac{ksd} introduces the risk of a misspecified model, which would bias the estimate, but reduces the variance.
The risk of a misspecified model can be ameliorated by choosing a more flexible target distribution, for example a GMM distribution instead of Gaussian distribution, at the cost of a potential increase in variance.
A full theoretical analysis of this tradeoff between bias and variance in calculating \ac{ksd} and \ac{mmd} for small sample sizes is a rich area for future work.

\section{Experiments}\label{sec:experiments}

We evaluate our method against 10 baseline \ac{uda} methods on benchmark datasets for \ac{uda}. 
Our code is available on \href{https://github.com/amvs/stein-disc}{Github}.
For datasets with multiple domain pairs, we report the average across domain pairs in this section.
Tables with full results are included in Appendix~\ref{app:add-experiments}.

\subsection{Setup}\label{subsec:setup}

\subsubsection{Datasets}
We evaluate on three standard \ac{uda} benchmarks: Office31 \citep{saenko2010adapting}, Office-Home \citep{venkateswara2017deep}, VisDA-2017 \citep{peng2017visda}.
Office31 includes approximately 4600 images across 31 classes and 3 domains, while Office-Home includes over 15,000 images across 65 classes and 4 domains.
VisDA-2017 includes over 280,000 images across synthetic (S) and real (R) domains.
We evaluate all domain pairs for Office31 and Office-Home; VisDA-2017 is evaluated on transfer from synthetic to real.

\subsubsection{Baseline methods}

We compare our method against Deep Adversarial Neural Network (DANN) \citep{ganin_unsupervised_2015}, Joint Adaptation Network (JAN) \citep{long_deep_2017}, Adaptive Feature Norm (AFN) \citep{AFN}, Margin Disparity Discrepancy (MDD) \citep{zhang_bridging_2019}, and Minimum Class Confusion (MCC) \citep{MCC} f-Domain Adversarial Learning (FDAL) \citep{acuna2021f} Graph Spectral Alignment (SPA) \citep{xiao2024spa}, Discriminator-free Adversarial Learning Network, which uses nuclear Wasserstein Distance (NWD) \citep{chen2022reusing}, Smooth Domain Adversarial Training (SDAT) \citep{rangwani2022closer}. Empirical Risk Minimization (ERM), which trains only on source data, is included as a naive model that all \ac{uda} methods are expected to outperform. 
JAN is of particular interest because of its use of \ac{mmd} as the domain discrepancy measure.

For our proposed method, we present several variants to demonstrate how it can be combined with recent advances in \ac{uda} to improve performance in the scarce target setting.
The standard implementation closely follows the framework of JAN, replacing the \ac{mmd}-based transfer loss with a Stein discrepancy-based loss.
To demonstrate integration with a self-training approach, the proposed method is combined with FixMatch by adding a Stein discrepancy term to the loss function, in addition to the classification loss and pseudo-label classification loss used in the original FixMatch method \citep{sohn2020fixmatch}.
To demonstrate integration with a recent method that achieves state-of-the-art performance, we integrate our method with SPA,
replacing the domain distance, which was DANN or CDAN in the original implementation, with Stein discrepancy \citep{xiao2024spa}.
Results are grouped by implementation (plain, FixMatch, and SPA).

\subsubsection{Implementation and training}

To ensure fair comparisons, all methods were implemented within a consistent framework based on the \ac{tll} \citep{tllib, jiang2022transferability}.
Baseline methods implemented in \ac{tll}, including DANN, JAN, AFN, MDD, MCC, and ERM, were used directly, with small modifications to implement the scarce target setting.
FDAL, SPA, NWD, and SDAT were implemented following the code made available in the original papers, with modifications to ensure data preprocessing and augmentation was consistent with TLL.
Details on accessing this code are found in Appendix \ref{app:implementation}.

To simulate scarce target data, we use 32 samples for Office31 and at least 1\% or 32 samples (whichever is larger) for Office-Home.
For VisDA-2017, we consider both 1\% and 0.1\% of the target data to ensure scarcity; 0.1\% corresponds to approximately 55 samples for VisDA-2017.
We modified \ac{tll}'s data preprocessing pipeline to allow specifying the percentage of target data and a minimum number of target samples that should be included in training.
The validation and test datasets are unchanged.

All models use ResNet-50 \citep{he2016deep} as the feature extractor on the Office31 and Office-Home datasets and ResNet-101 on VisDA-2017, followed by a single-layer classifier.
For the kernelized methods, an RBF kernel was used and the code to calculate \ac{ksd} is adapted from \citep{korba_kernel_2021}.

Our hyperparameter tuning strategy was designed to ensure a fair and rigorous comparison. 
For all baseline methods, we used the default hyperparameters provided in their original implementations.
For our proposed Stein discrepancy-based methods, we performed a hyperparameter search for each method using Raytune.
The process began by tuning the learning rate, followed by a search over other key parameters such as momentum, the transfer loss trade-off schedule, and the bottleneck dimension.
Method-specific hyperparameters, including kernel types and bandwidths for kernelized methods and architectural choices for adversarial methods, were also optimized.
Through this process, we identified several consistent findings; for instance, the radial basis function (RBF) kernel and a hyperbolic tangent rescaling function consistently yielded the best performance.
The same set of optimized hyperparameters was used for a given method across all domain adaptation tasks to ensure consistency.
A complete list of the final hyperparameter values for each variant of the proposed method on each dataset is provided in Appendix \ref{app:implementation} for full reproducibility.

Our methods are trained with stochastic gradient descent using the learning rate and momentum values listed in \S\ref{subsec:hyperparams}, and weight decay fixed at $5 \times 10^{-4}$. We use a Reduce‑On‑Plateau scheduler on each optimizer, with a batch size of 32 for both source and target data, and clip gradients to a maximum norm of 5.

\subsubsection{Evaluation}

To ensure reproducibility, each experiment was run three times with the random seed set to 0.
Reported results are the mean and standard deviation over those three runs unless otherwise stated, as in  \S\ref{subsec:visda-stat-sig}.
Following standard practice, we select the best model checkpoint for evaluation based on its accuracy on the target domain's validation set.

\subsection{Results}\label{subsec:results}

\begin{figure}[h!]
    \centering
    \includegraphics[width=0.7\columnwidth]{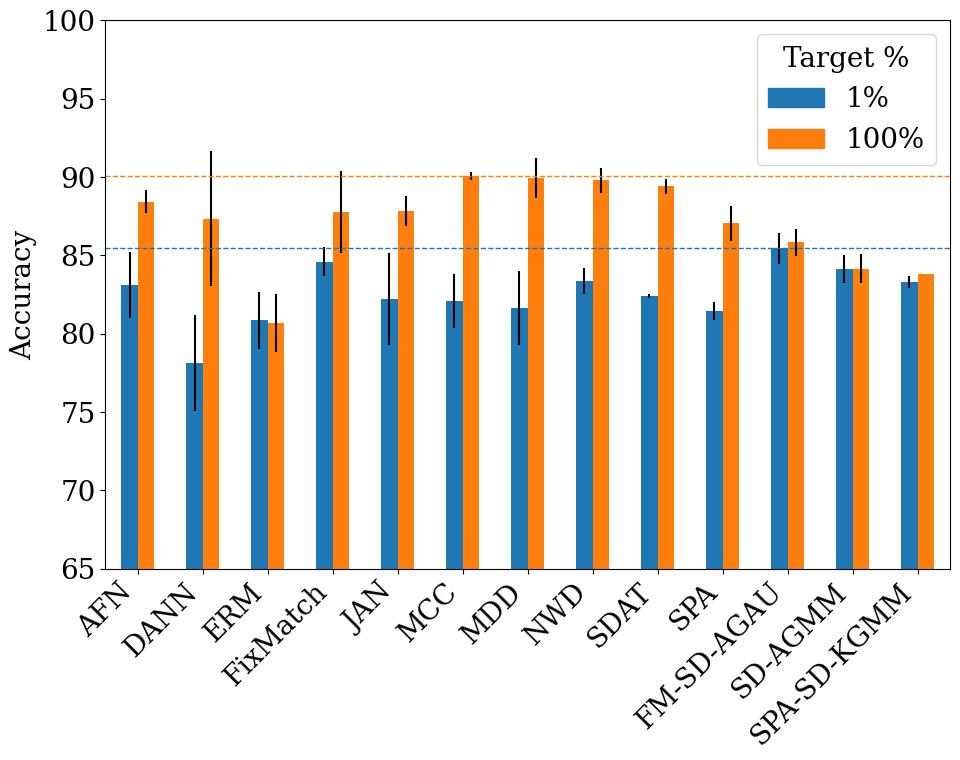}
    \caption{Accuracy on Office31, averaged across six domain pairs. Orange bars use all target data; blue bars use at most 1\% (or 32) target examples.}
    \label{fig:o31-main}
\end{figure}

The results on the three benchmark datasets are displayed in Figures~ \ref{fig:o31-main}-\ref{fig:vis-main}.
Stein discrepancy-based methods are labeled SD, with suffixes indicating the variant: A for adversarial loss, K for kernelized loss, and GAU, GMM, or VAE for the assumed target distribution.
Results are reported for all baseline methods and for the highest performing variant of Stein discrepancy in each framework, plain, FixMatch, and SPA.
For Office-31 and OfficeHome, the reported results are averaged across domain pairs; per-domain tables are included in Appendix \ref{app:add-experiments}.
Each method was trained and evaluated three times; we report mean validation accuracy with error bars denoting standard deviation, pooled across domain pairs where applicable.
Random seeds were fixed so that each method used the same target data in the scarce setting, isolating performance variance from sampling effects.
Appendix \ref{app:add-experiments} also includes sensitivity analyses for different target samples in the scarce setting.

\subsubsection{Office31}

In the scarce target setting, Stein discrepancy-based methods outperform other methods.
The best performing variants in the plain, FixMatch, and SPA framework were AGMM, AGAU, and KGMM, respectively.
The combined Fixmatch and Stein discrepancy method has the best performance overall in the scarce target setting, with an improvement of 0.9\% over the next best method, FixMatch.
In the plain framework, SD-AGMM achieves the highest performance, and outperforms all baseline methods except for FixMatch.
Notably, given the connection between \ac{ksd} and \ac{mmd} and the fact that in theory \ac{mmd} and \ac{ksd} have the same convergence rate of $O(n^{-1/2} + \nolinebreak m^{-1/2})\nolinebreak$, several Stein discrepancy-based methods outperform the \ac{mmd}-based method JAN.
In the SPA framework, adding Stein discrepancy boosts performance by several percentage points in the scarce target setting, although proposed methods in the plain framework outperform SPA in the scarce target setting.
The results in the SPA and FixMatch frameworks demonstrate that Stein discrepancy-based methods can offer an advantage in the scarce target setting even to recent, state-of-the-art models.

In the full target setting, Stein discrepancy-based methods are not very competitive.
The best performing method is MCC, followed by MDD and the Stein discrepancy-based methods perform at least 5\% worse than the best methods in the plain framework, and similarly for FixMatch and SPA frameworks.

\subsubsection{Office-Home}

\begin{figure}[h!]
    \centering
    \includegraphics[width=0.7\columnwidth]{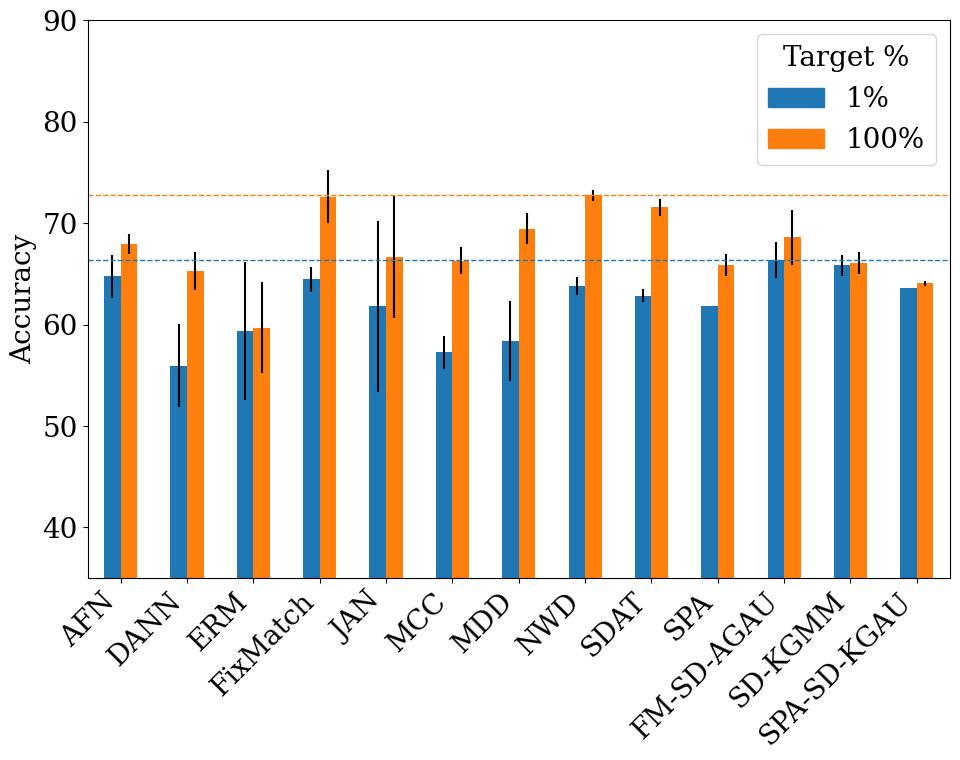}
    \caption{Accuracy on Office-Home, averaged across domain pairs.}
    \label{fig:oh-main}
\end{figure}

On Office-Home, the best-performing method in the full target setting was NWD, followed by FixMatch. 
The best performing method overall in the scarce target setting was FixMatch combined with SD-AGAU, which outperformed plain FixMatch by 2\%.
In the plain framework, SD-KGMM was the highest performing method, outperforming JAN, the \ac{mmd}-based method, by at least 4\%.
In the SPA framework, the KGAU method was the highest performing method, outperforming plain SPA by 2\%.

\subsubsection{VisDA-2017}

\begin{figure}[h!]
    \centering
    \includegraphics[width=0.7\columnwidth]{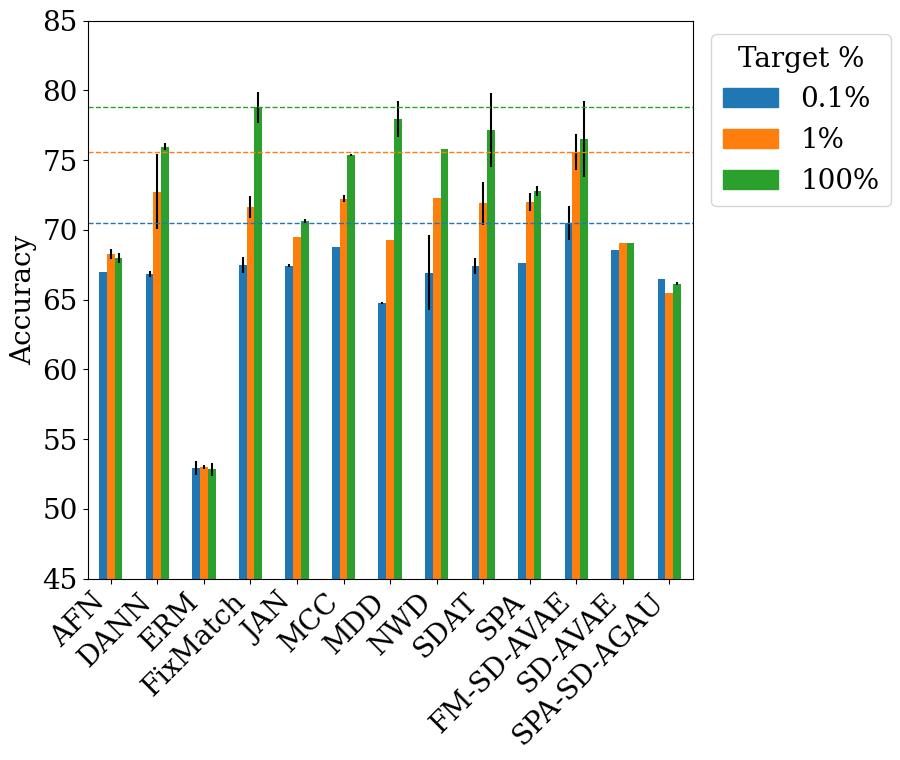}
    \caption{Accuracy on VisDA-2017.}
    \label{fig:vis-main}
\end{figure}

On VisDA-2017 in the full target setting, MDD was the highest performing method, followed by SDAT, both of which outperform Stein discrepancy-based methods in the plain framework by almost 10\%.
However, combining FixMatch with Stein discrepancy shows improvement even in the full target setting, with the best combined method achieving performance within 1\% of the best methods in the full target setting.

VisDA-2017 is the largest benchmark dataset used in these experiments, and 1\% of the target data is approximately 550 target samples.
In this setting, the best performing method overall was FixMatch combined with SD-AVAE.
In the plain framework, the best performing method was DANN, and the best performing Stein discrepancy-based method, SD-AVAE, was slightly outperformed by JAN.

It is likely that several hundred target data samples is too abundant to qualify as the scarce target setting, so we also examine the results with 0.1\% of target data available, which corresponds to approximately 55 target training samples.
In this setting the best method overall is still FixMatch combined with SD-AVAE.
In the plain framework, the best performing method  is MCC, closely followed by SD-AVAE.
In the SPA framework, combining with Stein discrepancy fails to show an advantage for this dataset, although Stein discrepancy-based methods in the plain framework outperform all methods in the SPA framework, including original SPA.

\subsubsection{Sensitivity to amount of target data}

To further explore the effect of the amount of available target data on \ac{uda} methods, we evaluate the methods on the Office31 dataset at the following levels of target data: $100\%$, $75\%$, $50\%$, $25\%$, $10\%$, $5\%$, $1\%$.
For the sake of readability, we display the same subset of methods as in Figure~\ref{fig:o31-main}: all baseline methods and the best Stein discrepancy-based method from each framework, SD-AGMM, FixMatch with SD-AGAU, and SPA with SD-KGMM.
We display the methods on the average across domains in Figure \ref{fig:target-pct-comparison}; results for each domain separately are included in Appendix \ref{app:add-experiments}.

\begin{figure*}[h]
    \centering
    \includegraphics[width=0.75\linewidth]{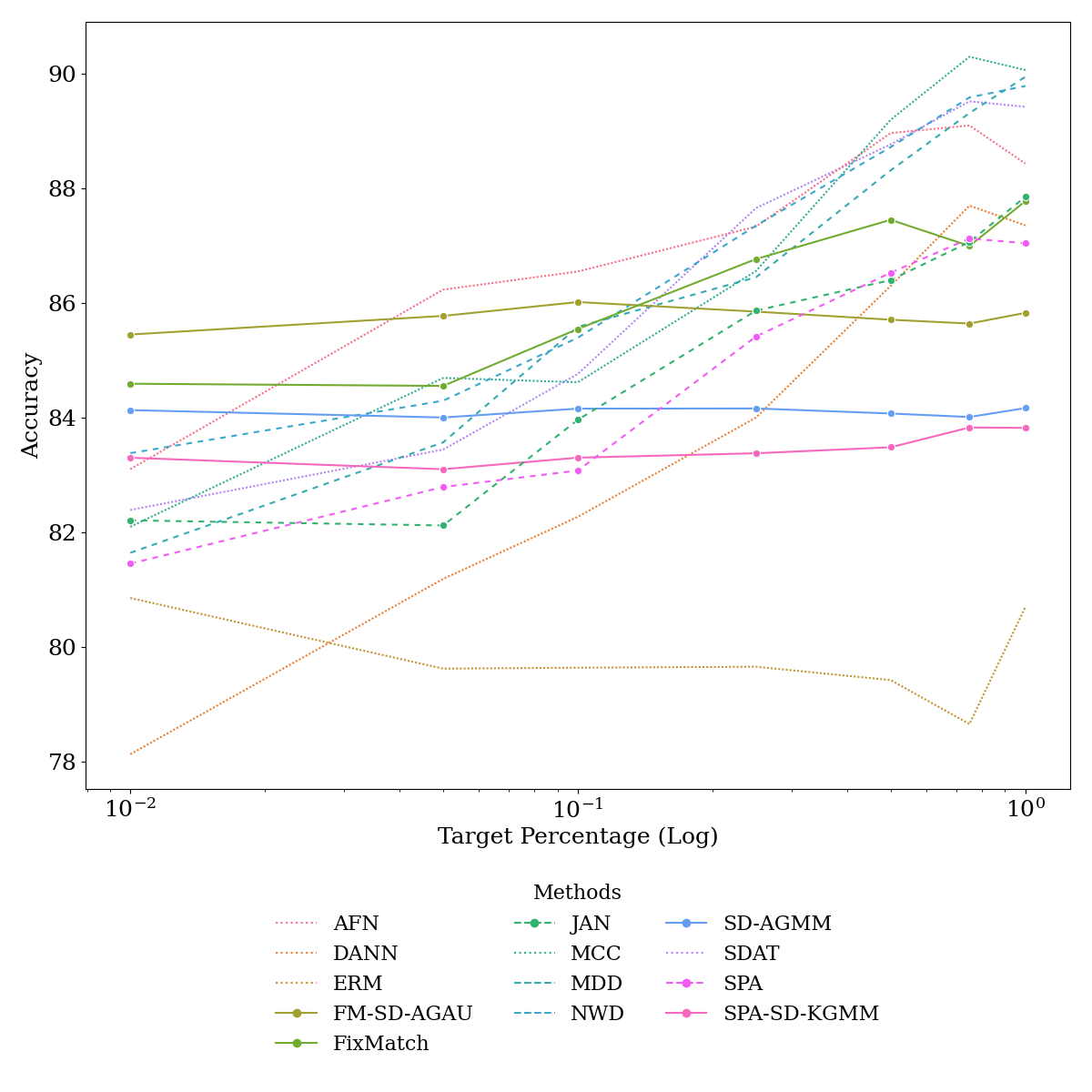}
    \caption{Accuracy vs. target data percentage on Office31 (log-scale). Stein discrepancy methods (solid lines) are more stable under data scarcity. FixMatch combined with SD-SD-AGMM performs best at low target availability.}
    \label{fig:target-pct-comparison}
\end{figure*}

Most methods see a decline in accuracy when the available data is below $10\%$.
The Stein discrepancy-based methods are the most stable to the change of percentages, and have a minimal decline in accuracy as the amount of target data decreases.
This suggests that Stein discrepancy-based methods have an advantage when target data is very scarce, which aligns with the results on the VisDA-2017 dataset, which showed little advantage for Stein discrepancy-based methods at the 1\% level but significant advantage at the 0.1\% level.

\subsubsection{Feature visualization}

We use t-SNE to visualize the learned features and provide an intuitive view of how well each method separates classes under different levels of target data.
Results are shown for the W2A domain pair from the Office-31 dataset, which is particularly challenging, as evidenced by the relatively low accuracy across all methods (see Table~\ref{table:office31-scarce} and \ref{table:office31-scarce}).
They include features trained on 100\% of the target data and 1\% of the target data for all methods except ERM, for which only 100\% of the target data is included, since the target data is not used in training ERM.

\begin{figure}[h!]
    \centering
        
            \includegraphics[width=0.65\columnwidth]{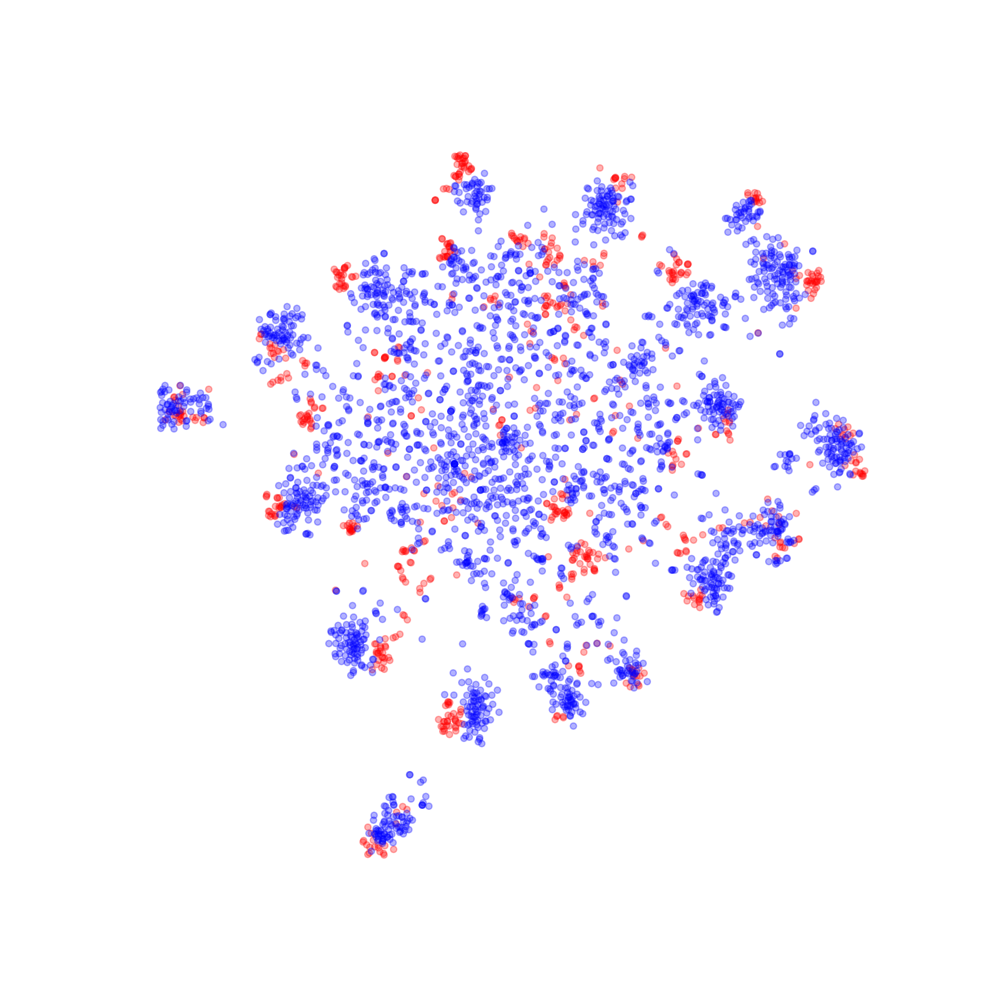}
            \caption{ERM: full target setting. We do not include the scarce target setting for ERM since it does not use any target data in training.}
            \label{fig:feature-vis-erm}

\end{figure}

\begin{figure}[h!]
    \centering
    \begin{tabular}{cc}
            \centering
            \includegraphics[width=0.45\columnwidth]{TSNE_W2A_targetPct_1.0.png}
            
       & 
            \centering
            \includegraphics[width=0.45\columnwidth]{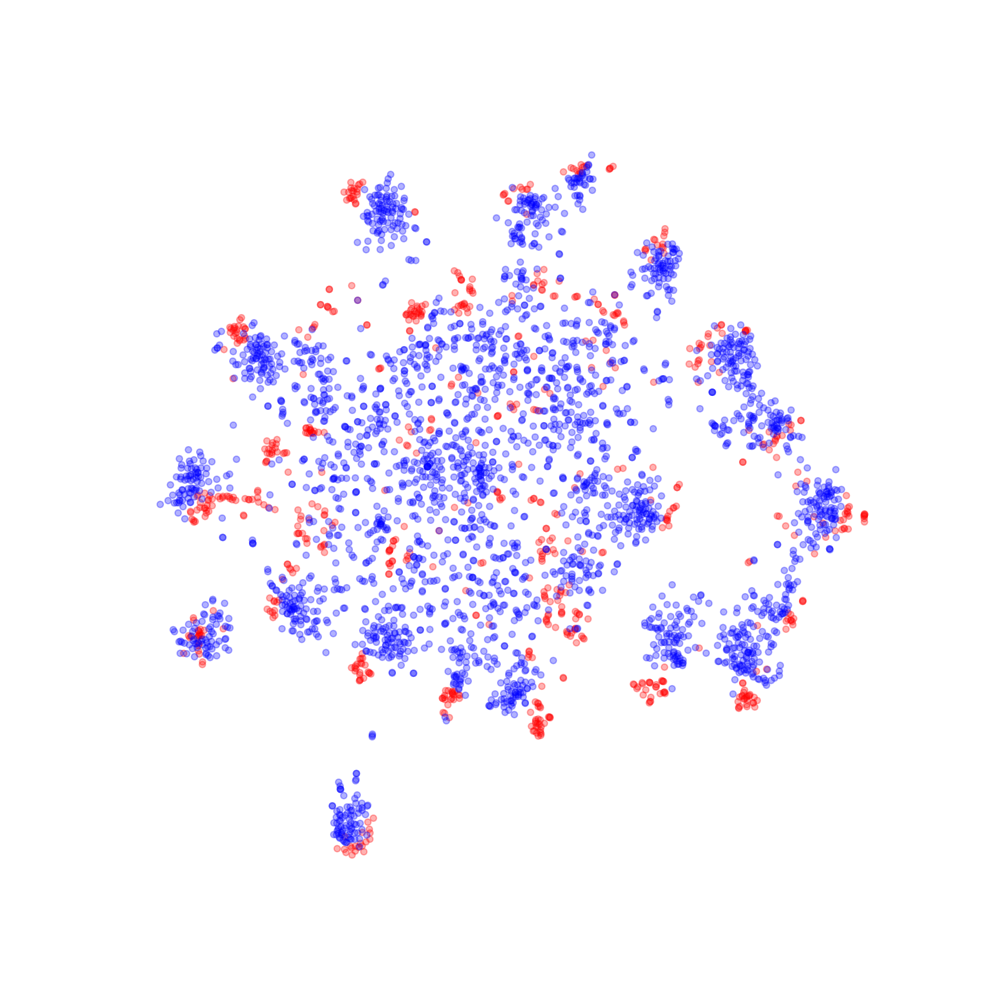}
    \end{tabular}
\label{fig:feature-vis-jan}
\caption{JAN: full target setting (left) and scarce target setting (right).}
\end{figure}

\begin{figure}[h!]
    \centering
    \begin{tabular}{cc}
            \centering
            \includegraphics[width=0.45\columnwidth]{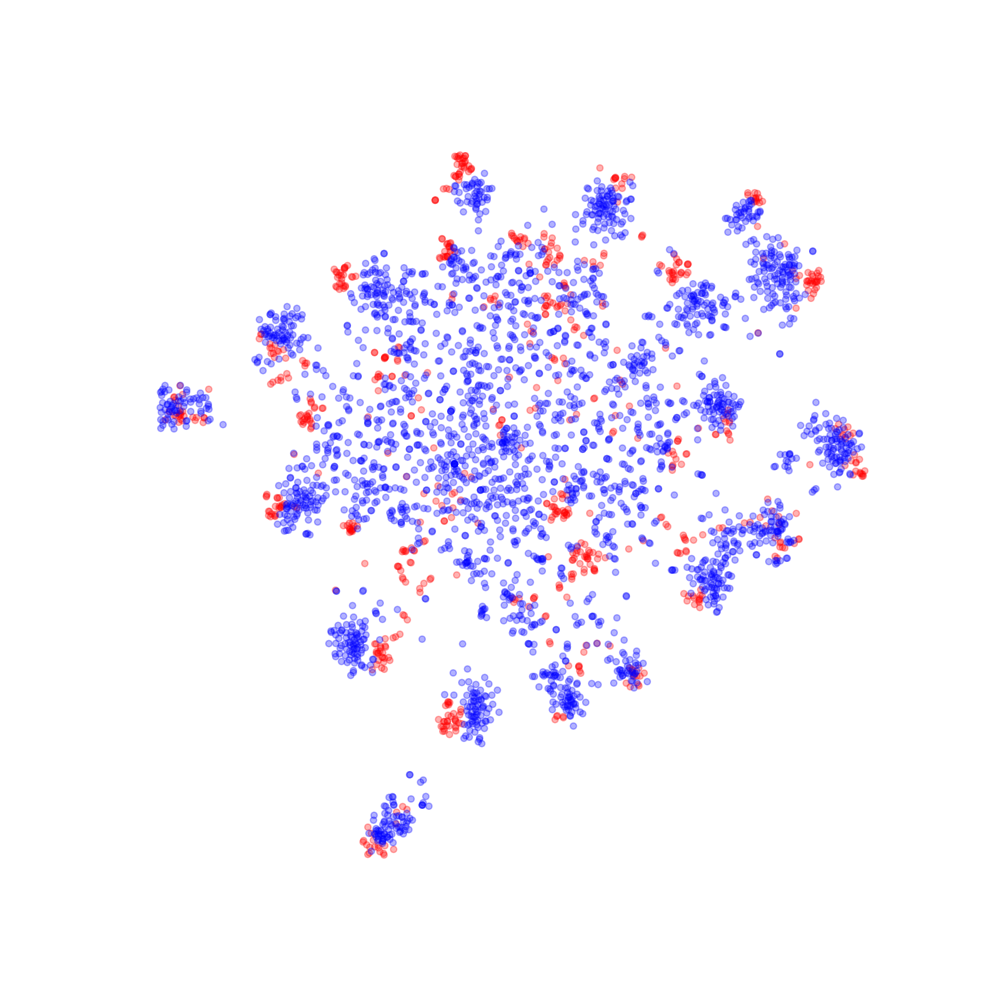}
       & 
            \centering
            \includegraphics[width=0.45\columnwidth]{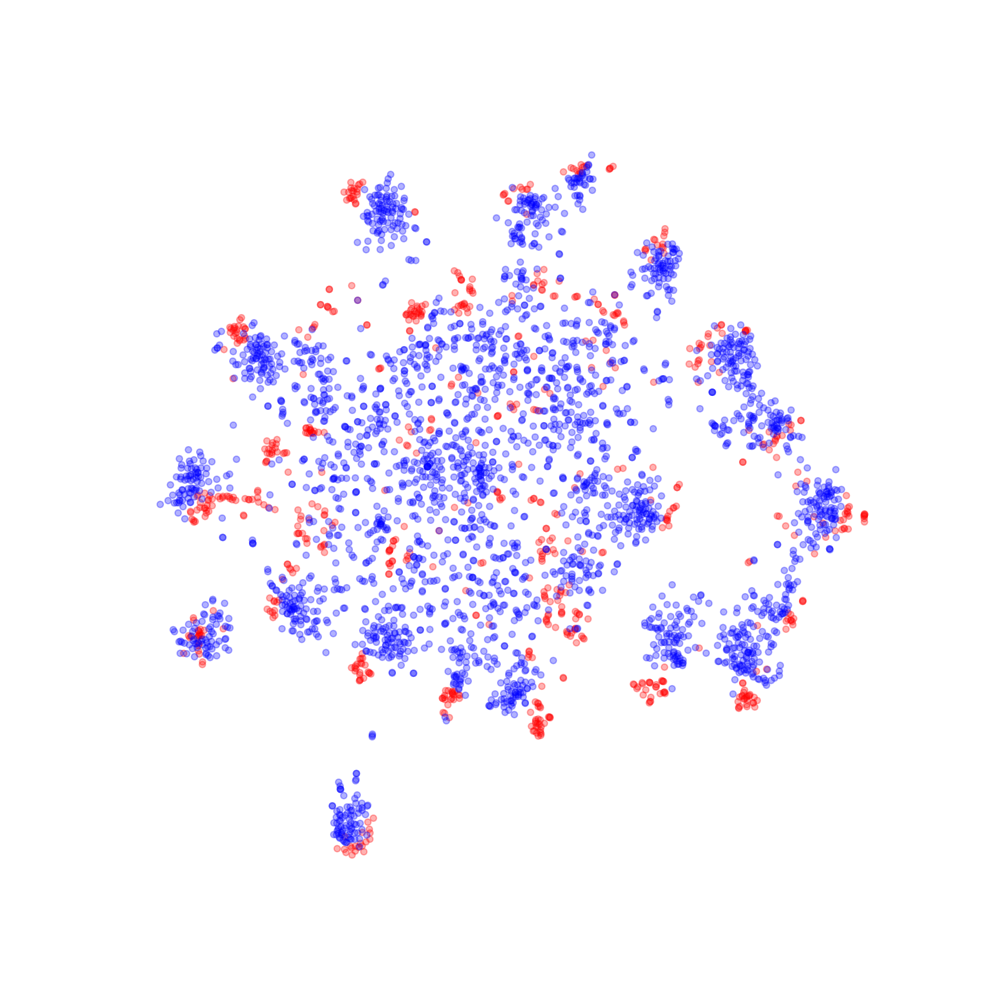}
            
    \end{tabular}
    \label{fig:feature-vis-sd-agmm}
\caption{SD-AGMM: full target setting (left) and scarce target setting (right).}
\end{figure}

\begin{figure}[h!]
    \centering
    \begin{tabular}{cc}
            \centering
            \includegraphics[width=0.45\columnwidth]{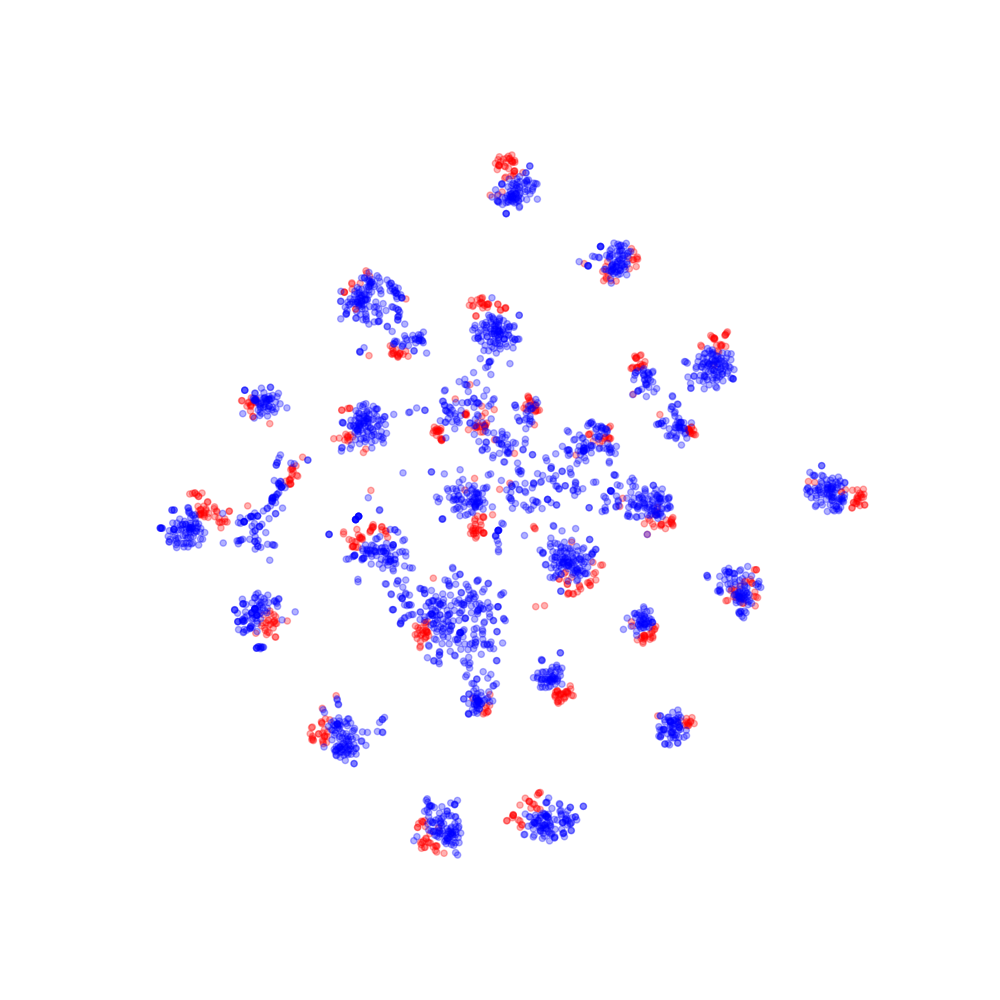}
            
       & 
            \centering
            \includegraphics[width=0.45\columnwidth]{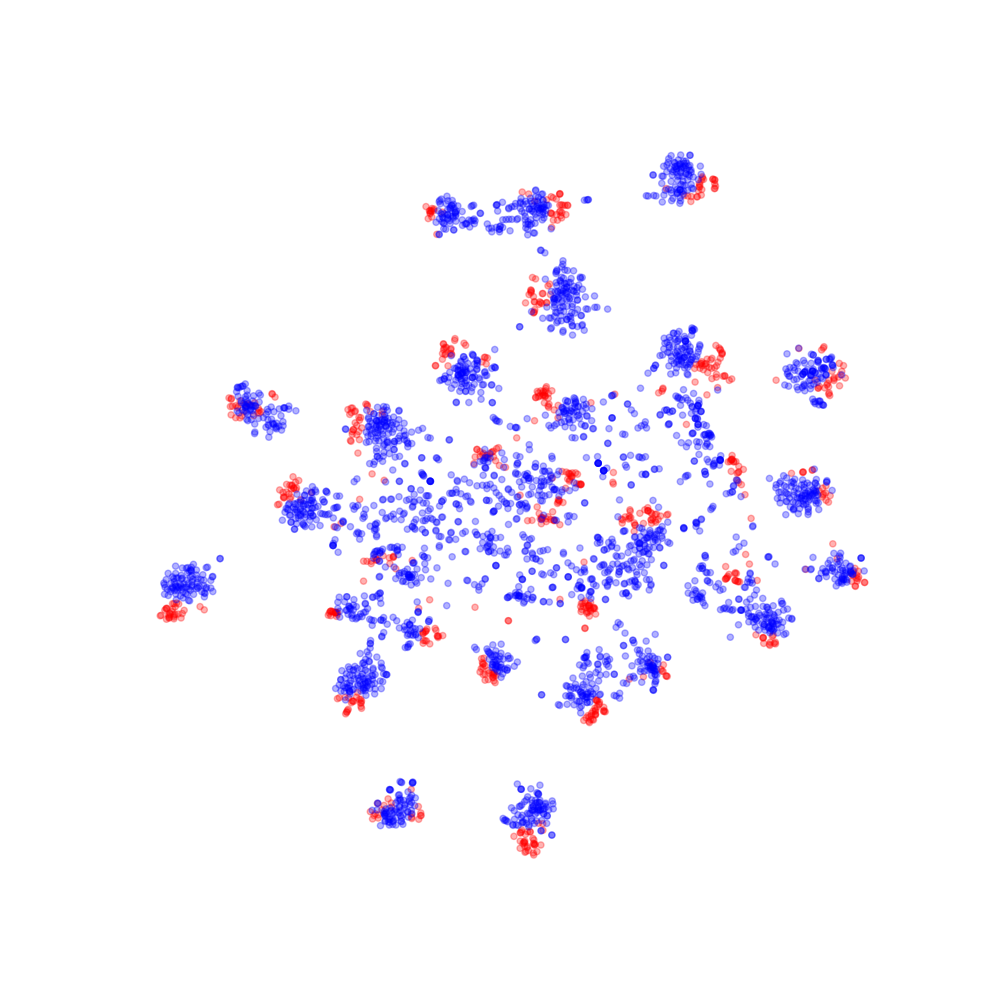}
    \end{tabular}
\label{fig:feature-vis-fixmatch}
\caption{FixMatch: full target setting (left) and scarce target setting (right).}
\end{figure}

\begin{figure}[h!]
    \centering
    \begin{tabular}{cc}
            \centering
            \includegraphics[width=0.45\columnwidth]{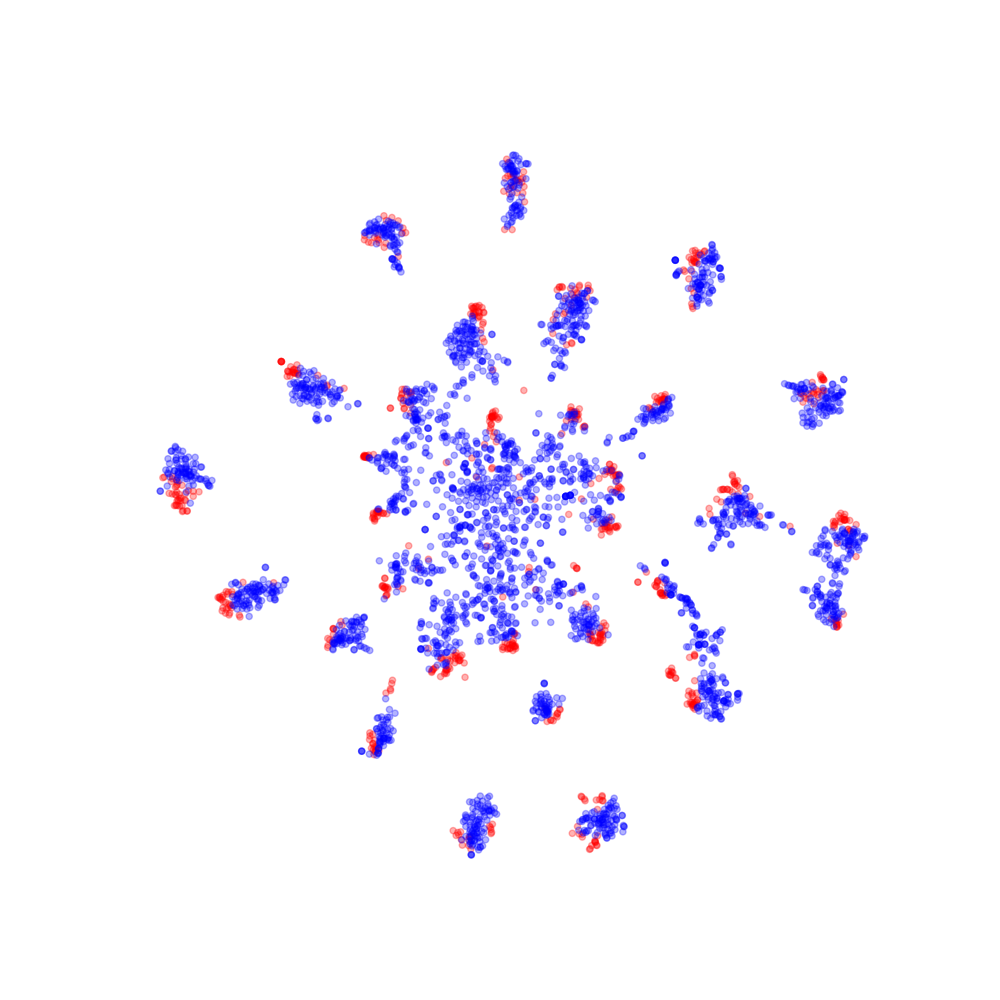}
        &
            \centering
            \includegraphics[width=0.45\columnwidth]{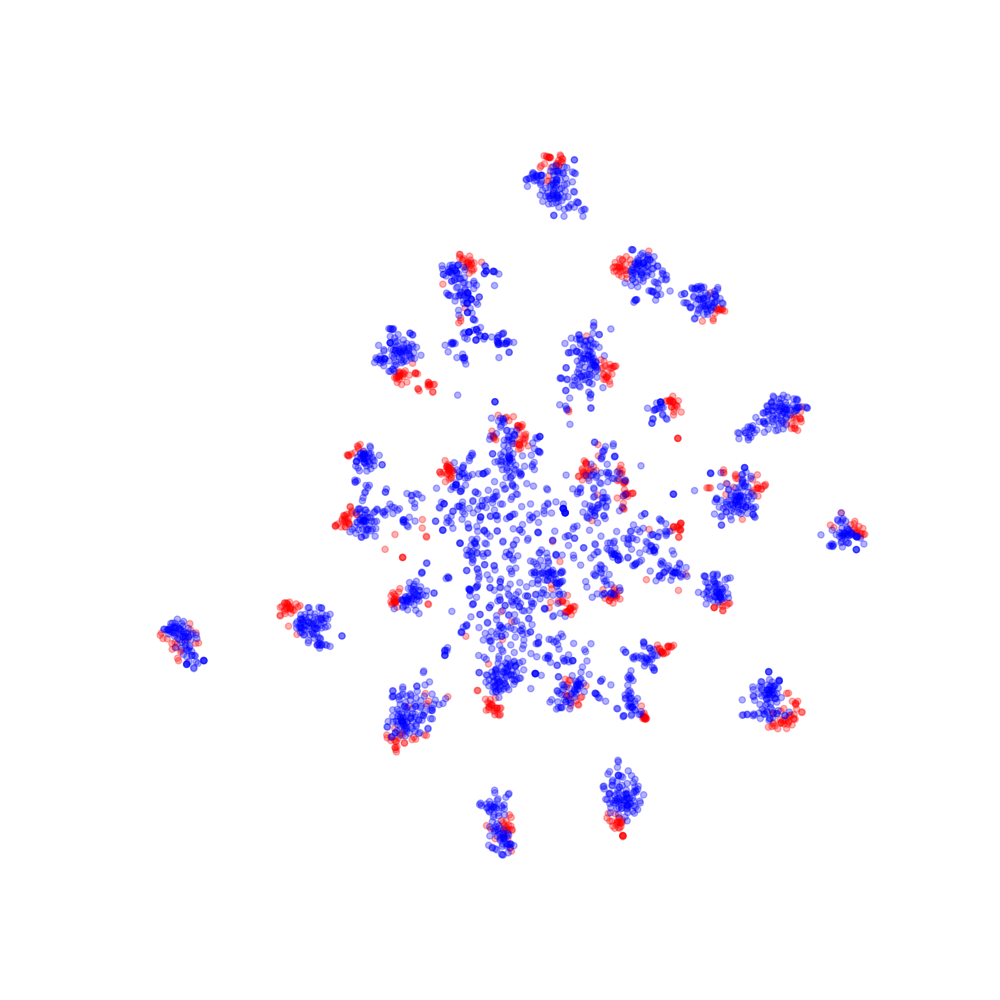}
            
    \end{tabular}
    \label{fig:feature-vis-fm-sd-agau}
\caption{FixMatch-SD-AGAU: full target setting (left) and scarce target setting (right).}
\end{figure}

\begin{figure}[h!]
    \centering
    \begin{tabular}{cc}
            \centering
            \includegraphics[width=0.45\columnwidth]{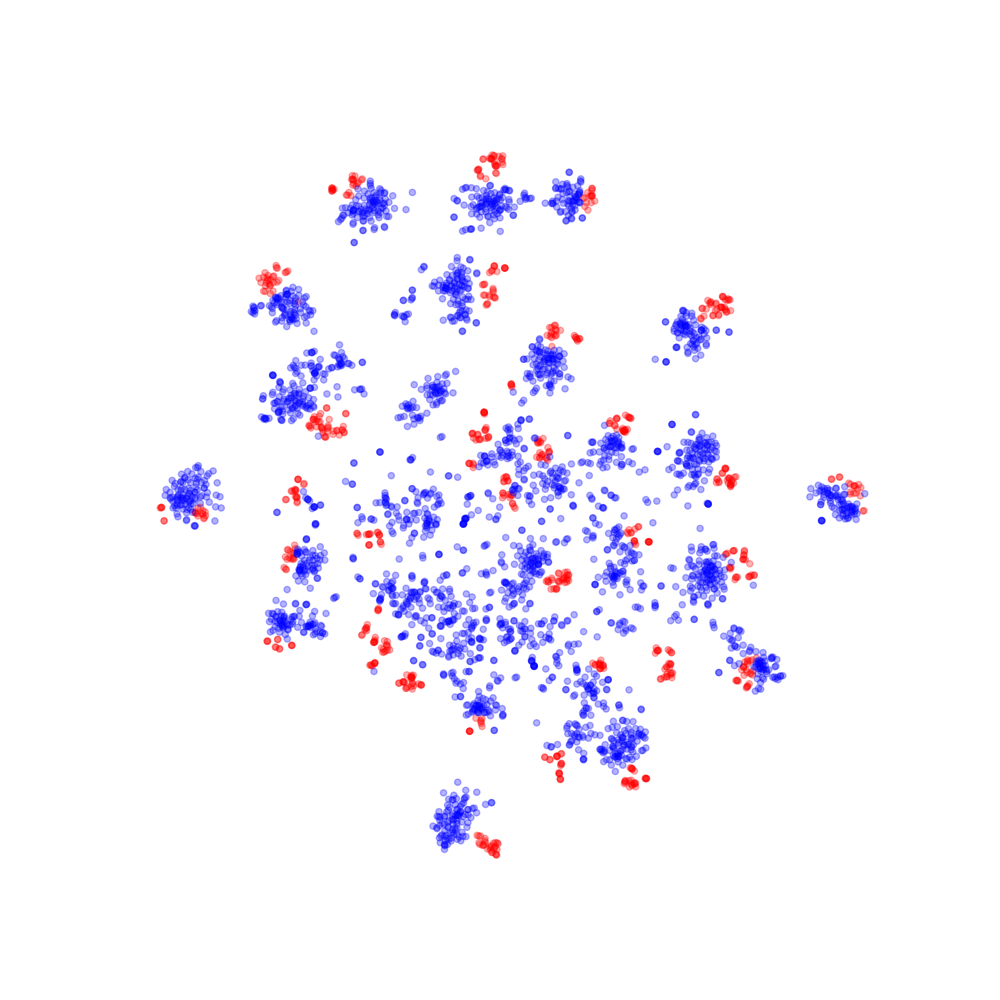}
 & 
            \centering
            \includegraphics[width=0.45\columnwidth]{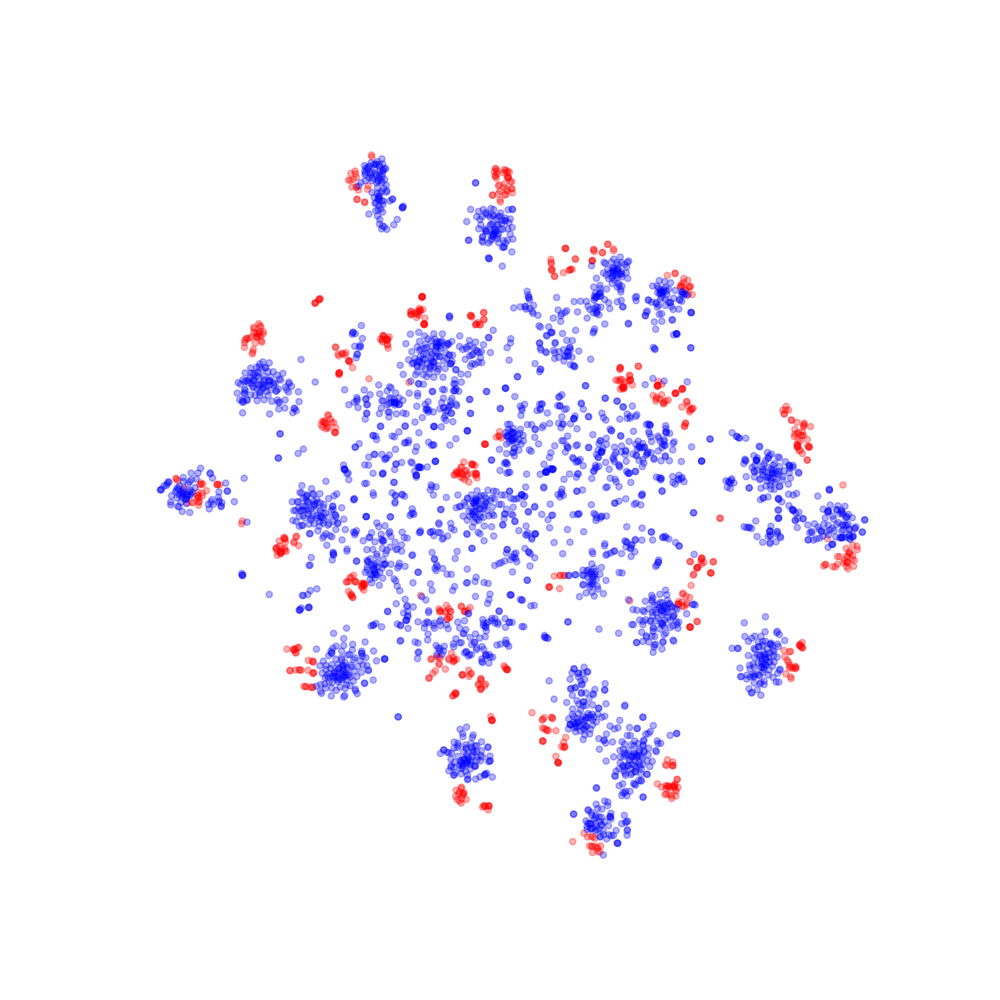}
    \end{tabular}
    \label{fig:feature-vis-spa}
            \caption{SPA: full target setting (left) and scarce target setting (right).}

\end{figure}

\begin{figure}[h!]
    \centering
    \begin{tabular}{cc}
            \centering
            \includegraphics[width=0.45\columnwidth]{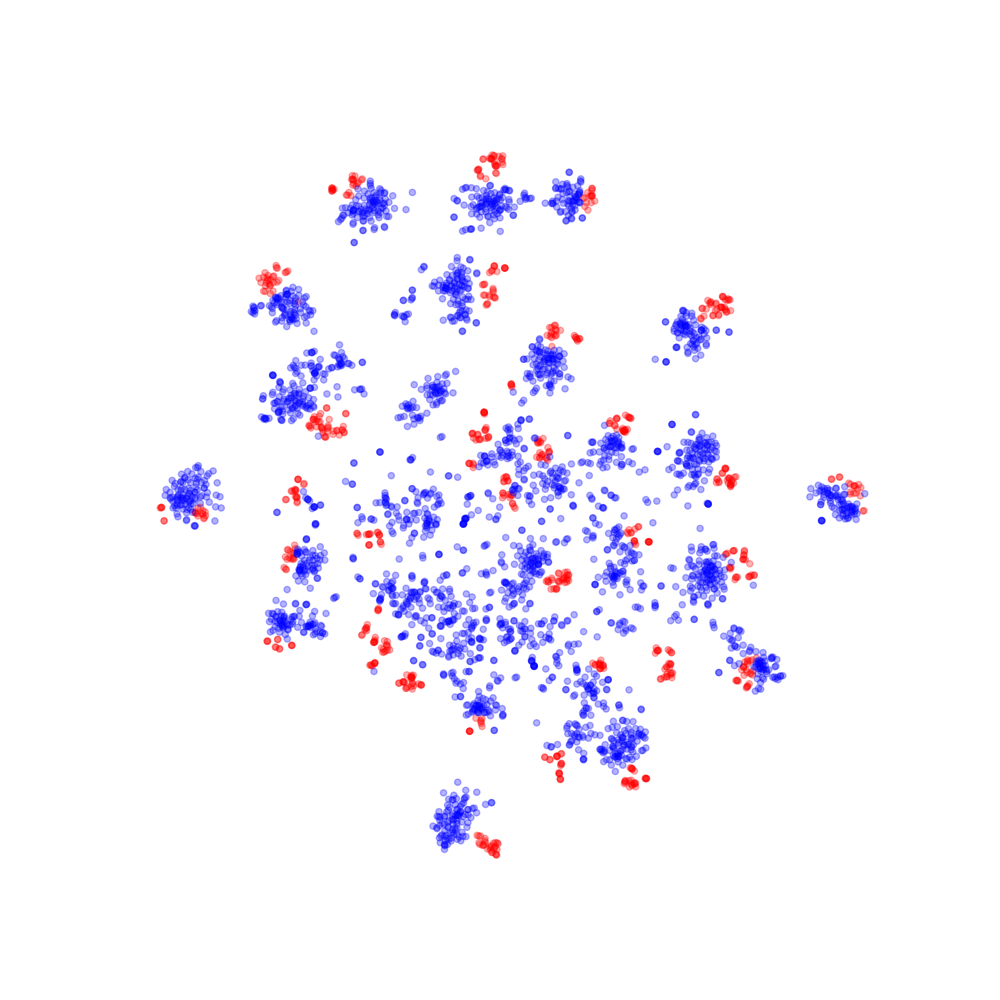}
        & 
            \centering
            \includegraphics[width=0.45\columnwidth]{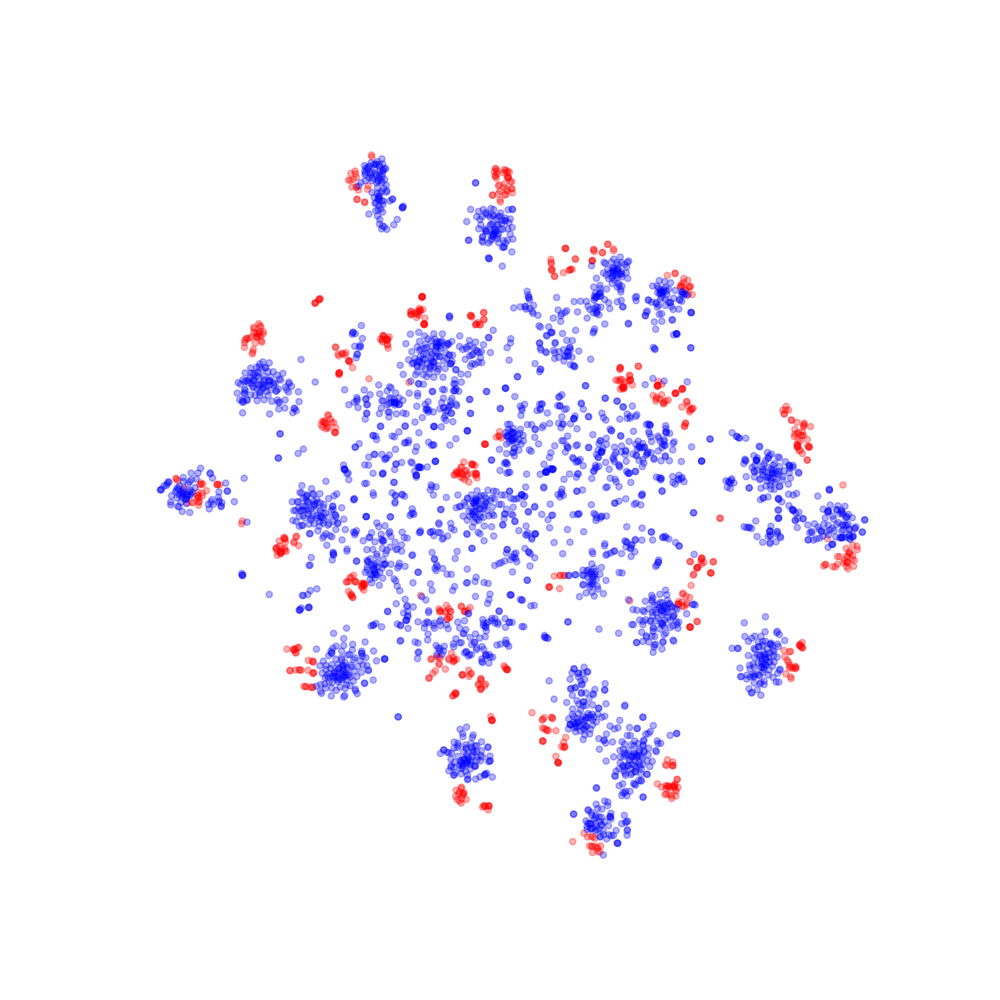}
            
    \end{tabular}
\caption{SPA-SD-KGAU: full target setting (left) and scarce target setting (right).}
    \label{fig:feature-vis-spa-sd-kgau}
\end{figure}

We compare four baseline methods (ERM, JAN, FixMatch, and SPA) with three Stein discrepancy-based methods (SD-AGMM, FM-SD-AGAU, and SPA-SD-KGAU), all of which are also included in Figure~\ref{fig:target-pct-comparison}.
The corresponding visualizations appear in Figures~\ref{fig:feature-vis-erm}–\ref{fig:feature-vis-spa-sd-kgau}.

All methods show improved class separation compared to ERM.
For all of the methods, the full target setting has better separation of classes compared to the scarce setting.
If we consider pairs of methods, JAN and SD-AGMM, FixMatch and FixMatch-SD-AGAU, and SPA and SPA-SD-KGAU, the method that does not include Stein discrepancy shows better separation between classes in the full target setting, while the method that includes Stein discrepancy generally shows better separation in the scarce target setting.
Fixmatch-SD-AGAU shows particularly good separation between classes in both settings, matching the accuracy results, in which FixMatch-SD-AGAU was the best overall performer on Office31.


\subsubsection{Regularization}

\begin{figure}[h!]
    \centering
    \includegraphics[width=0.7\columnwidth]{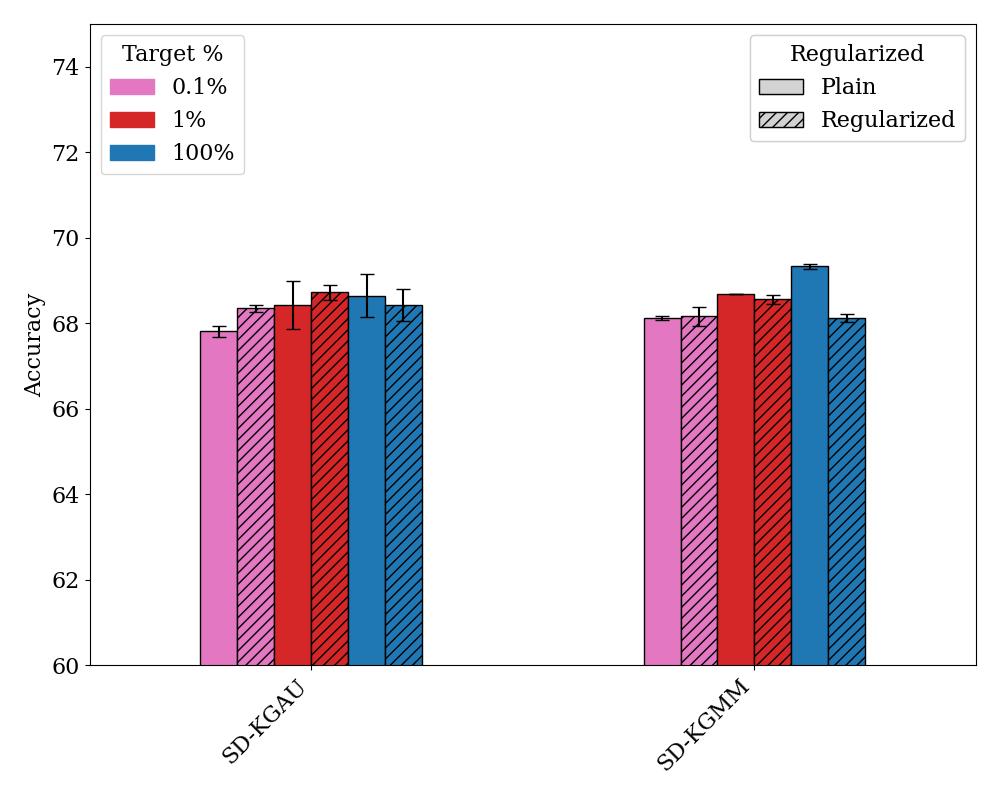}
    \caption{Accuracy on VisDA-2017 for regularized methods.}
    \label{fig:regularized}
\end{figure}

Finally, we compare \ac{ksd} with regularized \ac{ksd} \citep{hagrass2024minimax}, to investigate the intuition that the minimax optimal test statistic, regularized \ac{ksd}, will use scarce data more efficiently than the original \ac{ksd}.
The spectral regularization can only be applied to the kernelized method, so we compare only the SD-KGAU and SD-GMM methods on the VisDA-2017 dataset.
The regularized version with a Gaussian target distribution shows a slight improvement over unregularized \ac{ksd} when the amount of target data is 1\% and 0.1\%, but the unregularized version outperforms when 100\% of target data is used for training.
With a GMM target distribution, the regularized method is competitive with the unregularized method for smaller amounts of target data, 1\% and 0.1\%, but performs significantly worse when 100\% of target data is used for training.
The improved performance for the regularized method in the scarce data regime aligns with the intuition that the regularized version makes more efficient use of target data when it is extremely limited.
The small overall performance gain, particularly the lack of improvement with the GMM, suggest that in \ac{uda}, regularization offers only a marginal boost compared to other factors, such as source domain classification accuracy and the accuracy of target distribution parameter estimates.


\section{Discussion and Conclusion}\label{sec:discussion}

Overall, the results demonstrate that Stein discrepancy-based methods consistently outperform baseline methods in low-target-data regimes across multiple datasets.
While their advantage is less pronounced when thousands of unlabeled target examples are available, they show strong robustness and stability as target data becomes scarce, particularly with 100 or fewer target data samples.
Among the variants of the proposed method in the plain framework, SD-KGAU and SD-AGMM perform consistently well across all datasets.
Combining Stein discrepancy with other \ac{uda} methods also offers improved performance in the scarce target setting over both original methods.
SD-AGAU combined with FixMatch offers a particularly strong combination across multiple datasets and achieved the highest overall accuracy on several datasets, making it a strong default choice for new domain adaptation tasks when the amount of target data is limited (e.g. 100 examples or fewer).
Finally, the results on VisDA-2017 suggest that on larger, more complicated datasets, more flexible target distributions, such as VAE, have an advantage over a Gaussian target distribution.
These findings not only demonstrate strong empirical performance but also complement the theoretical motivation for using Stein discrepancy in \ac{uda}, reinforcing its value in settings with limited target supervision.

The results on VisDA-2017 also help to clarify when the scarce target assumption applies and Stein discrepancy can offer an advantage for \ac{uda}.
With at least several hundred target examples, \ac{ksd} offers limited gains over state-of-the-art \ac{uda} methods.
However, with less than 100 samples, Stein discrepancy-based methods outperform all others, indicating the data regimes where \ac{ksd}-based methods are most effective.

A practical limitation of Stein discrepancy-based methods is that their performance saturates as more unlabeled target data becomes available.
While they are effective in low-data regimes, they offer limited improvements in settings with abundant target data.
Another practical limitation is limited compatibility with other \ac{uda} approaches.
It complements methods that focus on aspects of adaptation other than feature alignment, such as self-training methods like FixMatch \citep{sohn2020fixmatch} or optimization-focused methods such as SDAT \citep{rangwani2022closer}, but it cannot be directly combined with  feature alignment methods whose main contribution is also a domain discrepancy measure, such as JAN \citep{long_learning_2015} and NWD \citep{chen2022reusing}.
Finally, the empirical validation was conducted on a standard set of domain adaptation benchmarks, but the datasets are relatively limited in diversity and size.
As a result, further testing is needed to evaluate the robustness and generalizability of the method to real-world conditions.

From a theoretical standpoint, we provide a generalization bound on the target error, verifying that Stein discrepancy is an effective measure of distance between distributions of \ac{uda}, and a convergence rate of the empirical estimate of \ac{ksd} to the true value when the score function is estimated.
The convergence rate, $O(n^{-1/2} + m^{-1/2})$ is identical between \ac{ksd} and \ac{mmd}; the better empirical performance of \ac{ksd} with unbalanced sample sizes is not currently supported by theory.
The method assumes a smooth, differentiable target density, and the convergence result in our setting further assumes that the target distribution has an M-estimator and satisfies certain regularity conditions.
Verifying that the GMM and VAE target distributions follow these assumptions is left for future work.
Performance also depends on the choice of Stein operator and target approximation, and there is limited guidance for selecting between the kernelized and adversarial variants.
Future work should focus on when \ac{ksd} outperforms adversarial Stein discrepancy, and when recent extensions of \ac{ksd} offer an advantage.
We began this work by analyzing regularized \ac{ksd} \citep{hagrass2024minimax}, but other extensions, particularly non-parametric KSD \citep{xu_kernelised_2022}, deserve attention and evaluation for their applications to \ac{uda}.

Finally, our method is designed for scenarios where labeled target data is limited.
This has the potential to improve access to high-performing models in low-resource settings.
As with other domain adaptation techniques, care should be taken to ensure that performance gains do not mask bias or inaccuracies due to distribution shift, especially when applied in sensitive domains such as medicine.

We have proposed a novel method for \ac{uda} based on Stein discrepancy, and derived two theoretical results: a generalization bound that motivates minimizing the Stein discrepancy, and a convergence rate for empirical \ac{ksd} in two-sample testing.
The proposed method is adaptable and has both a kernelized form and a non-kernelized, adversarial form, with several possible parametric models for the target distribution: Gaussian, GMM, or VAE.
In numerical experiments, our method outperformed baseline methods in the scarce target setting, where only a small amount of target data is available.

\backmatter

\bmhead{Acknowledgements}

We thank Lester Mackey for his insightful comments on the historical development of Stein discrepancies, and Krishnakumar Balasubramanian for valuable discussion of regularized \ac{ksd} and sharing code. 
A.V. and G.L. thank Larry Goldstein for his inspiring talk and discussion on Stein discrepancy and kernels. 
A.V. and G.L. were partially supported by NSF award DMS 2427955.
D.Z. was partially supported by National Natural Science Foundation of China (NSFC) award 12301117.

\section*{Declarations}

\bmhead{Competing Interests} The authors declare no competing interests.

\section*{Supplementary Material} The supplementary material contains three appendices.
The first appendix includes additional experimental results, including tables reporting results on individual domain pairs for Office31 and Office-Home, where results in the main paper are averaged across domain pairs.
The second appendix contains information about experimental implementation and reproducibility, and the third appendix contains information about our use of existing code and datasets in numerical experiments.
Code is available at \href{https://github.com/amvs/stein-disc}{Github}.

\bibliography{DomainAdaptationStein}

\clearpage

\begin{appendices}
\appendix
{
\begin{center}
\Large{\textbf{Supplementary Material}}
\end{center}
}

\noindent Appendix~\ref{app:add-experiments} includes additional experimental results, including tables reporting results on individual domain pairs for Office31 and Office-Home, where results in the main paper are averaged across domain pairs.
Appendix~\ref{app:implementation} contains information about experimental implementation and reproducibility.
Appendix~\ref{app:existing-assets} describes our use of existing code and datasets in numerical experiments.
Code is available at \href{https://github.com/amvs/stein-disc}{Github}.
All references that do not start with A refer to the numbering in the main paper. Similarly, we refer to the bibliography from the main text. 


\section{Additional experimental results}\label{app:add-experiments}

\subsection{Accuracy}


To supplement the results in the main paper, we provide tables of results, including accuracy on each domain pair for Office31 and OfficeHome datasets, since the results of the main paper are averaged across domain pairs.
Results on Office-31 are displayed in Tables \ref{table:office31-full} and \ref{table:office31-scarce}.
Results on Office-Home are displayed in Tables \ref{table:officehome-mean}, \ref{table:oh-100pct}, and \ref{table:oh-1pct}.
Results on VisDA-2017 are displayed in Table \ref{table:visda}.

\begin{figure}[h!]
    \centering
    \includegraphics[width=0.7\columnwidth]{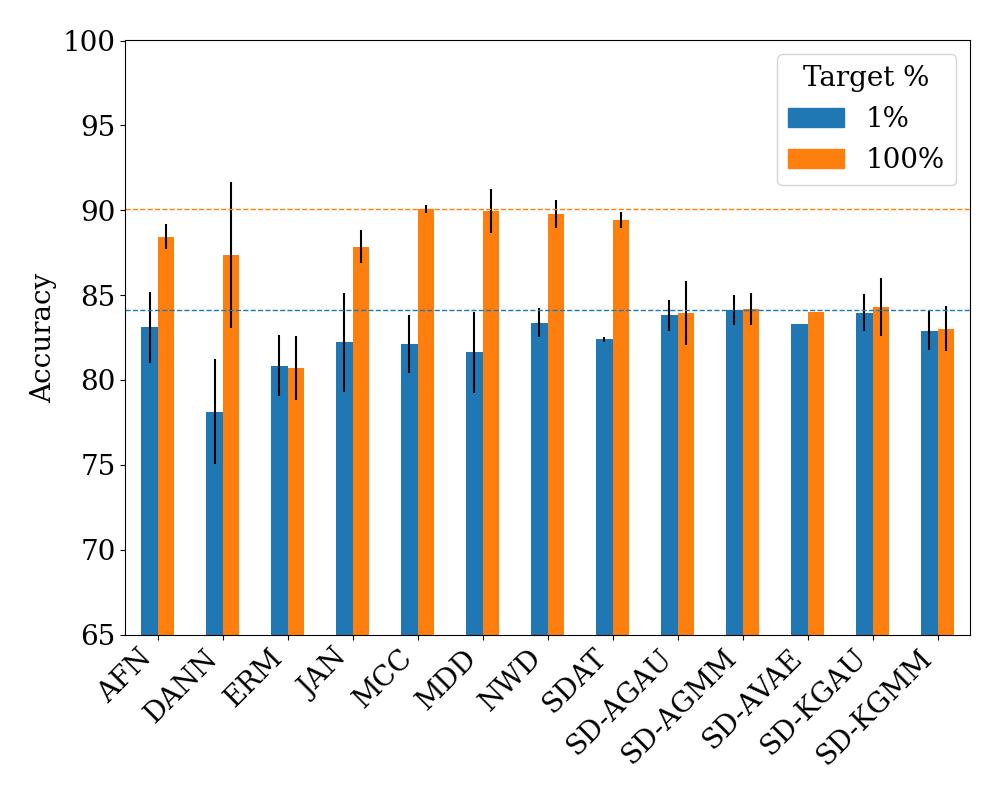}
    \caption{Accuracy on Office31, averaged across six domain pairs. Orange bars use all target data; blue bars use at most 1\% (or 32) target examples.}
    \label{fig:o31-plain}
\end{figure}

\begin{figure}[h!]
    \centering
    \includegraphics[width=0.7\columnwidth]{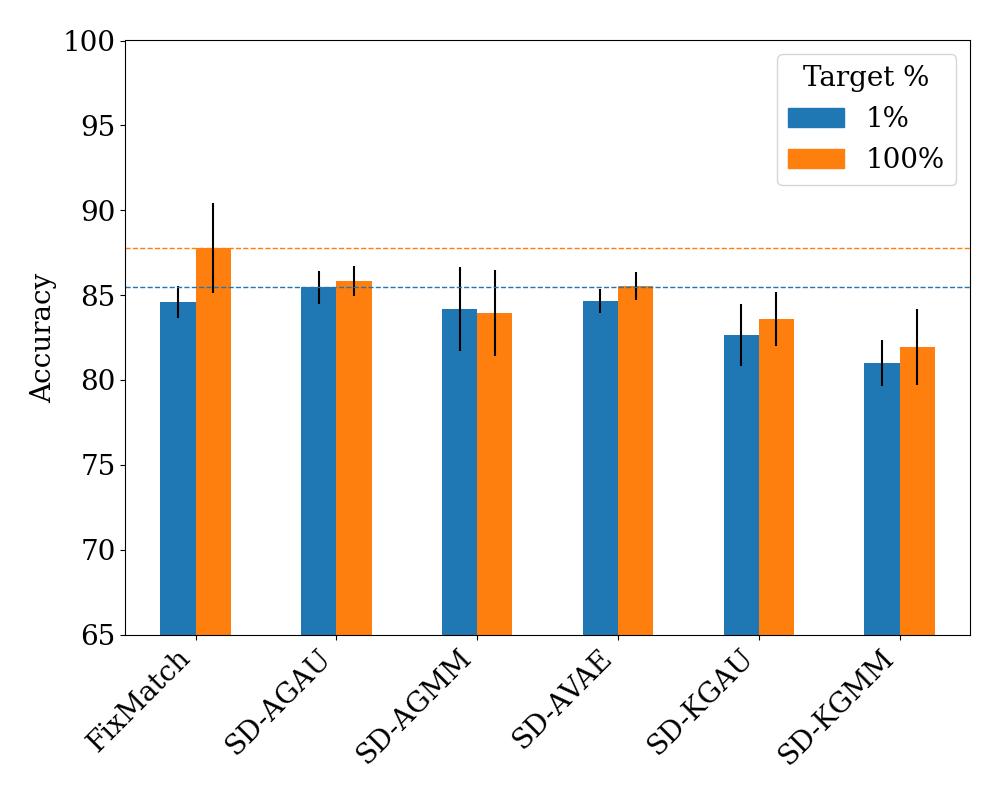}
    \caption{Accuracy on Office31, averaged over six domain pairs, for FixMatch-based methods. Unlike Figure~\ref{fig:o31-plain}, all methods here include FixMatch. “SD–” methods combine Stein discrepancy with FixMatch; “FixMatch” alone is the baseline. Orange bars use all target data; blue bars use at most 1\% (or 32) target examples.}
    \label{fig:o31-fm}
\end{figure}

\begin{figure}[h!]
    \centering
    \includegraphics[width=0.7\columnwidth]{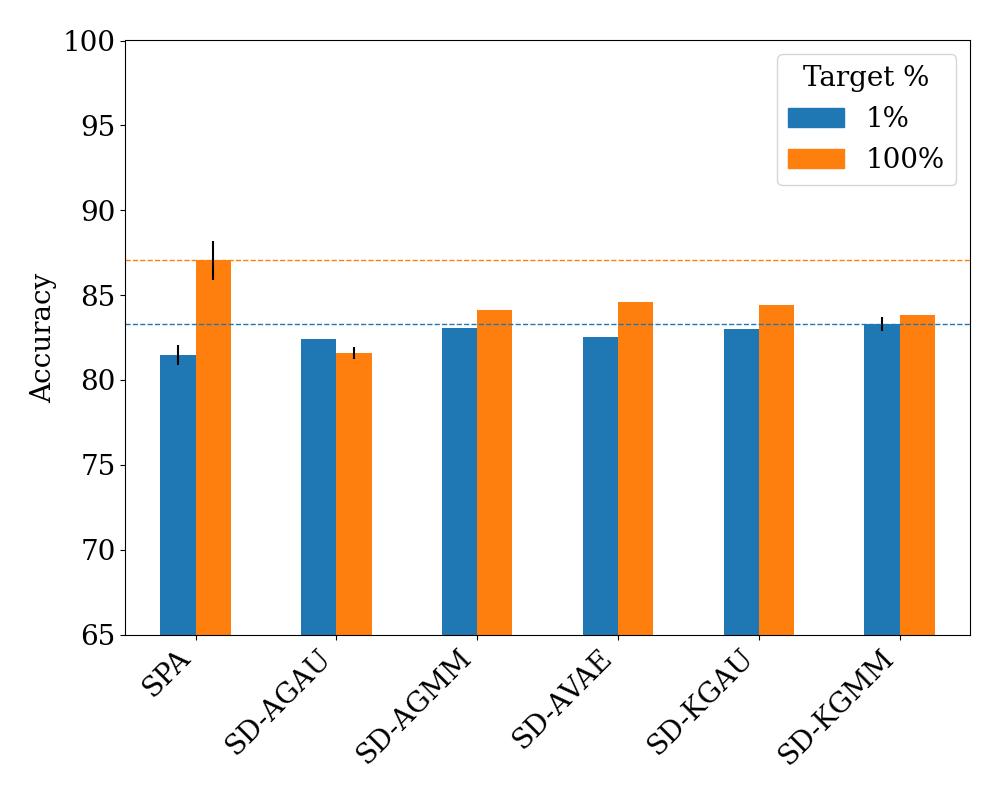}
    \caption{Accuracy on Office31, averaged over six domain pairs, for SPA-based methods. Unlike Figure~\ref{fig:o31-plain}, all methods here include SPA. “SD–” methods combine Stein discrepancy with SPA; “SPA” alone is the baseline. Orange bars use all target data; blue bars use at most 1\% (or 32) target examples.}
    \label{fig:o31-spa}
\end{figure}

\begin{figure}[h!]
    \centering
    \includegraphics[width=0.7\columnwidth]{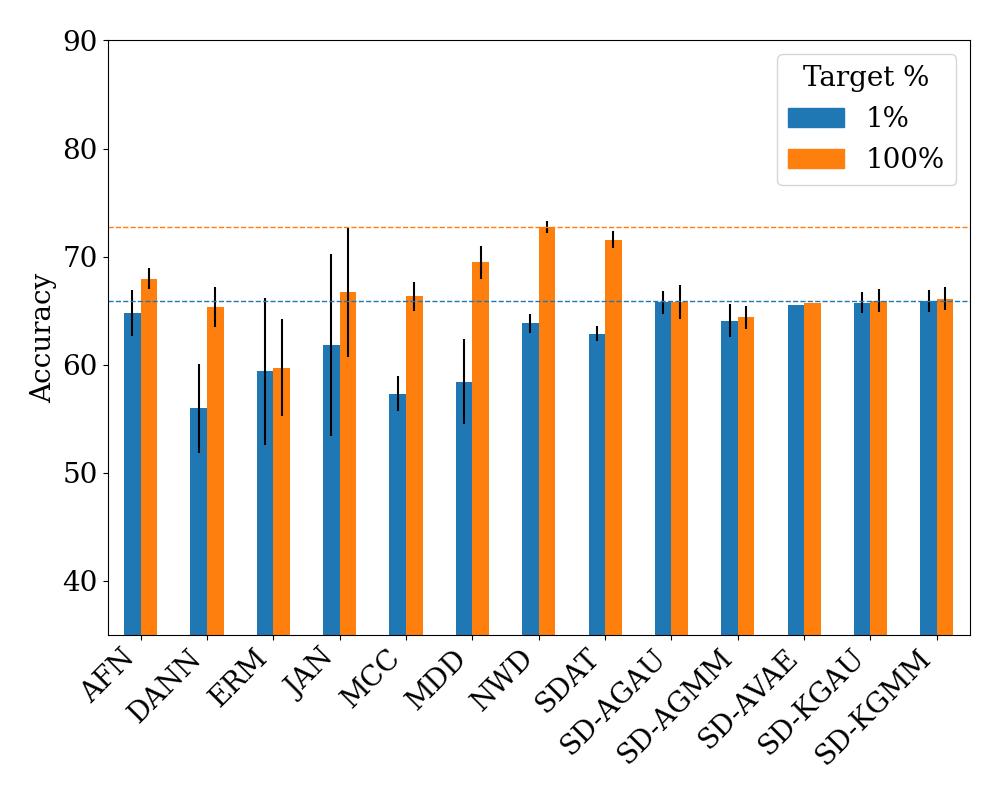}
    \caption{Accuracy on Office-Home, averaged across domain pairs.}
    \label{fig:oh-plain}
\end{figure}

\begin{figure}[h!]
    \centering
    \includegraphics[width=0.7\columnwidth]{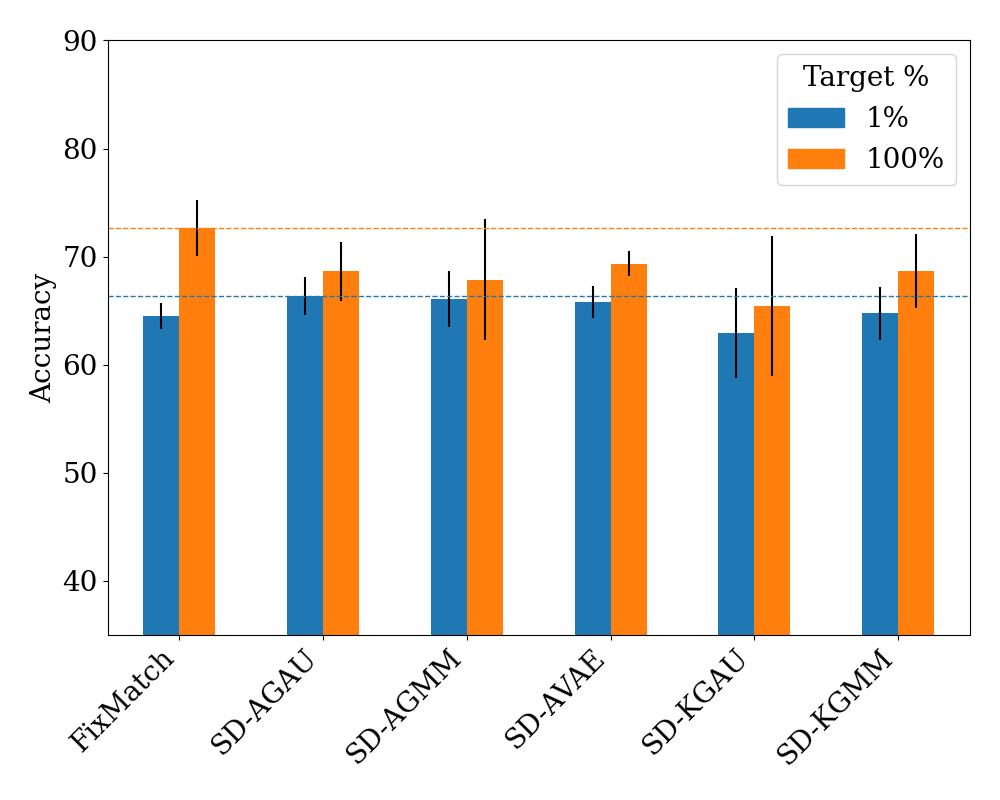}
    \caption{Accuracy on Office-Home, averaged across domain pairs. Unlike Figure~\ref{fig:oh-plain}, all methods here include FixMatch. “SD–” methods combine Stein discrepancy with FixMatch; “FixMatch” alone is the baseline.}
    \label{fig:oh-fm}
\end{figure}

\begin{figure}[h!]
    \centering
    \includegraphics[width=0.7\columnwidth]{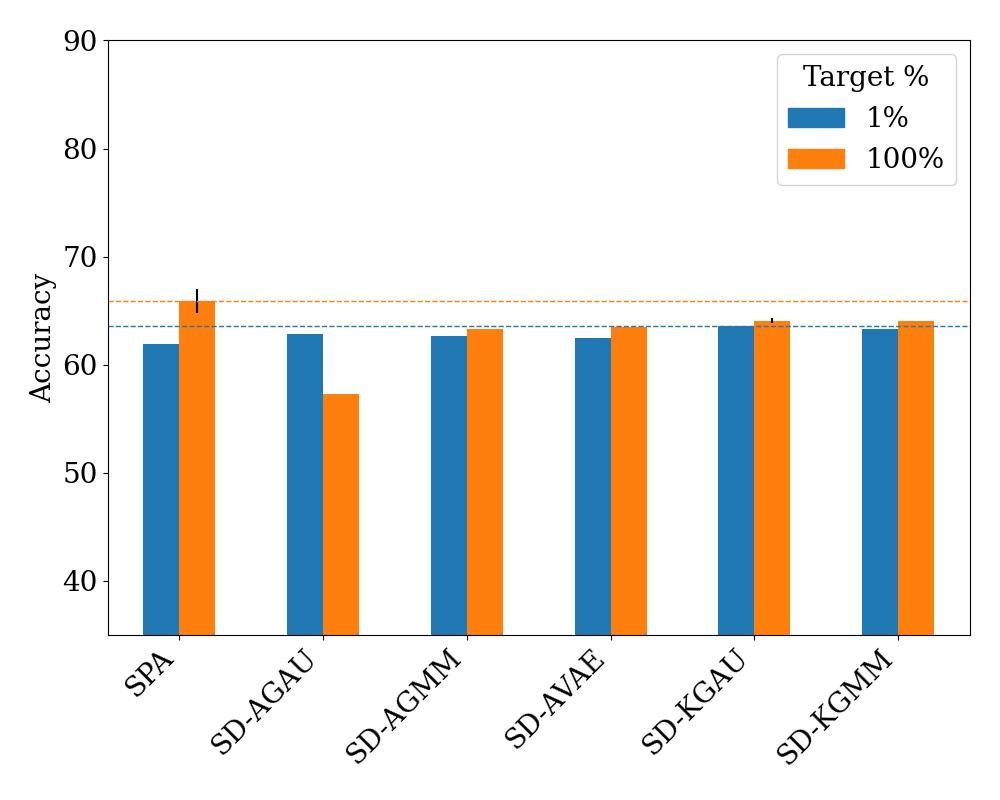}
    \caption{Accuracy on Office-Home, averaged across domain pairs. Unlike Figure~\ref{fig:oh-plain}, all methods here include SPA. “SD–” methods combine Stein discrepancy with SPA; “SPA” alone is the baseline.}
    \label{fig:oh-spa}
\end{figure}

\begin{figure}[h!]
    \centering
    \includegraphics[width=0.7\columnwidth]{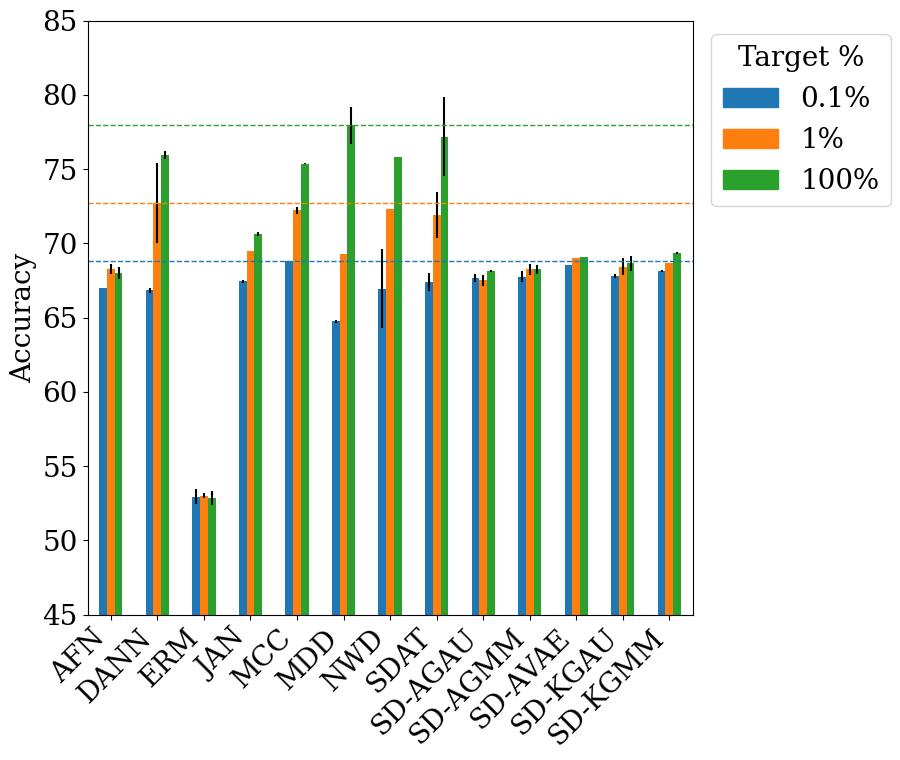}
    \caption{Accuracy on VisDA-2017.}
    \label{fig:vis-plain}
\end{figure}

\begin{figure}[h!]
    \centering
    \includegraphics[width=0.7\columnwidth]{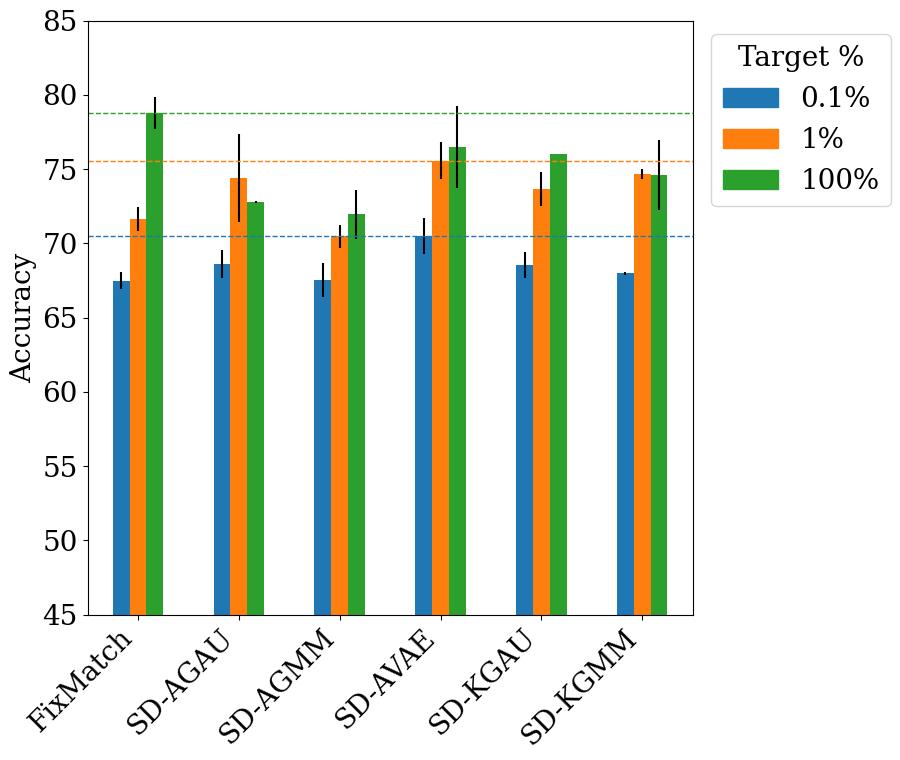}
    \caption{Accuracy on VisDA-2017. Unlike Figure~\ref{fig:vis-plain}, all methods here include FixMatch. “SD–” methods combine Stein discrepancy with FixMatch; “FixMatch” alone is the baseline.}
    \label{fig:vis-fm}
\end{figure}

\begin{figure}[h!]
    \centering
    \includegraphics[width=0.7\columnwidth]{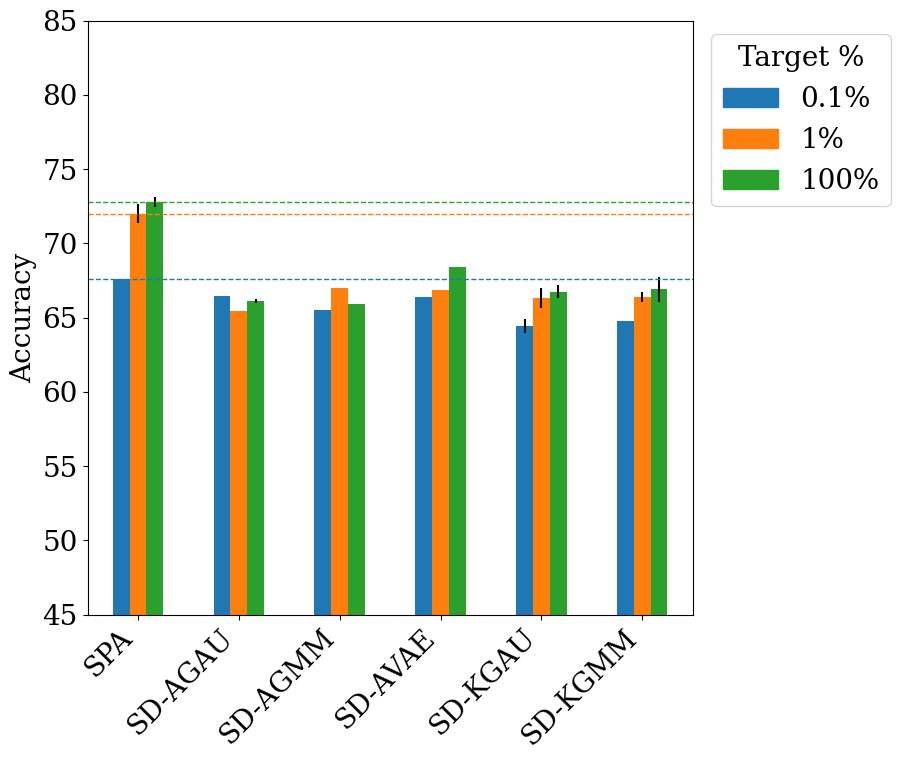}
    \caption{Accuracy on VisDA-2017. Unlike Figure~\ref{fig:vis-plain}, all methods here include SPA. “SD–” methods combine Stein discrepancy with SPA; “SPA” alone is the baseline.}
    \label{fig:vis-spa}
\end{figure}

We also include plots of all of the methods, averaged across domain pairs, to supplement the results in Figure~\ref{fig:o31-main}-\ref{fig:vis-main}.
Results are grouped by dataset and by framework and are displayed in Figures~\ref{fig:o31-plain}-\ref{fig:vis-spa}.

\begin{table*}[ht!]
    \centering
    \scriptsize
    \caption{Accuracy and standard deviation on Office31 in the full target setting (100\% of target data available). The best accuracy is bolded and the second best is underlined.}
    \label{table:office31-full}
    \resizebox{\textwidth}{!}{
\begin{tabular}{llllllll}
\toprule
Method & A2D & A2W & D2A & D2W & W2A & W2D & Mean \\
\midrule
AFN & \underline{95.1 (0.23)} & 92.5 (0.38) & 73.0 (0.57) & 98.9 (0.06) & 71.0 (0.06) & 100.0 (0.00) & 88.4 (0.73) \\
DANN & 85.2 (2.55) & 92.0 (2.25) & 73.6 (0.85) & 97.7 (0.50) & \underline{75.7 (2.43)} & 99.9 (0.12) & 87.4 (4.29) \\
ERM & 81.8 (0.31) & 77.3 (0.51) & 65.1 (1.76) & 96.6 (0.10) & 64.3 (0.15) & 99.2 (0.00) & 80.7 (1.87) \\
FixMatch & 93.0 (2.48) & 89.4 (0.67) & 72.9 (0.53) & 98.3 (0.07) & 73.1 (0.21) & 100.0 (0.00) & 87.8 (2.63) \\
JAN & 90.4 (0.35) & \underline{94.6 (0.45)} & 71.5 (0.69) & 98.2 (0.15) & 72.4 (0.32) & 100.0 (0.00) & 87.9 (0.96) \\
MCC & \textbf{97.4 (0.00)} & 94.1 (0.23) & 75.6 (0.00) & 98.4 (0.06) & 75.1 (0.00) & 99.8 (0.00) & \textbf{90.1 (0.24)} \\
MDD & 94.7 (0.58) & \textbf{96.1 (0.51)} & 76.3 (0.91) & \underline{99.0 (0.06)} & 73.5 (0.51) & 100.0 (0.00) & \underline{90.0 (1.30)} \\
NWD & 93.9 (0.12) & 92.8 (0.64) & \underline{76.4 (0.23)} & 98.6 (0.12) & \textbf{77.3 (0.44)} & 99.8 (0.00) & 89.8 (0.82) \\
SDAT & 92.7 (0.12) & 91.4 (0.46) & \textbf{78.3 (0.01)} & 98.7 (0.02) & 75.5 (0.00) & 100.0 (0.00) & 89.4 (0.48) \\
SPA & 89.4 (0.93) & 92.7 (0.33) & 67.5 (0.38) & 98.5 (0.13) & 74.3 (0.39) & 99.8 (0.00) & 87.0 (1.13) \\
FM-SD-AGAU & 87.9 (0.58) & 85.3 (0.61) & 71.3 (0.25) & 97.9 (0.00) & 72.5 (0.06) & 100.0 (0.00) & 85.8 (0.88) \\
FM-SD-AGMM & 85.5 (1.40) & 84.3 (0.17) & 68.6 (2.06) & 97.2 (0.25) & 69.1 (0.32) & 98.9 (0.23) & 83.9 (2.54) \\
FM-SD-AVAE & 89.5 (0.50) & 85.7 (0.64) & 69.5 (0.10) & 98.0 (0.00) & 70.6 (0.10) & 100.0 (0.00) & 85.6 (0.82) \\
FM-SD-KGAU & 83.0 (1.40) & 79.2 (0.53) & 69.0 (0.25) & 98.6 (0.06) & 72.0 (0.35) & 99.8 (0.35) & 83.6 (1.60) \\
FM-SD-KGMM & 81.7 (0.69) & 77.3 (1.65) & 67.8 (0.92) & 98.5 (0.06) & 67.3 (0.42) & 99.1 (0.90) & 81.9 (2.25) \\
SD-AGAU & 87.5 (1.45) & 82.7 (0.73) & 66.9 (0.37) & 98.2 (0.13) & 68.5 (0.87) & 100.0 (0.00) & 84.0 (1.88) \\
SD-AGMM & 86.5 (0.72) & 82.3 (0.40) & 68.3 (0.38) & 98.9 (0.13) & 69.0 (0.12) & 99.9 (0.12) & 84.2 (0.94) \\
SD-AVAE & 86.3 (0.00) & 82.0 (0.00) & 68.3 (0.00) & 98.4 (0.00) & 68.9 (0.00) & 100.0 (0.00) & 84.0 (0.00) \\
SD-KGAU & 87.1 (1.59) & 82.5 (0.48) & 67.9 (0.28) & 98.8 (0.07) & 69.5 (0.31) & 100.0 (0.00) & 84.3 (1.72) \\
SD-KGMM & 83.9 (0.51) & 80.8 (1.07) & 66.9 (0.54) & 98.6 (0.14) & 68.1 (0.06) & 99.9 (0.12) & 83.0 (1.31) \\
SPA-SD-AGAU & 85.3 (0.00) & 79.5 (0.00) & 61.8 (0.37) & \underline{99.0 (0.00)} & 64.0 (0.00) & 100.0 (0.00) & 81.6 (0.37) \\
SPA-SD-AGMM & 86.5 (0.00) & 84.0 (0.00) & 66.6 (0.00) & 98.6 (0.00) & 68.8 (0.00) & 100.0 (0.00) & 84.1 (0.00) \\
SPA-SD-AVAE & 88.0 (0.00) & 85.7 (0.00) & 66.8 (0.00) & 98.7 (0.00) & 68.6 (0.00) & 100.0 (0.00) & 84.6 (0.00) \\
SPA-SD-KGAU & 87.2 (0.00) & 84.3 (0.00) & 66.8 (0.00) & \textbf{99.1 (0.00)} & 69.3 (0.00) & 100.0 (0.00) & 84.4 (0.00) \\
SPA-SD-KGMM & 85.7 (0.00) & 83.3 (0.00) & 66.5 (0.00) & 98.9 (0.00) & 68.5 (0.00) & 100.0 (0.00) & 83.8 (0.00) \\
\bottomrule
\end{tabular}
    }
    
\end{table*}

\begin{table*}
    \centering
    \scriptsize
    \caption{Accuracy and standard deviation on Office31 in the scarce target setting (32 training samples available; all domains are small enough that 1\% of target data would be less than 32 samples). The best accuracy is bolded and the second best is underlined.}
    \label{table:office31-scarce}
    \resizebox{\textwidth}{!}{
\begin{tabular}{llllllll}
\toprule
Method & A2D & A2W & D2A & D2W & W2A & W2D & Mean \\
\midrule
AFN & 85.6 (0.70) & \underline{83.8 (0.71)} & 66.6 (1.64) & 97.8 (0.31) & 65.5 (0.06) & 99.3 (0.81) & 83.1 (2.11) \\
DANN & 78.6 (0.64) & 76.5 (2.70) & 60.2 (0.90) & 96.6 (0.50) & 57.5 (0.75) & 99.3 (0.42) & 78.1 (3.08) \\
ERM & 82.0 (0.31) & 77.0 (0.64) & 65.2 (1.64) & 96.7 (0.17) & 64.9 (0.10) & 99.3 (0.23) & 80.8 (1.82) \\
FixMatch & 86.7 (0.51) & 83.6 (0.57) & 68.6 (0.53) & 97.2 (0.00) & \underline{71.5 (0.08)} & 100.0 (0.00) & \underline{84.6 (0.93)} \\
JAN & 84.2 (0.50) & 80.8 (0.99) & 65.4 (1.19) & 97.6 (0.06) & 65.6 (2.40) & 99.6 (0.35) & 82.2 (2.92) \\
MCC & 84.8 (1.10) & 81.8 (0.44) & 64.5 (0.91) & 98.0 (0.55) & 63.8 (0.55) & 99.7 (0.31) & 82.1 (1.71) \\
MDD & 85.7 (2.08) & 80.4 (0.90) & 64.4 (0.18) & 98.1 (0.40) & 61.4 (0.55) & 99.8 (0.20) & 81.6 (2.39) \\
NWD & 86.2 (0.22) & 81.0 (0.61) & 67.5 (0.15) & 98.2 (0.00) & 68.3 (0.45) & 99.1 (0.23) & 83.4 (0.84) \\
SDAT & 83.7 (0.00) & 79.5 (0.12) & 67.5 (0.06) & 98.6 (0.01) & 65.0 (0.00) & 100.0 (0.00) & 82.4 (0.13) \\
SPA & 84.1 (0.35) & 78.9 (0.25) & 63.1 (0.00) & 97.5 (0.32) & 65.4 (0.00) & 99.7 (0.23) & 81.4 (0.58) \\
FM-SD-AGAU & \textbf{89.2 (0.40)} & \textbf{86.0 (0.87)} & \underline{68.7 (0.15)} & 97.3 (0.12) & \textbf{71.6 (0.15)} & 100.0 (0.00) & \textbf{85.5 (0.98)} \\
FM-SD-AGMM & 87.5 (0.36) & 83.7 (0.17) & \textbf{68.8 (0.06)} & 96.8 (0.80) & 69.5 (1.88) & 98.5 (1.33) & 84.2 (2.47) \\
FM-SD-AVAE & \underline{88.5 (0.42)} & 83.6 (0.53) & 67.9 (0.06) & 97.0 (0.06) & 70.7 (0.20) & 100.0 (0.00) & \underline{84.6 (0.71)} \\
FM-SD-KGAU & 82.9 (0.87) & 80.4 (1.32) & 67.5 (0.35) & 96.6 (0.55) & 68.8 (0.57) & 99.7 (0.23) & 82.7 (1.82) \\
FM-SD-KGMM & 80.0 (0.90) & 76.9 (0.55) & 68.0 (0.51) & 95.9 (0.59) & 67.1 (0.15) & 98.0 (0.20) & 81.0 (1.34) \\
SD-AGAU & 87.2 (0.46) & 82.0 (0.57) & 67.2 (0.50) & 98.2 (0.13) & 68.2 (0.18) & 100.0 (0.00) & 83.8 (0.91) \\
SD-AGMM & 86.5 (0.60) & 81.9 (0.29) & 68.4 (0.07) & \underline{98.7 (0.26)} & 69.2 (0.49) & 100.0 (0.00) & 84.1 (0.87) \\
SD-AVAE & 85.3 (0.00) & 81.5 (0.00) & 66.5 (0.00) & 98.4 (0.00) & 68.3 (0.00) & 100.0 (0.00) & 83.3 (0.00) \\
SD-KGAU & 86.7 (0.84) & 81.6 (0.38) & 67.6 (0.46) & 98.5 (0.07) & 69.3 (0.39) & 100.0 (0.00) & 84.0 (1.10) \\
SD-KGMM & 84.6 (1.03) & 79.8 (0.44) & 66.1 (0.00) & 98.6 (0.25) & 68.3 (0.11) & 100.0 (0.00) & 82.9 (1.15) \\
SPA-SD-AGAU & 83.9 (0.00) & 80.0 (0.00) & 64.8 (0.00) & 98.6 (0.00) & 67.2 (0.00) & 100.0 (0.00) & 82.4 (0.00) \\
SPA-SD-AGMM & 84.3 (0.00) & 82.3 (0.00) & 65.5 (0.00) & 98.6 (0.00) & 67.5 (0.00) & 100.0 (0.00) & 83.0 (0.00) \\
SPA-SD-AVAE & 83.5 (0.00) & 80.4 (0.00) & 65.0 (0.00) & \textbf{98.9 (0.00)} & 67.4 (0.00) & 100.0 (0.00) & 82.5 (0.00) \\
SPA-SD-KGAU & 85.7 (0.00) & 80.5 (0.00) & 65.3 (0.00) & \textbf{98.9 (0.00)} & 67.5 (0.00) & 100.0 (0.00) & 83.0 (0.00) \\
SPA-SD-KGMM & 86.5 (0.35) & 82.9 (0.00) & 64.3 (0.18) & \textbf{98.9 (0.00)} & 67.2 (0.00) & 100.0 (0.00) & 83.3 (0.40) \\
\bottomrule
\end{tabular}
}

    \vspace{2.5em}
\end{table*}



\begin{table}[ht!]
    \centering
    \scriptsize
    \caption{Accuracy on Office-Home, averaged across 12 domain pairs.  Full refers to accuracy trained on full target dataset (100\% of available data); scarce refers to accuracy trained in scarce target setting (maximum of 1\% of target data or 32 samples).}
    \label{table:officehome-mean}
\begin{tabular}{lll}
\toprule
Method & 100\% & 1\% \\
\midrule
AFN & 68.0 (0.99) & 64.8 (2.14) \\
DANN & 65.3 (1.87) & 55.9 (4.08) \\
ERM & 59.7 (4.50) & 59.4 (6.79) \\
FixMatch & \underline{72.6 (2.58)} & 64.5 (1.22) \\
JAN & 66.7 (6.03) & 61.8 (8.42) \\
MCC & 66.3 (1.33) & 57.3 (1.62) \\
MDD & 69.5 (1.54) & 58.4 (3.95) \\
NWD & \textbf{72.8 (0.56)} & 63.8 (0.90) \\
SDAT & 71.6 (0.82) & 62.9 (0.67) \\
SPA & 65.9 (1.10) & 61.9 (0.00) \\
FM-SD-AGAU & 68.6 (2.73) & \textbf{66.4 (1.78)} \\
FM-SD-AGMM & 67.8 (5.58) & \underline{66.1 (2.58)} \\
FM-SD-AVAE & 69.3 (1.14) & 65.8 (1.47) \\
FM-SD-KGAU & 65.4 (6.48) & 62.9 (4.18) \\
FM-SD-KGMM & 68.6 (3.41) & 64.7 (2.42) \\
SD-AGAU & 65.8 (1.56) & 65.8 (1.03) \\
SD-AGMM & 64.4 (1.06) & 64.1 (1.52) \\
SD-AVAE & 65.7 (0.00) & 65.5 (0.00) \\
SD-KGAU & 65.9 (1.09) & 65.7 (0.97) \\
SD-KGMM & 66.1 (1.07) & 65.9 (1.04) \\
SPA-SD-AGAU & 57.3 (0.00) & 62.9 (0.00) \\
SPA-SD-AGMM & 63.3 (0.00) & 62.6 (0.00) \\
SPA-SD-AVAE & 63.5 (0.00) & 62.5 (0.00) \\
SPA-SD-KGAU & 64.1 (0.20) & 63.6 (0.00) \\
SPA-SD-KGMM & 64.0 (0.00) & 63.3 (0.00) \\
\bottomrule
\end{tabular}

\end{table}


\begin{table*}[h]
    \centering
    \scriptsize
    \caption{Results on OfficeHome dataset in full target setting (trained on 100\% of target data). The first six domain pairs are shown in the top half of the table; the remaining six appear in the bottom half.}
    \label{table:oh-100pct}
\begin{tabular}{lllllll}
\toprule
Method & Ar2Cl & Ar2Pr & Ar2Rw & Cl2Ar & Cl2Pr & Cl2Rw \\
\midrule
AFN & 53.2 (0.25) & 72.4 (0.40) & 77.1 (0.29) & 65.0 (0.21) & 71.2 (0.10) & 72.3 (0.00) \\
DANN & 53.0 (0.49) & 62.9 (0.15) & 73.9 (0.14) & 56.1 (0.23) & 66.2 (0.57) & 68.5 (0.49) \\
ERM & 43.6 (1.24) & 68.7 (1.51) & 75.1 (1.63) & 53.3 (0.51) & 62.3 (0.36) & 64.4 (0.55) \\
FixMatch & \textbf{57.8 (1.20)} & \textbf{79.5 (0.06)} & \underline{82.2 (0.20)} & \underline{67.6 (0.00)} & 73.1 (1.62) & \underline{75.0 (1.14)} \\
JAN & 50.6 (0.25) & 72.9 (2.29) & 78.2 (2.05) & 61.4 (2.37) & 68.8 (1.54) & 70.3 (1.39) \\
MCC & 49.8 (0.57) & 78.6 (0.14) & 80.7 (0.21) & 49.6 (0.64) & 68.5 (0.14) & 68.4 (0.44) \\
MDD & 55.9 (0.21) & 74.8 (0.36) & 79.2 (0.64) & 62.6 (0.38) & 72.8 (0.49) & 73.5 (0.31) \\
NWD & \underline{57.5 (0.21)} & \underline{79.0 (0.00)} & \textbf{82.7 (0.06)} & \textbf{71.9 (0.21)} & \textbf{76.6 (0.35)} & \textbf{77.8 (0.23)} \\
SDAT & 55.6 (0.45) & 75.7 (0.12) & 81.0 (0.08) & 66.0 (0.00) & \underline{73.8 (0.29)} & 73.5 (0.35) \\
SPA & 50.6 (0.55) & 69.4 (0.28) & 77.4 (0.31) & 59.5 (0.19) & 69.0 (0.10) & 70.2 (0.19) \\
FM-SD-AGAU & 49.1 (0.47) & 74.0 (0.20) & 77.9 (2.54) & 66.2 (0.06) & 71.5 (0.15) & 73.3 (0.10) \\
FM-SD-AGMM & 50.8 (0.67) & 71.5 (1.69) & 79.2 (0.23) & 64.6 (0.29) & 68.7 (0.35) & 71.6 (0.36) \\
FM-SD-AVAE & 48.5 (0.61) & 74.1 (0.23) & 79.0 (0.12) & 66.5 (0.25) & 71.6 (0.40) & 73.5 (0.31) \\
FM-SD-KGAU & 48.1 (1.87) & 71.1 (1.99) & 77.0 (1.04) & 56.9 (0.35) & 65.6 (2.48) & 71.5 (2.02) \\
FM-SD-KGMM & 50.7 (0.51) & 74.3 (0.06) & 78.6 (0.32) & 64.7 (1.50) & 69.8 (2.41) & 70.4 (1.73) \\
SD-AGAU & 47.2 (0.54) & 71.9 (0.00) & 78.4 (0.08) & 64.7 (0.31) & 68.8 (0.01) & 71.5 (0.06) \\
SD-AGMM & 46.3 (0.16) & 70.4 (0.54) & 78.2 (0.42) & 60.1 (0.43) & 68.6 (0.24) & 70.8 (0.35) \\
SD-AVAE & 47.5 (0.00) & 71.0 (0.00) & 78.6 (0.00) & 64.0 (0.00) & 70.1 (0.00) & 72.3 (0.00) \\
SD-KGAU & 48.3 (0.24) & 71.5 (0.22) & 78.8 (0.28) & 64.0 (0.06) & 69.5 (0.27) & 72.4 (0.36) \\
SD-KGMM & 48.6 (0.15) & 72.0 (0.11) & 79.0 (0.21) & 64.0 (0.54) & 70.0 (0.17) & 72.4 (0.29) \\
SPA-SD-AGAU & 40.6 (0.00) & 66.0 (0.00) & 69.3 (0.00) & 53.9 (0.00) & 64.9 (0.00) & 59.0 (0.00) \\
SPA-SD-AGMM & 47.5 (0.00) & 68.8 (0.00) & 77.1 (0.00) & 57.6 (0.00) & 66.1 (0.00) & 67.3 (0.00) \\
SPA-SD-AVAE & 46.3 (0.00) & 68.4 (0.00) & 76.3 (0.00) & 59.1 (0.00) & 65.4 (0.00) & 68.3 (0.00) \\
SPA-SD-KGAU & 47.5 (0.00) & 70.2 (0.00) & 77.6 (0.00) & 58.8 (0.00) & 66.3 (0.00) & 69.8 (0.05) \\
SPA-SD-KGMM & 47.6 (0.00) & 70.2 (0.00) & 77.8 (0.00) & 58.9 (0.00) & 67.1 (0.00) & 68.8 (0.00) \\
\bottomrule
\end{tabular}
\begin{tabular}{lllllll}
\toprule
Method & Pr2Ar & Pr2Cl & Pr2Rw & Rw2Ar & Rw2Cl & Rw2Pr \\
\midrule
AFN & 64.0 (0.42) & 51.5 (0.50) & 77.7 (0.21) & 72.2 (0.23) & 57.3 (0.25) & 81.6 (0.21) \\
DANN & 57.2 (0.57) & 54.1 (1.27) & 78.5 (0.57) & 71.0 (0.45) & 61.0 (0.07) & 81.1 (0.35) \\
ERM & 53.3 (0.60) & 39.1 (0.44) & 72.7 (0.21) & 65.2 (0.93) & 42.3 (3.38) & 76.4 (0.40) \\
FixMatch & \underline{70.3 (0.61)} & \textbf{58.4 (0.44)} & \underline{82.0 (0.40)} & \underline{77.4 (0.06)} & \textbf{63.7 (0.70)} & \underline{84.7 (0.15)} \\
JAN & 63.0 (2.90) & 49.6 (0.81) & 78.0 (1.10) & 72.0 (0.78) & 54.1 (2.29) & 81.6 (0.90) \\
MCC & 62.0 (0.21) & 42.1 (0.75) & 81.0 (0.07) & 74.2 (0.21) & 56.2 (0.21) & \underline{84.7 (0.25)} \\
MDD & 63.1 (0.99) & 54.5 (0.32) & 79.7 (0.14) & 73.6 (0.21) & 60.1 (0.21) & 84.0 (0.35) \\
NWD & \textbf{71.7 (0.06)} & 54.8 (0.17) & \textbf{82.3 (0.12)} & 76.4 (0.00) & 56.8 (0.00) & \textbf{85.7 (0.00)} \\
SDAT & 69.3 (0.06) & \underline{56.0 (0.00)} & 81.9 (0.30) & \textbf{78.9 (0.12)} & \underline{61.4 (0.23)} & \textbf{85.7 (0.28)} \\
SPA & 61.1 (0.51) & 49.7 (0.26) & 78.1 (0.45) & 70.7 (0.28) & 54.2 (0.17) & 80.8 (0.09) \\
FM-SD-AGAU & 68.2 (0.23) & 51.0 (0.35) & 80.1 (0.31) & 74.8 (0.06) & 54.4 (0.62) & 83.0 (0.15) \\
FM-SD-AGMM & 68.0 (0.46) & 51.3 (2.28) & 79.2 (0.23) & 73.5 (0.06) & 52.7 (4.69) & 82.9 (0.23) \\
FM-SD-AVAE & 67.8 (0.15) & 53.3 (0.42) & 79.7 (0.14) & 76.8 (0.35) & 57.7 (0.45) & 83.5 (0.10) \\
FM-SD-KGAU & 66.8 (1.59) & 44.2 (2.50) & 79.5 (0.46) & 72.4 (0.40) & 48.1 (3.64) & 83.6 (0.81) \\
FM-SD-KGMM & 67.5 (0.25) & 52.7 (0.23) & 79.6 (0.21) & 75.9 (0.25) & 57.8 (0.10) & 81.8 (0.06) \\
SD-AGAU & 64.3 (0.07) & 44.1 (0.66) & 78.6 (0.40) & 71.4 (0.23) & 48.3 (1.16) & 80.0 (0.15) \\
SD-AGMM & 59.6 (0.05) & 42.6 (0.26) & 77.1 (0.27) & 69.8 (0.32) & 48.8 (0.07) & 80.3 (0.07) \\
SD-AVAE & 62.6 (0.00) & 43.2 (0.00) & 77.9 (0.00) & 71.7 (0.00) & 48.2 (0.00) & 81.2 (0.00) \\
SD-KGAU & 62.8 (0.29) & 43.6 (0.60) & 77.8 (0.10) & 71.4 (0.33) & 49.4 (0.42) & 81.5 (0.22) \\
SD-KGMM & 63.1 (0.43) & 43.8 (0.56) & 78.6 (0.20) & 71.5 (0.21) & 49.0 (0.28) & 81.2 (0.08) \\
SPA-SD-AGAU & 47.7 (0.00) & 37.9 (0.00) & 67.1 (0.00) & 60.5 (0.00) & 45.6 (0.00) & 74.6 (0.00) \\
SPA-SD-AGMM & 58.1 (0.00) & 42.3 (0.00) & 75.4 (0.00) & 70.3 (0.00) & 48.3 (0.00) & 80.2 (0.00) \\
SPA-SD-AVAE & 59.8 (0.00) & 43.0 (0.00) & 76.7 (0.00) & 70.0 (0.00) & 48.7 (0.00) & 79.9 (0.00) \\
SPA-SD-KGAU & 59.7 (0.00) & 41.5 (0.00) & 76.0 (0.20) & 70.8 (0.00) & 49.2 (0.00) & 81.3 (0.00) \\
SPA-SD-KGMM & 59.7 (0.00) & 41.9 (0.00) & 76.3 (0.00) & 70.5 (0.00) & 49.4 (0.00) & 80.4 (0.00) \\
\bottomrule
\end{tabular}

\end{table*}

\begin{table*}[h]
    \centering
    \scriptsize
    \caption{Results on OfficeHome dataset in scarce target setting (trained on maximum of 1\% of target data or 32 samples). The first six domain pairs are shown in the top half of the table; the remaining six appear in the bottom half.}
    \label{table:oh-1pct}
\begin{tabular}{lllllll}
\toprule
Method & Ar2Cl & Ar2Pr & Ar2Rw & Cl2Ar & Cl2Pr & Cl2Rw \\
\midrule
AFN & \textbf{50.3 (0.47)} & 69.5 (0.75) & 76.4 (0.21) & 59.4 (0.98) & 67.9 (0.90) & 70.2 (0.49) \\
DANN & 41.2 (0.32) & 61.9 (1.56) & 70.2 (1.10) & 48.0 (0.71) & 56.1 (1.25) & 59.3 (0.57) \\
ERM & 44.0 (0.49) & 68.3 (1.97) & 74.5 (0.55) & 52.5 (0.50) & 62.3 (0.06) & 63.8 (0.28) \\
FixMatch & 47.2 (0.17) & 70.4 (0.20) & 77.6 (0.21) & 59.2 (0.45) & 67.3 (0.46) & 69.1 (0.13) \\
JAN & 44.8 (0.61) & 68.4 (2.80) & 76.2 (2.14) & 55.9 (3.30) & 63.7 (2.98) & 66.1 (2.19) \\
MCC & 41.4 (0.32) & 63.4 (0.42) & 72.8 (0.21) & 48.5 (0.57) & 58.6 (0.62) & 58.8 (0.24) \\
MDD & 42.2 (1.21) & 64.6 (0.85) & 72.5 (0.46) & 52.9 (0.71) & 60.8 (0.67) & 62.3 (3.24) \\
NWD & 46.4 (0.47) & \textbf{72.0 (0.00)} & 78.4 (0.00) & 61.1 (0.17) & 66.2 (0.21) & 69.2 (0.29) \\
SDAT & 45.5 (0.00) & 70.9 (0.00) & 78.2 (0.33) & 54.3 (0.00) & 66.4 (0.00) & 65.7 (0.16) \\
SPA & 45.1 (0.00) & 67.5 (0.00) & 76.2 (0.00) & 57.2 (0.00) & 64.5 (0.00) & 66.8 (0.00) \\
FM-SD-AGAU & 48.1 (0.00) & 71.2 (0.50) & 78.5 (0.21) & 63.1 (1.46) & \textbf{70.2 (0.25)} & 71.9 (0.65) \\
FM-SD-AGMM & \underline{48.9 (0.21)} & 70.8 (0.26) & \textbf{79.0 (0.12)} & 63.6 (0.81) & 69.3 (0.20) & 71.4 (0.00) \\
FM-SD-AVAE & 46.6 (0.26) & 70.4 (0.36) & 78.4 (0.10) & 63.0 (0.76) & 68.6 (0.40) & 72.2 (0.10) \\
FM-SD-KGAU & 44.5 (3.06) & 68.1 (0.83) & 74.8 (0.26) & 57.8 (0.17) & 63.6 (0.15) & 66.6 (0.32) \\
FM-SD-KGMM & 47.0 (0.98) & 70.2 (0.17) & 78.0 (1.27) & 57.8 (0.92) & 66.3 (0.35) & 69.3 (0.52) \\
SD-AGAU & 47.8 (0.32) & 71.3 (0.15) & 78.6 (0.20) & \textbf{65.0 (0.04)} & 67.4 (0.48) & 71.8 (0.60) \\
SD-AGMM & 46.6 (0.29) & 70.7 (0.36) & 77.5 (0.32) & 59.5 (0.63) & 67.4 (0.44) & 70.5 (0.61) \\
SD-AVAE & 46.0 (0.00) & 70.9 (0.00) & 78.6 (0.00) & \underline{64.5 (0.00)} & 69.2 (0.00) & 72.0 (0.00) \\
SD-KGAU & 47.9 (0.30) & 71.5 (0.16) & 78.6 (0.07) & 63.1 (0.43) & \underline{69.5 (0.06)} & \textbf{72.6 (0.43)} \\
SD-KGMM & 47.9 (0.03) & \underline{71.6 (0.23)} & \underline{78.9 (0.00)} & 63.5 (0.50) & \underline{69.5 (0.01)} & \underline{72.5 (0.18)} \\
SPA-SD-AGAU & 46.9 (0.00) & 67.3 (0.00) & 74.7 (0.00) & 59.4 (0.00) & 66.3 (0.00) & 68.3 (0.00) \\
SPA-SD-AGMM & 45.8 (0.00) & 68.2 (0.00) & 76.7 (0.00) & 57.5 (0.00) & 64.5 (0.00) & 66.8 (0.00) \\
SPA-SD-AVAE & 44.5 (0.00) & 66.9 (0.00) & 75.6 (0.00) & 58.0 (0.00) & 64.1 (0.00) & 68.9 (0.00) \\
SPA-SD-KGAU & 46.0 (0.00) & 69.5 (0.00) & 77.7 (0.00) & 58.2 (0.00) & 66.5 (0.00) & 69.7 (0.00) \\
SPA-SD-KGMM & 46.5 (0.00) & 69.8 (0.00) & 77.8 (0.00) & 58.1 (0.00) & 65.3 (0.00) & 69.1 (0.00) \\
\bottomrule
\end{tabular}
\begin{tabular}{lllllll}
\toprule
Method & Pr2Ar & Pr2Cl & Pr2Rw & Rw2Ar & Rw2Cl & Rw2Pr \\
\midrule
AFN & 58.5 (0.06) & \textbf{47.0 (0.96)} & 76.7 (0.44) & 69.0 (0.52) & \textbf{52.3 (0.51)} & 79.7 (0.31) \\
DANN & 48.8 (0.80) & 36.2 (0.78) & 67.9 (2.94) & 62.6 (0.14) & 44.3 (0.78) & 74.7 (0.14) \\
ERM & 52.7 (0.95) & 35.7 (6.30) & 73.2 (0.17) & 65.0 (0.84) & 44.5 (0.10) & 75.7 (0.21) \\
FixMatch & 62.5 (0.10) & 42.6 (0.46) & 76.6 (0.12) & 71.4 (0.12) & 49.7 (0.82) & 79.9 (0.15) \\
JAN & 58.0 (3.95) & 40.4 (1.39) & 75.0 (2.04) & 67.4 (2.62) & 47.9 (0.79) & 78.0 (2.09) \\
MCC & 51.5 (0.70) & 36.3 (0.14) & 71.4 (0.44) & 64.3 (0.14) & 44.2 (0.55) & 76.2 (0.71) \\
MDD & 51.9 (0.38) & 37.5 (0.78) & 71.0 (0.63) & 63.5 (0.78) & 45.8 (0.21) & 76.0 (0.00) \\
NWD & 62.2 (0.17) & 38.7 (0.52) & 77.2 (0.23) & 70.0 (0.15) & 44.5 (0.23) & 79.8 (0.06) \\
SDAT & 59.1 (0.00) & 40.2 (0.52) & 75.9 (0.09) & 69.9 (0.19) & 46.4 (0.01) & \textbf{81.8 (0.01)} \\
SPA & 57.4 (0.00) & 39.7 (0.00) & 75.0 (0.00) & 68.5 (0.00) & 46.5 (0.00) & 78.0 (0.00) \\
FM-SD-AGAU & \underline{64.6 (0.06)} & \underline{45.9 (0.45)} & \underline{78.7 (0.20)} & \underline{72.6 (0.06)} & \underline{50.6 (0.00)} & \underline{81.1 (0.06)} \\
FM-SD-AGMM & 63.8 (0.46) & 44.6 (0.46) & \textbf{79.0 (0.40)} & \textbf{72.7 (0.12)} & 49.3 (2.17) & 80.7 (0.72) \\
FM-SD-AVAE & \textbf{64.7 (0.69)} & 45.4 (0.58) & 77.8 (0.10) & \textbf{72.7 (0.58)} & 49.2 (0.20) & 80.2 (0.10) \\
FM-SD-KGAU & 63.3 (0.00) & 43.4 (1.65) & 77.3 (0.46) & 71.6 (0.17) & 45.4 (1.79) & 78.3 (1.01) \\
FM-SD-KGMM & 62.8 (0.87) & 45.4 (0.86) & 78.1 (0.36) & 72.4 (0.53) & 49.3 (0.10) & 80.3 (0.31) \\
SD-AGAU & 64.5 (0.13) & 44.2 (0.07) & 78.5 (0.31) & 71.5 (0.06) & 48.3 (0.29) & 80.3 (0.31) \\
SD-AGMM & 59.8 (0.56) & 41.6 (0.25) & 77.0 (0.64) & 69.7 (0.40) & 48.0 (0.20) & 80.5 (0.22) \\
SD-AVAE & 63.3 (0.00) & 43.2 (0.00) & 78.4 (0.00) & 71.7 (0.00) & 47.4 (0.00) & 80.4 (0.00) \\
SD-KGAU & 62.7 (0.02) & 43.7 (0.21) & 78.1 (0.08) & 71.7 (0.13) & 48.6 (0.60) & 80.8 (0.15) \\
SD-KGMM & 63.2 (0.17) & 43.5 (0.23) & 78.0 (0.20) & 71.6 (0.55) & 49.3 (0.34) & 80.8 (0.45) \\
SPA-SD-AGAU & 59.4 (0.00) & 39.6 (0.00) & 75.8 (0.00) & 70.0 (0.00) & 48.7 (0.00) & 78.1 (0.00) \\
SPA-SD-AGMM & 58.3 (0.00) & 41.2 (0.00) & 75.0 (0.00) & 69.9 (0.00) & 48.0 (0.00) & 79.8 (0.00) \\
SPA-SD-AVAE & 59.2 (0.00) & 40.6 (0.00) & 75.1 (0.00) & 69.0 (0.00) & 47.7 (0.00) & 79.8 (0.00) \\
SPA-SD-KGAU & 60.1 (0.00) & 40.8 (0.00) & 76.1 (0.00) & 70.4 (0.00) & 47.5 (0.00) & 80.4 (0.00) \\
SPA-SD-KGMM & 58.5 (0.00) & 40.5 (0.00) & 75.9 (0.00) & 69.9 (0.00) & 47.1 (0.00) & 80.8 (0.00) \\
\bottomrule
\end{tabular}

\end{table*}


\begin{table*}[ht!]
    \centering
    \scriptsize
    \caption{Accuracy and standard deviation on VisDA-2017 from sythetic domain to real domain, with 100\%, 1\% and 0.1\% of target data available.}
    \label{table:visda}
\begin{tabular}{llll}
\toprule
Method & 100\% & 1\% & 0.1\% \\
\midrule
AFN & 68.0 (0.38) & 68.3 (0.34) & 67.0 (0.00) \\
DANN & 75.9 (0.25) & 72.7 (2.69) & 66.8 (0.19) \\
ERM & 52.8 (0.45) & 53.0 (0.17) & 52.9 (0.51) \\
FixMatch & \textbf{78.8 (1.10)} & 71.6 (0.81) & 67.5 (0.58) \\
JAN & 70.6 (0.12) & 69.5 (0.00) & 67.4 (0.12) \\
MCC & 75.4 (0.06) & 72.2 (0.23) & \underline{68.8 (0.00)} \\
MDD & \underline{77.9 (1.27)} & 69.3 (0.00) & 64.8 (0.09) \\
NWD & 75.8 (0.00) & 72.3 (0.00) & 66.9 (2.67) \\
SDAT & 77.2 (2.66) & 71.9 (1.53) & 67.4 (0.61) \\
SPA & 72.8 (0.34) & 72.0 (0.63) & 67.6 (0.00) \\
FM-SD-AGAU & 72.8 (0.06) & 74.4 (2.94) & 68.6 (0.96) \\
FM-SD-AGMM & 72.0 (1.65) & 70.5 (0.76) & 67.5 (1.12) \\
FM-SD-AVAE & 76.5 (2.75) & \textbf{75.6 (1.27)} & \textbf{70.5 (1.21)} \\
FM-SD-KGAU & 76.0 (0.00) & 73.6 (1.15) & 68.5 (0.85) \\
FM-SD-KGMM & 74.6 (2.36) & \underline{74.7 (0.35)} & 68.0 (0.12) \\
SD-AGAU & 68.1 (0.07) & 67.5 (0.35) & 67.7 (0.29) \\
SD-AGMM & 68.2 (0.29) & 68.2 (0.36) & 67.7 (0.35) \\
SD-AVAE & 69.1 (0.00) & 69.0 (0.00) & 68.5 (0.00) \\
SD-KGAU & 68.6 (0.50) & 68.4 (0.56) & 67.8 (0.13) \\
SD-KGAU-REG & 68.4 (0.37) & 68.7 (0.17) & 68.4 (0.08) \\
SD-KGMM & 69.3 (0.06) & 68.7 (0.01) & 68.1 (0.05) \\
SD-KGMM-REG & 68.1 (0.09) & 68.6 (0.09) & 68.2 (0.23) \\
SPA-SD-AGAU & 66.1 (0.12) & 65.5 (0.00) & 66.5 (0.00) \\
SPA-SD-AGMM & 65.9 (0.00) & 67.0 (0.00) & 65.5 (0.00) \\
SPA-SD-AVAE & 68.4 (0.00) & 66.9 (0.00) & 66.4 (0.00) \\
SPA-SD-KGAU & 66.7 (0.45) & 66.3 (0.67) & 64.5 (0.47) \\
SPA-SD-KGMM & 66.9 (0.83) & 66.4 (0.32) & 64.7 (0.00) \\
\bottomrule
\end{tabular}

\end{table*}


\begin{figure}[t]
    \centering
    \begin{minipage}[t]{0.48\textwidth}
        \centering
        \includegraphics[width=\linewidth]{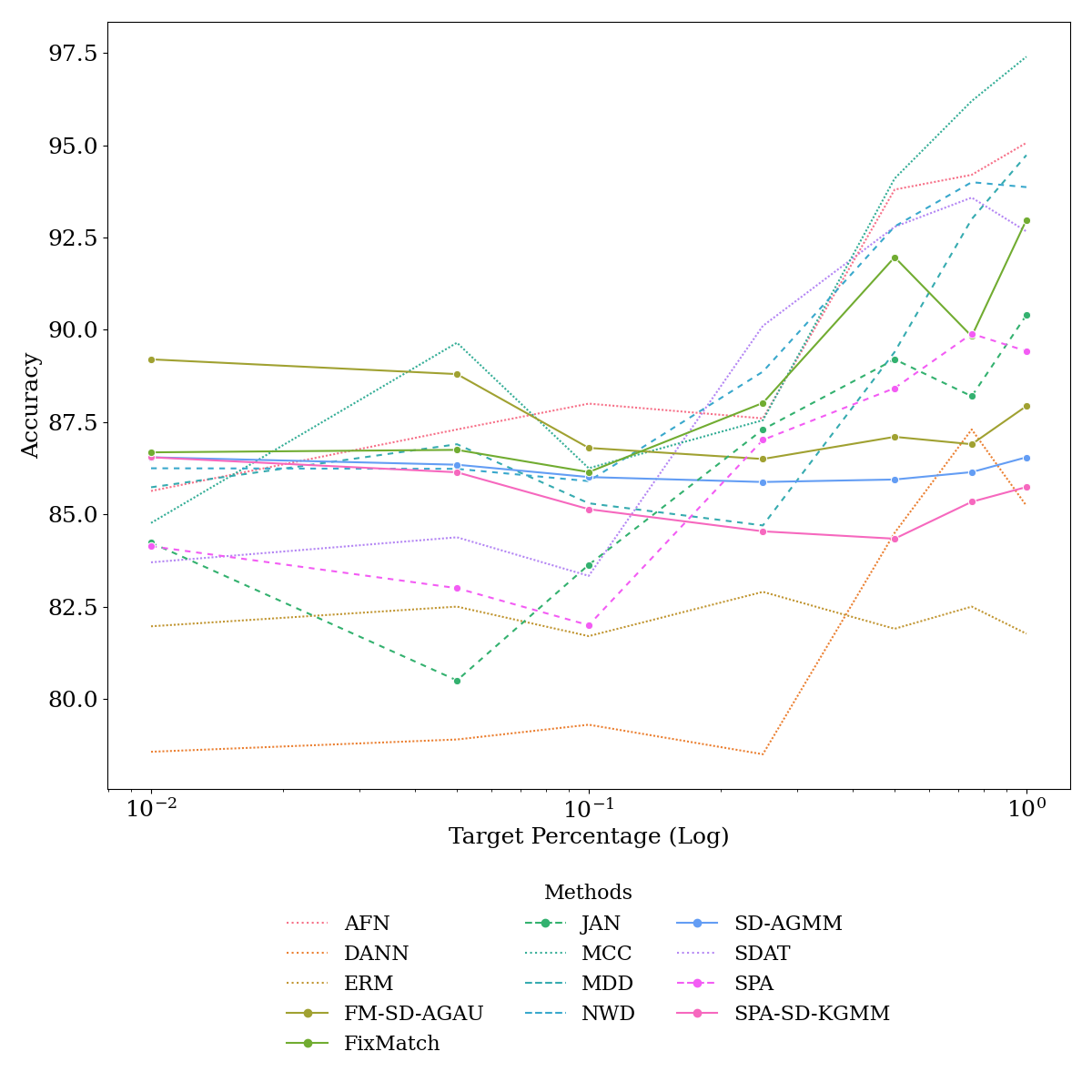}
    \end{minipage}\hfill
    \begin{minipage}[t]{0.48\textwidth}
        \centering
        \includegraphics[width=\linewidth]{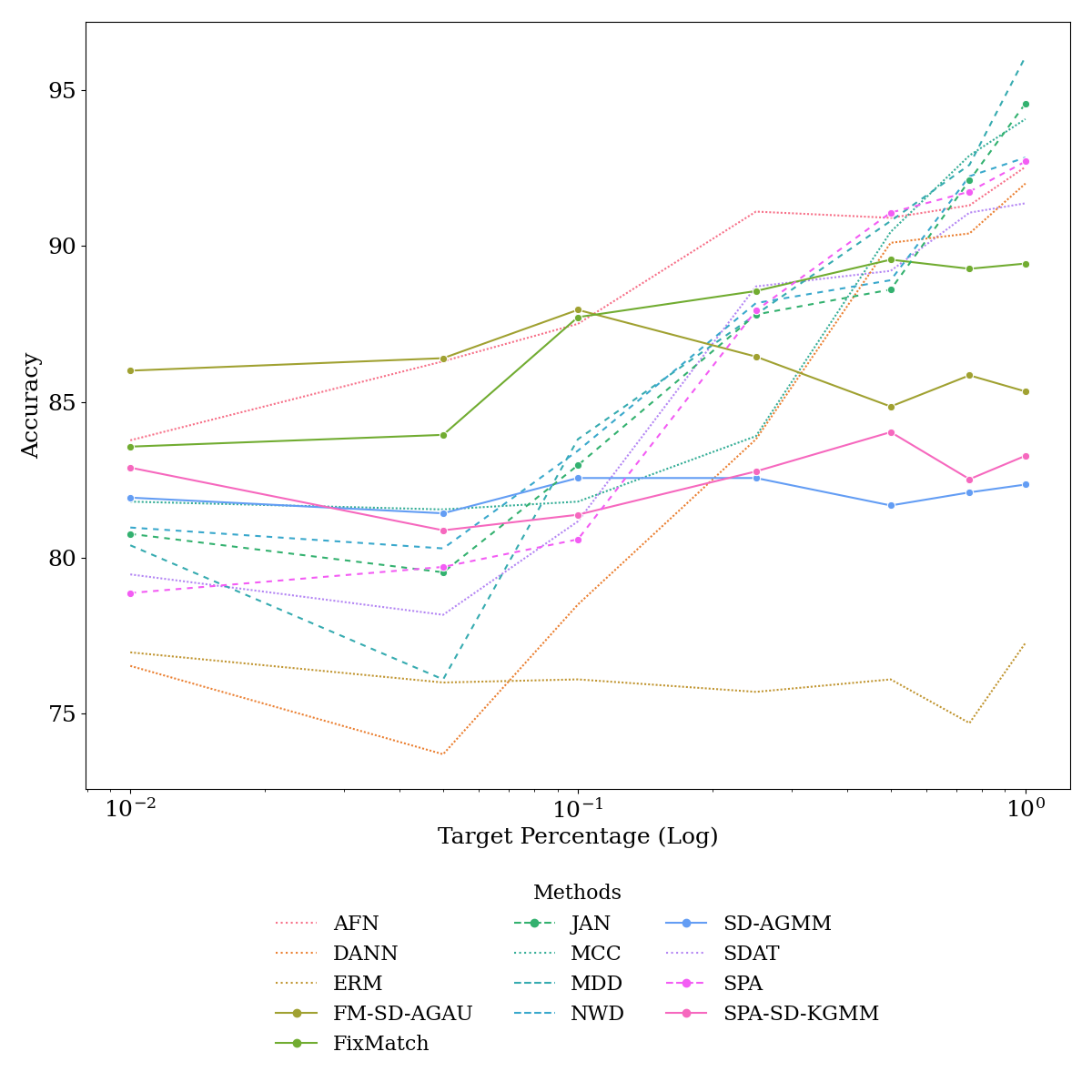}
    \end{minipage}
    \caption{Accuracy on A2D (left) and A2W (right) domain pairs of Office31.}
    \label{fig:ablation_A2D_A2W}
\end{figure}

\begin{figure}[t]
    \centering
    \begin{minipage}[t]{0.48\textwidth}
        \centering
        \includegraphics[width=\linewidth]{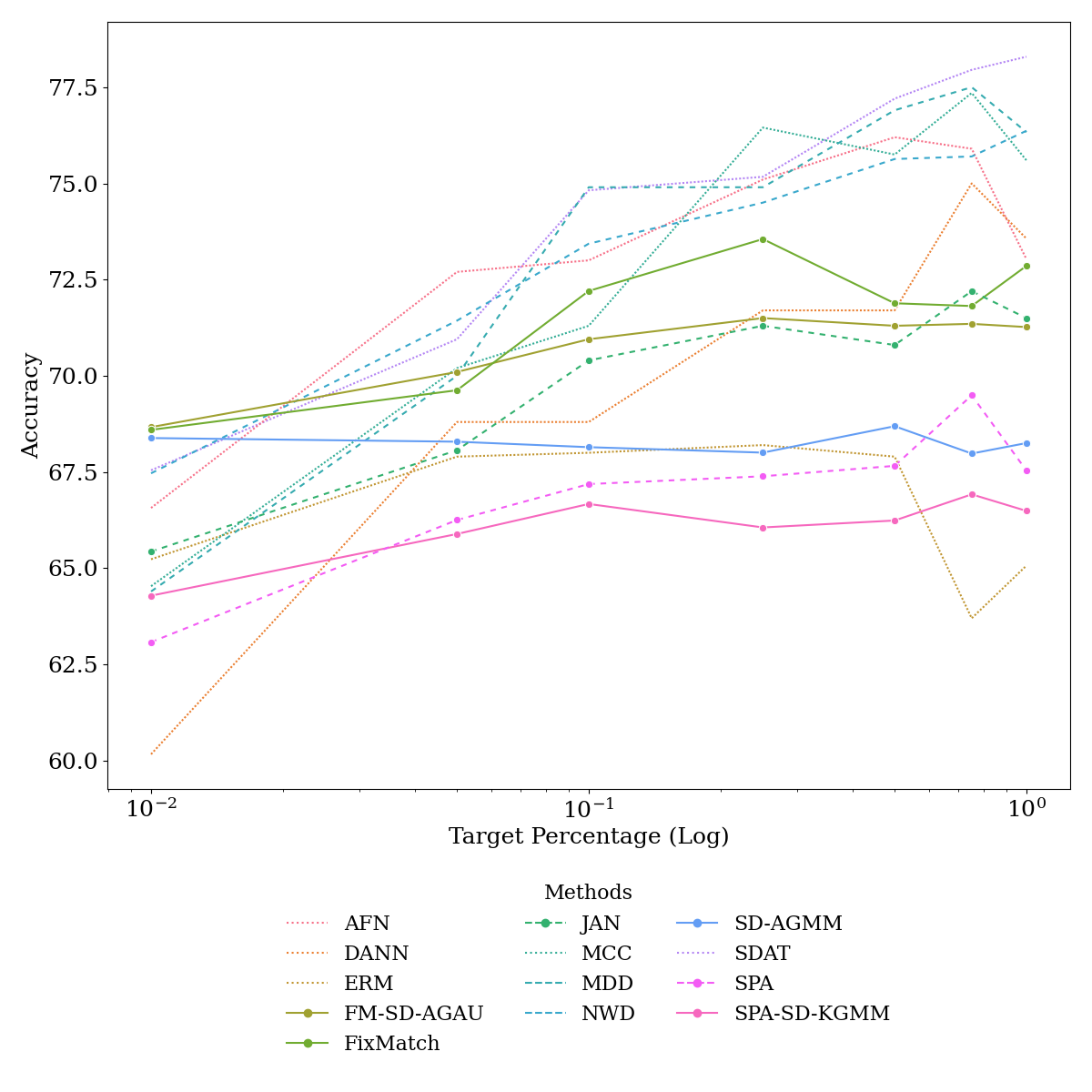}
    \end{minipage}\hfill
    \begin{minipage}[t]{0.48\textwidth}
        \centering
        \includegraphics[width=\linewidth]{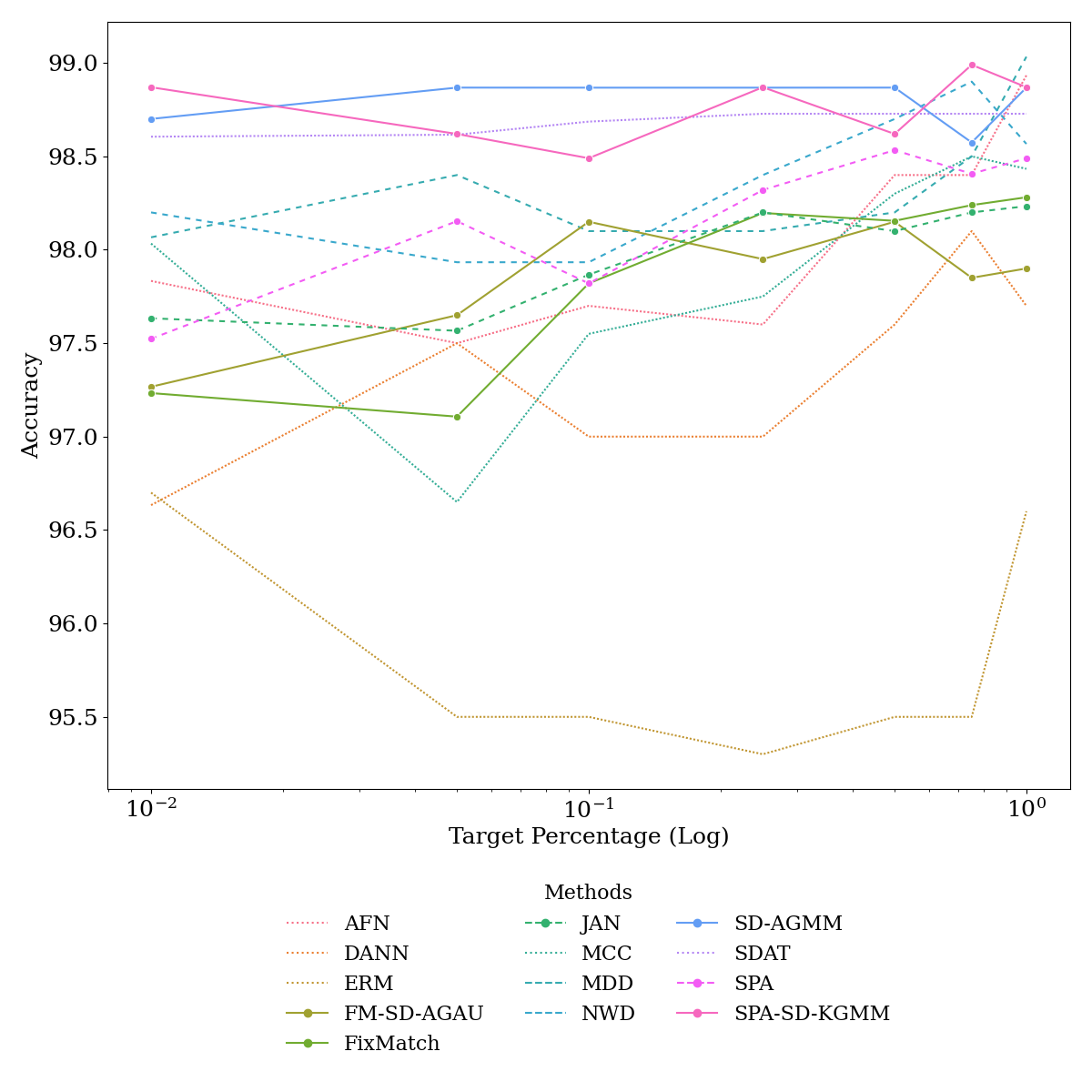}
    \end{minipage}
    \caption{Accuracy on D2A (left) and D2W (right) domain pairs of Office31.}
    \label{fig:ablation_D2A_D2W}
\end{figure}

\begin{figure}[t]
    \centering
    \begin{minipage}[t]{0.48\textwidth}
        \centering
        \includegraphics[width=\linewidth]{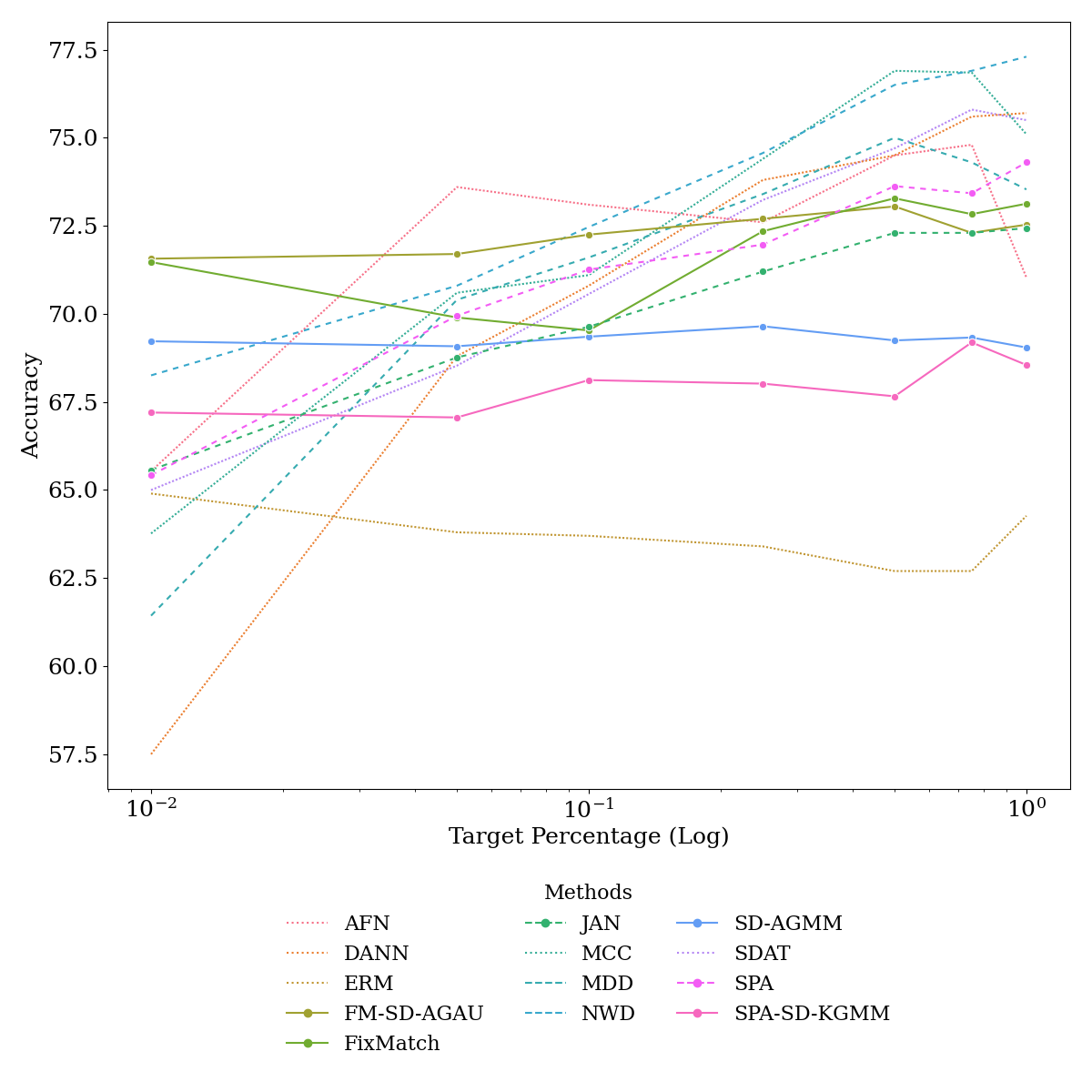}
    \end{minipage}\hfill
    \begin{minipage}[t]{0.48\textwidth}
        \centering
        \includegraphics[width=\linewidth]{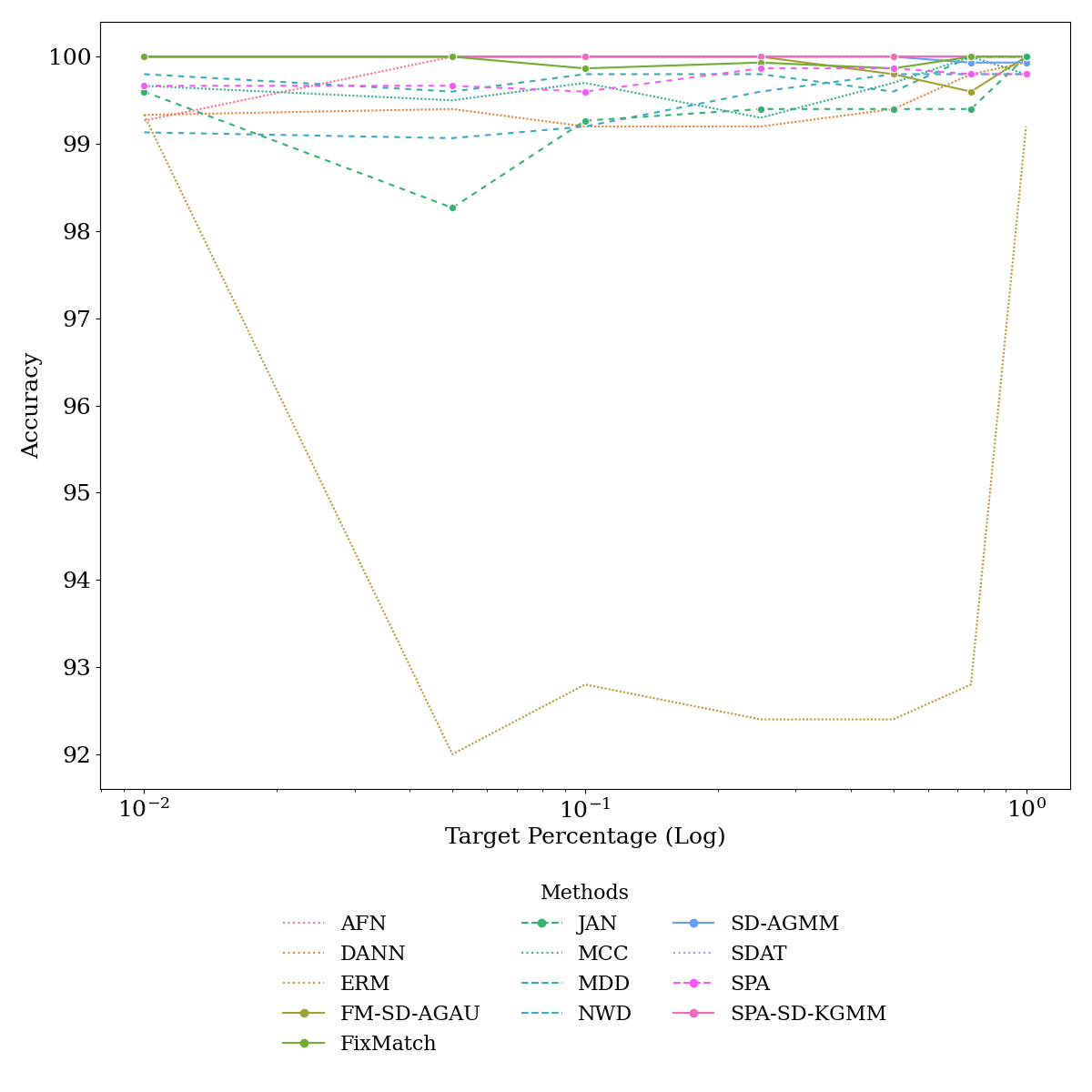}
    \end{minipage}
    \caption{Accuracy on W2A (left) and W2D (right) domain pairs of Office31.}
    \label{fig:ablation_W2A_W2D}
\end{figure}

To supplement the results in Figure \ref{fig:target-pct-comparison}, we include the results on each domain pair individually in Figures \ref{fig:ablation_A2D_A2W}-\ref{fig:ablation_W2A_W2D}.
Each figure shows the accuracy of \ac{uda} methods on the Office31 dataset, when different amounts of target data are made available during training.
Although Stein discrepancy-based methods are not competitive when large amounts of target data are available, they are very stable to decreasing amounts of target data and are among the best performers when the amount of target data is reduced to 1\%.

\subsection{Sensitivity to target sample}\label{subsec:visda-stat-sig}
To ensure reproducibility, random seeds were fixed across all experiments.
For scarce target settings, the seed was reinitialized prior to subsampling the target data so that all methods were trained on the same subset.

\begin{figure}
    \centering
    \includegraphics[width=0.7\columnwidth]{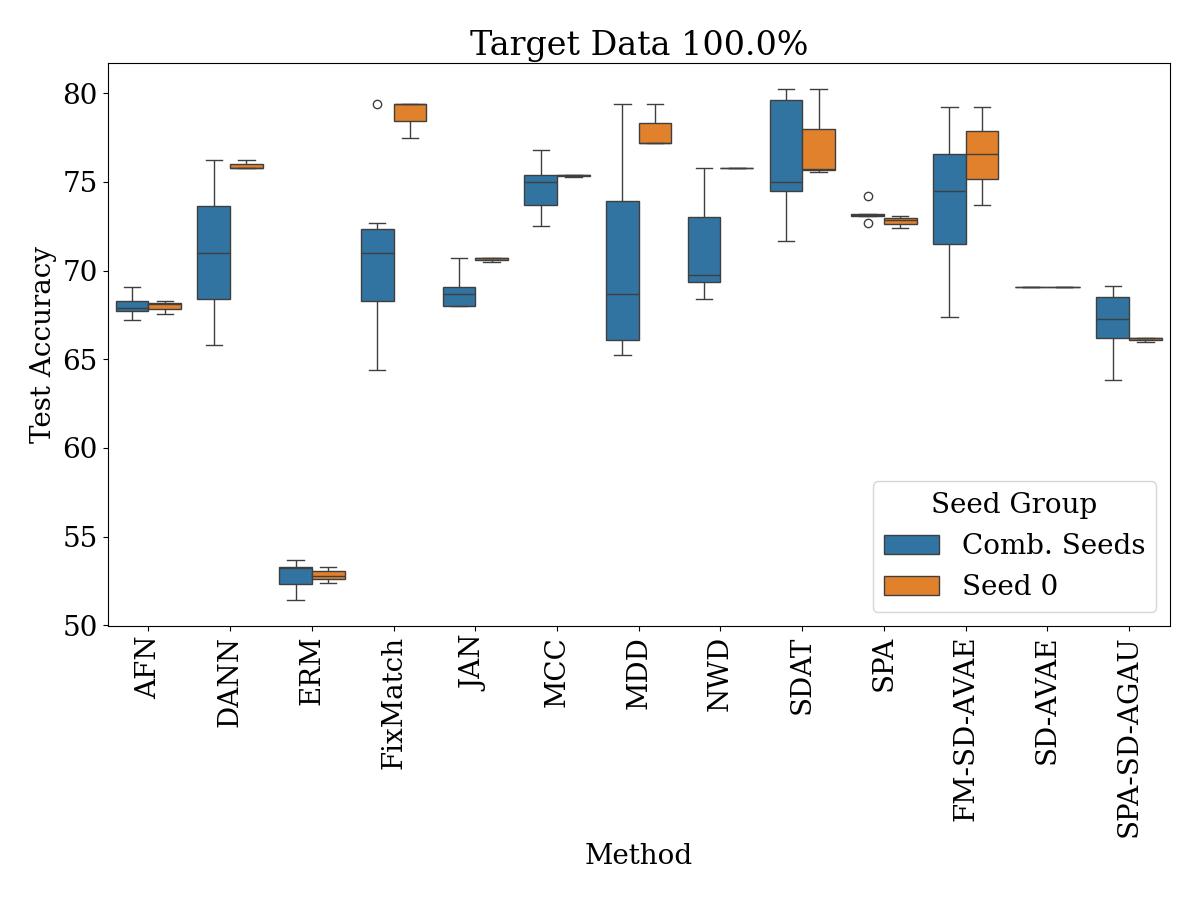}
    \caption{Comparing variance on VisDA-2017 dataset with 100\% of target data used in training.}
    \label{fig:seed-100pct}
\end{figure}

\begin{figure}
    \centering
    \includegraphics[width=0.7\linewidth]{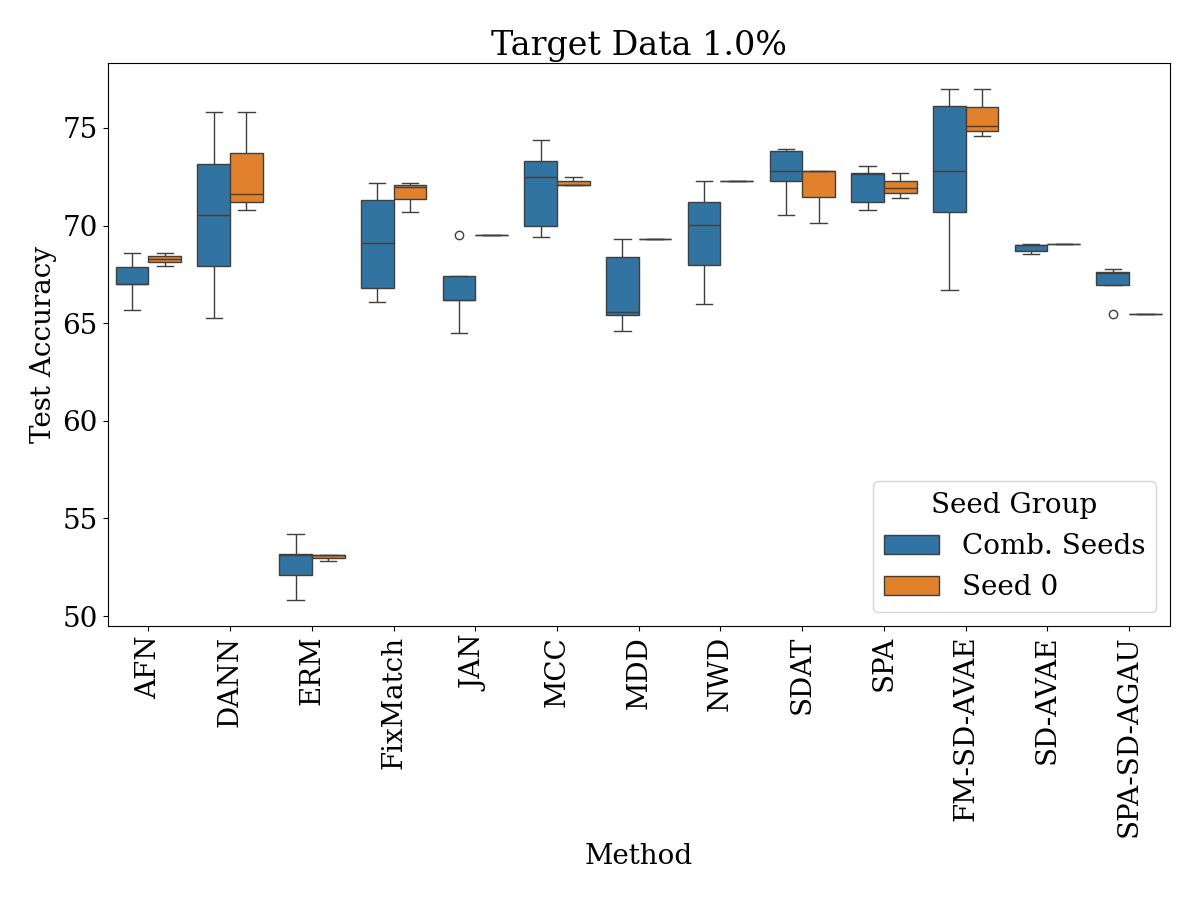}
    \caption{Comparing variance on VisDA-2017 dataset with 1\% of target data used in training.}
    \label{fig:seed-1pct}
\end{figure}

\begin{figure}
    \centering
    \includegraphics[width=0.7\linewidth]{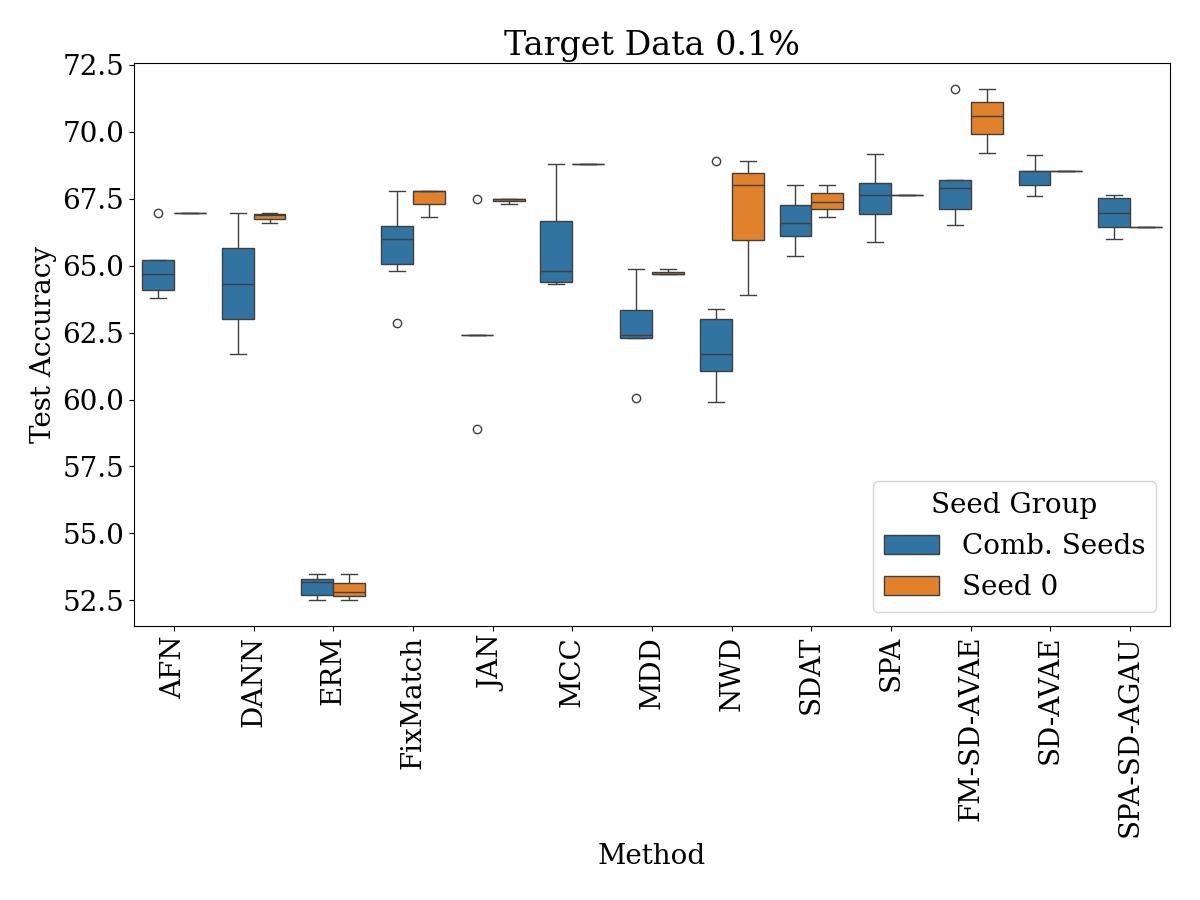}
    \caption{Comparing variance on VisDA-2017 dataset with 0.1\% of target data used in training.}
    \label{fig:seed-point1pct}
\end{figure}

Despite this control, some methods still exhibited noticeable variance between runs.
To better understand this variability and how much of an effect the specific sample of target data has, we ran all methods on the VisDA-2017 dataset using five different seeds (0 through 4).
The main results presented in the paper use seed 0.
Here, we separate the results into two groups: runs with seed 0, and runs with seeds 1–4.
We visualize the results using boxplots in Figures~\ref{fig:seed-100pct}–\ref{fig:seed-point1pct}.
Similarly to Figures ~\ref{fig:o31-main}-\ref{fig:vis-main}, we report only the higest performing variant among Stein discrepancy-based methods in each framework (plain, FixMatch, and SPA).

Even in the 100\% target setting, where all target data is used and no subsampling occurs, we observe higher variance across the full seed group compared to seed 0 alone.
This indicates that randomness in model initialization (e.g., weight initialization) contributes meaningfully to performance variability.
In the two scarce settings (1\% and 0.1\% target data), all methods show greater variance in the full seed group than in the seed 0 runs, reflecting sensitivity to the specific target subset.
Notably, Stein discrepancy-based methods show less sensitivity to this variation than other approaches.
Adversarial methods such as DANN, FDAL, and JAN tend to exhibit higher variance in both conditions.
These results suggest that methods with lower sensitivity to the target subsample may offer more stable performance in practical scenarios, where target data is limited and selection effects can meaningfully impact outcomes.


\section{Experiment implementation}\label{app:implementation}

\subsection{Hyperparameter selection}\label{subsec:hyperparams}

Hyperparameter tuning for Stein discrepancy-based methods was performed using Raytune \citep{liaw2018tune}.
Hyperparameters were tuned independently; the learning rate was tuned using default values for each method and then additional hyperparameters were tuned using the best learning rate.
The same hyperparameters were used across domains and target percentages.

The following hyperparameters were tuned for all Stein discrepancy-based methods: learning rate, momentum, epoch trade-off, bottleneck dimension, and rescaling function.
The epoch trade-off parameter controls how fast the transfer loss is phased in as part of training; recall that the loss is a combination of classification loss on the source domain and transfer loss, which captures distance between source and target domains.
For all of the methods, they are trained for one epoch with no transfer loss (only maximizing classification loss).
The transfer loss is rescaled using a sigmoid function and the trade-off parameter controls how quickly the sigmoid function grows.
The rescaling function, either hyperbolic tangent or sigmoid, is used to rescale the Stein discrepancy before it is combined with the classification loss.
For all methods, hyperbolic tangent outperformed sigmoid and was used in all final experiments.

For the adversarial methods, the following additional hyperparameters were tuned: learning rate for the adversarial network and dimension of the hidden layers in the adversarial network.
The adversarial network always has two layers.
The kernelized methods tuned the following additional hyperparameters: the kernel bandwidth and the type of kernel, radial basis function or inverse multiquadric.
For all methods and datasets, the radial basis function kernel was used for final experiments.

Methods using a GMM target distribution tuned the following hyperparameters: the learning rate for the GMM, the number of components in the GMM, and the covariance type of the GMM, full or diagonal.
The best covariance type was always diagonal.
The SD-AVAE method had several hyperparameters related to training the VAE: a learning rate for the VAE, the hidden and latent dimensions of the VAE, and the number of training steps for the VAE.
One training step was used for all datasets for SD-AVAE.

Finally, methods in the FixMatch and SPA frameworks had additional hyperparameters from the baseline method.
Methods in the FixMatch framework trained the following additional hyperparameters:
a confidence threshold to determine when pseudo-labels are confident enough to use and a tradeoff between the pseudo-labeling loss and other terms in the loss function.
Methods in the SPA framework trained the following additional hyperparameters: pseudo-labeling type, pseudo-labeling trade-off, SVD trade-off, adjacency-similarity function, and Laplacian type.
Recall that SPA constructs a graph from the training data; the adjacency-similarity function and Laplacian are used in constructing the graph.
The other hyperparameters control whether or not other terms are added to the loss function and how much weight they have relative to the classification and transfer losses.
These are all parameters in the original SPA code and the best values after hyperparameter tuning were the same as those used for the SPA experiments without Stein discrepancy with the exception of the pseudo-labeling type.
The best results were obtained with no pseudo-labeling term when Stein discrepancy was included, while the original SPA experiments used soft pseudo-label assignment.

The hyperparameters for each method on each dataset are reported in Tables \ref{table:hyper-sd-agau}-\ref{table:hyper-sd-kgmm}.

\begin{table*}
    \centering
    \tiny
    \caption{Hyperparameters for SD-AGAU.}
    \label{table:hyper-sd-agau}
\begin{tabular}{lccc>{\columncolor[HTML]{EEEEEE}}c>{\columncolor[HTML]{EEEEEE}}c>{\columncolor[HTML]{EEEEEE}}cccc}
\toprule
Dataset & \multicolumn{3}{c}{Office31} & \multicolumn{3}{>{\columncolor[HTML]{EEEEEE}}c}{OfficeHome} & \multicolumn{3}{c}{VisDA2017} \\
Framework & FixMatch & Plain & SPA & FixMatch & Plain & SPA & FixMatch & Plain & SPA \\
\midrule
LR & $0.003$ & $0.1$ & $0.03$ & $0.005$ & $0.1$ & $0.025$ & $0.01$ & $0.1$ & $0.01$ \\
Momentum & $0.8$ & $0.05$ & $0.3$ & $0.8$ & $0.05$ & $0.2$ & $0.94$ & $0$ & $0.8$ \\
Bottleneck Dim & 256 & 256 & 256 & 256 & 256 & 256 & 256 & 256 & 256 \\
Epoch Tradeoff & $1.5$ & $0.65$ & $1$ & $1.9$ & $0.65$ & $1$ & $0.1$ & $0.25$ & $1$ \\
Adv. LR & $0.0001$ & $0.0001$ & $0.01$ & $0.0001$ & $0.0001$ & $0.01$ & $0.0001$ & $0.0001$ & $0.01$ \\
Adv. Dim & 1024 & 512 & 1024 & 512 & 512 & 1024 & 512 & 256 & 1024 \\
Threshold & $0.85$ & — & — & $0.9$ & — & — & $0.9$ & — & — \\
FM Tradeoff & $0.35$ & — & — & $1$ & — & — & $1$ & — & — \\
Target Par. & — & — & $0.15$ & — & — & $0.2$ & — & — & $0.1$ \\
SVD Par. & — & — & $1$ & — & — & $1$ & — & — & $1$ \\
Laplacian & — & — & laplac1 & — & — & laplac1 & — & — & laplac1 \\
AP & — & — & gauss & — & — & gauss & — & — & gauss \\
\bottomrule
\end{tabular}

\end{table*}

\begin{table*}
    \centering
    \tiny
    \caption{Hyperparameters for SD-AGMM.}
    \label{table:hyper-sd-agmm}
\begin{tabular}{lccc>{\columncolor[HTML]{EEEEEE}}c>{\columncolor[HTML]{EEEEEE}}c>{\columncolor[HTML]{EEEEEE}}cccc}
\toprule
Dataset & \multicolumn{3}{c}{Office31} & \multicolumn{3}{>{\columncolor[HTML]{EEEEEE}}c}{OfficeHome} & \multicolumn{3}{c}{VisDA2017} \\
Framework & FixMatch & Plain & SPA & FixMatch & Plain & SPA & FixMatch & Plain & SPA \\
\midrule
LR & $0.001$ & $0.1$ & $0.01$ & $0.005$ & $0.15$ & $0.03$ & $0.001$ & $0.1$ & $0.01$ \\
Momentum & $0.8$ & $0$ & $0.2$ & $0.9$ & $0.1$ & $0.3$ & $0.95$ & $0$ & $0.8$ \\
Bottleneck Dim & 256 & 256 & 256 & 256 & 1024 & 256 & 256 & 256 & 256 \\
Epoch Tradeoff & $1.5$ & $0.65$ & $1$ & $0.85$ & $0.25$ & $1$ & $0.1$ & $0.25$ & $1$ \\
Adv. LR & $0.0001$ & $0.0001$ & $1 \times 10^{-05}$ & $0.0001$ & $1 \times 10^{-05}$ & $0.01$ & $0.0001$ & $0.0001$ & $1 \times 10^{-05}$ \\
Adv. Dim & 512 & 512 & 1024 & 1024 & 512 & 1024 & 512 & 256 & 1024 \\
GMM LR & $0.05$ & $1 \times 10^{-06}$ & $0.0001$ & $1 \times 10^{-05}$ & $1 \times 10^{-06}$ & $0.0001$ & $0.01$ & $1 \times 10^{-06}$ & $0.0001$ \\
GMM \# Comp. & 8 & 4 & 8 & 16 & 4 & 8 & 4 & 4 & 8 \\
Threshold & $0.7$ & — & — & $0.85$ & — & — & $0.85$ & — & — \\
FM Tradeoff & $0.1$ & — & — & $0.35$ & — & — & $0.35$ & — & — \\
Target Par. & — & — & $0.1$ & — & — & $0.2$ & — & — & $0.1$ \\
SVD Par. & — & — & $1$ & — & — & $1$ & — & — & $1$ \\
Laplacian & — & — & laplac1 & — & — & laplac1 & — & — & laplac1 \\
AP & — & — & gauss & — & — & gauss & — & — & gauss \\
\bottomrule
\end{tabular}

\end{table*}

\begin{table*}
    \centering
    \tiny
    \caption{Hyperparameters for SD-AVAE. Bottleneck dimension, adversarial dimension, and momentum are ommitted because they were the same across all three datasets. Their values were 256, 512, and 0, respectively.}
    \label{table:hyper-sd-avae}
\begin{tabular}{lccc>{\columncolor[HTML]{EEEEEE}}c>{\columncolor[HTML]{EEEEEE}}c>{\columncolor[HTML]{EEEEEE}}cccc}
\toprule
Dataset & \multicolumn{3}{c}{Office31} & \multicolumn{3}{>{\columncolor[HTML]{EEEEEE}}c}{OfficeHome} & \multicolumn{3}{c}{VisDA2017} \\
Framework & FixMatch & Plain & SPA & FixMatch & Plain & SPA & FixMatch & Plain & SPA \\
\midrule
LR & $0.001$ & $0.1$ & $0.02$ & $0.005$ & $0.1$ & $0.03$ & $0.005$ & $0.05$ & $0.01$ \\
Momentum & $0.8$ & $0$ & $0.8$ & $0.9$ & $0$ & $0.2$ & $0.9$ & $0$ & $0.8$ \\
Bottleneck Dim & 256 & 256 & 256 & 256 & 256 & 256 & 256 & 256 & 256 \\
Epoch Tradeoff & $1.2$ & $0.5$ & $2$ & $1.9$ & $0.5$ & $1$ & $1.9$ & $0.5$ & $1$ \\
Adv. LR & $0.0001$ & $0.1$ & $0.01$ & $0.0001$ & $0.1$ & $0.01$ & $0.0001$ & $0.1$ & $0.01$ \\
Adv. Dim & 512 & 512 & 1024 & 512 & 512 & 1024 & 512 & 512 & 1024 \\
VAE LR & $1 \times 10^{-05}$ & $1 \times 10^{-08}$ & $0.001$ & $0.001$ & $1 \times 10^{-08}$ & $0.001$ & $0.001$ & $1 \times 10^{-08}$ & $0.001$ \\
Hidden Dim & 128 & 256 & 512 & 256 & 256 & 512 & 256 & 256 & 512 \\
Latent Dim & 32 & 256 & 128 & 64 & 64 & 128 & 64 & 256 & 128 \\
VAE Steps & 1 & 1 & 1 & 1 & 1 & 1 & 1 & 1 & 1 \\
Threshold & $0.85$ & — & — & $0.9$ & — & — & $0.75$ & — & — \\
FM Tradeoff & $0.35$ & — & — & $1$ & — & — & $1$ & — & — \\
Target Par. & — & — & $0.1$ & — & — & $0.2$ & — & — & $0.1$ \\
SVD Par. & — & — & $1$ & — & — & $1$ & — & — & $1$ \\
Laplacian & — & — & laplac1 & — & — & laplac1 & — & — & laplac1 \\
AP & — & — & gauss & — & — & gauss & — & — & gauss \\
\bottomrule
\end{tabular}

\end{table*}

\begin{table*}
    \centering
    \tiny
    \caption{Hyperparameters for SD-KGAU.}
    \label{table:hyper-sd-kgau}
\begin{tabular}{lccc>{\columncolor[HTML]{EEEEEE}}c>{\columncolor[HTML]{EEEEEE}}c>{\columncolor[HTML]{EEEEEE}}cccc}
\toprule
Dataset & \multicolumn{3}{c}{Office31} & \multicolumn{3}{>{\columncolor[HTML]{EEEEEE}}c}{OfficeHome} & \multicolumn{3}{c}{VisDA2017} \\
Framework & FixMatch & Plain & SPA & FixMatch & Plain & SPA & FixMatch & Plain & SPA \\
\midrule
LR & $0.001$ & $0.1$ & $0.01$ & $0.01$ & $0.15$ & $0.01$ & $0.01$ & $0.15$ & $0.01$ \\
Momentum & $0.95$ & $0$ & $0.1$ & $0.8$ & $0$ & $0.1$ & $0.95$ & $0$ & $0.1$ \\
Bottleneck Dim & 256 & 256 & 256 & 256 & 256 & 256 & 256 & 256 & 256 \\
Epoch Tradeoff & $0.4$ & $0.5$ & $1$ & $0.4$ & $0.5$ & $1$ & $0.25$ & $0.25$ & $1$ \\
Kernel Bandwidth & $1$ & $1$ & $10$ & $1$ & $1$ & $10$ & $1$ & $1$ & $10$ \\
Threshold & $0.9$ & — & — & $0.9$ & — & — & $0.9$ & — & — \\
FM Tradeoff & $1$ & — & — & $1$ & — & — & $1$ & — & — \\
Target Par. & — & — & $0.2$ & — & — & $0.2$ & — & — & $0.2$ \\
SVD Par. & — & — & $1$ & — & — & $1$ & — & — & $1$ \\
Laplacian & — & — & laplac1 & — & — & laplac1 & — & — & laplac1 \\
AP & — & — & gauss & — & — & gauss & — & — & gauss \\
\bottomrule
\end{tabular}

\end{table*}

\begin{table*}
    \centering
    \tiny
    \caption{Hyperparameters for SD-KGMM.}
    \label{table:hyper-sd-kgmm}
\begin{tabular}{lccc>{\columncolor[HTML]{EEEEEE}}c>{\columncolor[HTML]{EEEEEE}}c>{\columncolor[HTML]{EEEEEE}}cccc}
\toprule
Dataset & \multicolumn{3}{c}{Office31} & \multicolumn{3}{>{\columncolor[HTML]{EEEEEE}}c}{OfficeHome} & \multicolumn{3}{c}{VisDA2017} \\
Framework & FixMatch & Plain & SPA & FixMatch & Plain & SPA & FixMatch & Plain & SPA \\
\midrule
LR & $0.003$ & $0.1$ & $0.01$ & $0.01$ & $0.1$ & $0.01$ & $0.01$ & $0.05$ & $0.01$ \\
Momentum & $0.8$ & $0.05$ & $0.1$ & $0.8$ & $0.05$ & $0.1$ & $0.95$ & $0.05$ & $0.3$ \\
Bottleneck Dim & 256 & 1024 & 256 & 256 & 256 & 256 & 256 & 256 & 256 \\
Epoch Tradeoff & $1.5$ & $0.25$ & $1$ & $0.1$ & $0.25$ & $1$ & $0.15$ & $10$ & $1$ \\
Kernel Bandwidth & $1$ & $0.5$ & $1$ & $1$ & $1$ & $1$ & $1$ & $10$ & $1$ \\
GMM LR & $0.001$ & $1 \times 10^{-05}$ & $0.1$ & $0.001$ & $1$ & $0.01$ & $0.001$ & $0.001$ & $0.01$ \\
GMM \# Comp. & 8 & 8 & 4 & 8 & 8 & 4 & 8 & 8 & 4 \\
Threshold & $0.9$ & — & — & $0.9$ & — & — & $0.7$ & — & — \\
FM Tradeoff & $1$ & — & — & $1$ & — & — & $0.5$ & — & — \\
Target Par. & — & — & $0.2$ & — & — & $0.2$ & — & — & $0.2$ \\
SVD Par. & — & — & $1$ & — & — & $1$ & — & — & $1$ \\
Laplacian & — & — & laplac1 & — & — & laplac1 & — & — & laplac1 \\
AP & — & — & gauss & — & — & gauss & — & — & gauss \\
\bottomrule
\end{tabular}

\end{table*}

\subsection{Scarce target setting for existing methods}\label{subsec:scarce-target-mods}

We used existing implementations of \ac{uda} methods for baseline methods, with modifications to run experiments in the scarce target setting.
The baseline methods DANN, JAN, AFN, MDD, MCC, and ERM were implemented using the Transfer Learning Library (TLL) \citep{tllib, jiang2022transferability}.
The only modification to TLL's methods was to the function used to retrieve the datasets.
We modified it to take three extra parameters: the percent of target data used in training, a minimum number of target samples that will override the percent if it would be too small (always set to 32), and a random seed, to ensure that all methods get the same sample of target data.
The validation and test data sets are unchanged.
SDAT's implementation \cite{rangwani2022closer} closely follows TLL, and we applied the same modifications to implement experiments in the scarce target setting.

Implementations for SPA \cite{xiao2024spa}, fDAL \cite{acuna2021f}, and NWD \cite{chen2022reusing} had more differences with TLL's implementation.
For all three methods, we replaced the construction of data transforms and dataloaders to use methods from TLL, to ensure identical preprocessing and augmentation.
For SPA, we also modified the logging and check-pointing behavior to follow the format used in TLL.

\subsection{Hardware and runtime}

Experiments were completed on an A100 GPU with 8 CPUs.
The runtimes for a subset of methods, including Stein discrepancy-based methods, ERM, and the three most closely related baseline methods, JAN, Fixmatch, and SPA, are reported in Table \ref{table:runtimes}.
We report the average time to complete one iteration on the VisDA2017 dataset, using ResNet 101 as a feature extractor.
For each run, we measured the time to train one epoch and divided by the number of iterations in that epoch to estimate the average time per iteration.
These per-run estimates were then averaged across the three runs.

\begin{table}[ht!]
    \centering
\begin{tabular}{lrr}
\toprule
 & Mean & Std. \\
Method &  &  \\
\midrule
ERM & 0.072 & 0.003 \\
FM-SD-AGAU & 0.244 & 0.005 \\
FM-SD-AGMM & 0.397 & 0.001 \\
FM-SD-AVAE & 0.425 & 0.012 \\
FM-SD-KGAU & 0.349 & 0.007 \\
FM-SD-KGMM & 0.441 & 0.025 \\
FixMatch & 0.268 & 0.009 \\
JAN & 0.133 & 0.001 \\
Reg-SD-KGAU & 0.232 & 0.020 \\
Reg-SD-KGMM & 0.427 & 0.003 \\
SD-AGAU & 0.183 & 0.013 \\
SD-AGMM & 0.333 & 0.002 \\
SD-AVAE & 0.221 & 0.006 \\
SD-KGAU & 0.189 & 0.001 \\
SD-KGMM & 0.349 & 0.003 \\
SPA & 0.756 & 0.051 \\
SPA-SD-AGAU & 0.717 & 0.061 \\
SPA-SD-AGMM & 0.981 & 0.115 \\
SPA-SD-AVAE & 0.884 & 0.037 \\
SPA-SD-KGAU & 0.752 & 0.052 \\
SPA-SD-KGMM & 0.979 & 0.058 \\
\bottomrule
\end{tabular}
\caption{Run times for Stein discrepancy-based methods and four baseline methods, ERM, JAN, FixMatch, and SPA. Report mean and standard deviation of seconds to run one training step of the algorithm.}
\label{table:runtimes}
\vspace{2.5em}
\end{table}

Replacing \ac{mmd} by \ac{ksd} adds minimal time to the computation when the target distribution is Gaussian.
More complicated target distributions, including GMM and VAE,  lead to greater increases in runtime and computational cost.
All of the SPA methods are much slower, likely because they involve constructing graphs from the training data and then computing spectral properties of the graph, both of which are computationally expensive.
Nevertheless, replacing DANN, the default domain discrepancy measure in SPA, by \ac{ksd} does not significantly increase the computational cost for the Gaussian target distribution.

\section{Use of existing code}\label{app:existing-assets}

We use several existing implementations of \ac{uda} methods; TLL , as well as code implementing of \ac{ksd} and GMMs.

The baseline methods DANN, JAN, AFN, MDD, MCC, and ERM were implemented using the Transfer Learning Library (TLL) \citep{tllib, jiang2022transferability}, with small modifications to implement the scarce target setting, described in \S \ref{subsec:scarce-target-mods}.
TLL is accessible at \hyperlink{https://github.com/thuml/Transfer-Learning-Library/}{https://github.com/thuml/Transfer-Learning-Library/} and is distributed under an MIT license.
SPA has an official implementation accessible at \hyperlink{https://github.com/CrownX/SPA}{https://github.com/CrownX/SPA}.
NWD has an official implementation at \hyperlink{https://github.com/xiaoachen98/DALN}{https://github.com/xiaoachen98/DALN}.
f-DAL has an official implementation at \hyperlink{https://github.com/nv-tlabs/fDAL}{https://github.com/nv-tlabs/fDAL}, and is distributed under an NVIDIA Source Code License, which allows non-commercial and research use.
SDAT has an official implementation accessible at \hyperlink{https://github.com/val-iisc/SDAT/}{https://github.com/val-iisc/SDAT/} and is distributed under an MIT license.

Our code for \ac{ksd} is based on code from \cite{korba_kernel_2021}.
The code can be accessed at \hyperlink{https://github.com/pierreablin/ksddescent}{https://github.com/pierreablin/ksddescent} and is distributed under an MIT license.
Our implementation of GMMs is based on code accessible at \hyperlink{https://github.com/ldeecke/gmm-torch/}{https://github.com/ldeecke/gmm-torch/}, which is distributed under an MIT license.
Raytune was used for hyperparamter tuning.
Information about installing and using Raytune can be found at \hyperlink{https://docs.ray.io/en/latest/tune/index.html}{https://docs.ray.io/en/latest/tune/index.html}.
The source code can be found at \hyperlink{https://github.com/ray-project/ray/}{https://github.com/ray-project/ray/} and is distributed under an Apache 2.0 license.
Finally, our code for regularized \ac{ksd} is based on code from \citet{hagrass2024minimax}, adapted from R to Python.

\end{appendices}

\end{document}